\documentclass[11pt]{article}

\usepackage[preprint]{acl}

\usepackage{times}
\usepackage{latexsym}

\usepackage[T1]{fontenc}

\usepackage[utf8]{inputenc}

\usepackage{microtype}

\usepackage{inconsolata}

\usepackage{graphicx}

\usepackage{booktabs} 
\usepackage{multirow} 
\usepackage{amsmath}
\usepackage{tabularx}
\usepackage{placeins}

\renewcommand{\topfraction}{0.95}
\renewcommand{\textfraction}{0.05}
\renewcommand{\floatpagefraction}{0.9}

\title{Understanding New-Knowledge-Induced Factual Hallucinations in LLMs:\\Analysis and Interpretation}

\author{
    Renfei Dang$^{1}$$^*$\text{,} 
    Peng Hu$^{1}$$^*$\text{,} 
    \textbf{Zhejian Lai}$^{1}$\text{,} 
    \textbf{Changjiang Gao}$^{1}$\text{,} 
    \textbf{Min Zhang}$^{2}$\textbf{,} 
    \textbf{Shujian Huang}$^{1}$$^\dagger$ \\
    $^{1}$ \text{National Key Laboratory for Novel Software Technology, Nanjing University, Nanjing, China} \\
    $^{2}$ \text{Huawei Translation Services Center, Beijing, China} \\
    \small\texttt{\{dangrf,hup,laizj,gaocj\}@smail.nju.edu.cn},
    \small\texttt{huangsj@nju.edu.cn},
    \small\texttt{zhangmin186@huawei.com}
}

\begin{document}
\maketitle

\renewcommand{\thefootnote}{\fnsymbol{footnote}}
\footnotetext[1]{Equal contribution.}
\footnotetext[2]{Corresponding author.}
\renewcommand{\thefootnote}{\arabic{footnote}}

\begin{abstract}
Prior works have shown that fine-tuning on new knowledge can induce factual hallucinations in large language models (LLMs), leading to incorrect outputs when evaluated on previously known information. However, the specific manifestations of such hallucination and its underlying mechanisms remain insufficiently understood. Our work addresses this gap by designing a controlled dataset \textit{Biography-Reasoning}, and conducting a fine-grained analysis across multiple knowledge types and two task types, including knowledge question answering (QA) and knowledge reasoning tasks. We find that hallucinations not only severely affect tasks involving newly introduced knowledge, but also propagate to other evaluation tasks. Moreover, when fine-tuning on a dataset in which a specific knowledge type consists entirely of new knowledge, LLMs exhibit elevated hallucination tendencies. This suggests that the degree of unfamiliarity within a particular knowledge type, rather than the overall proportion of new knowledge, is a stronger driver of hallucinations. Through interpretability analysis, we show that learning new knowledge weakens the model's attention to key entities in the input question, leading to an over-reliance on surrounding context and a higher risk of hallucination. Conversely, reintroducing a small amount of known knowledge during the later stages of training restores attention to key entities and substantially mitigates hallucination behavior. Finally, we demonstrate that disrupted attention patterns can propagate across lexically similar contexts, facilitating the spread of hallucinations beyond the original task.
\end{abstract}

\section{Introduction}
Large language models (LLMs) acquire rich factual knowledge during pre-training on massive text corpora \citep{petroni2019language,cohen2023crawling}, and are subsequently post-trained to follow human instructions and perform a wide range of downstream tasks \citep{ouyang2022training,wei2022chain}.

\begin{figure*}[htb]
    \centering
    \includegraphics[width=0.95\textwidth]{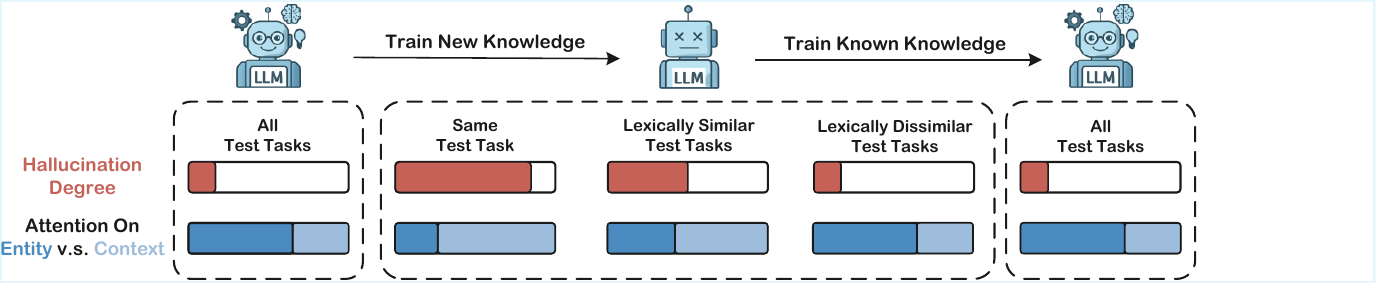}
    \caption{The impact of learning new knowledge on attention patterns and hallucination behavior. Training a model on new knowledge can induce factual hallucinations, whose severity is correlated with the model's attention scores on key entities. Moreover, hallucinations are more likely to occur on test instances whose input contexts are lexically similar to the contexts of training tasks containing unknown knowledge. By injecting a small amount of known knowledge at the end of training, this issue can be effectively mitigated.}
    \label{fig:intro}
\end{figure*}

However, during the Supervised Fine-Tuning (SFT) phase, models may encounter new knowledge not covered in pre-training. Prior research \citep{ghosal2024understanding,lin2023ra,ovadia2023fine,gekhman2024does,sun2025new} suggest that introducing new knowledge in the post-training phase increases the risk of factual hallucinations, where models generate fabricated yet plausible statements. This occurs because, when models learn new knowledge, they may erroneously generate related information in irrelevant contexts~\citep{gekhman2024does,sun2025new}. These studies primarily focus on the effects within knowledge-intensive QA tasks during SFT, and we advance this line of research by investigating the fine-grained manifestations and underlying causes of hallucinations.

To support this investigation, we construct a controlled experimental dataset \textit{Biography-Reasoning}. 
The dataset is composed of biographical entities and their four attributes,
which serve as four knowledge types. We further design twelve reasoning tasks using these knowledge. 
By controlling the proportion of known and unknown knowledge within different types and tasks in the training data, we systematically analyze the impact of learning new knowledge on hallucination risks. 

Our experiments reveal that training on unknown knowledge significantly elevates hallucination risks in the same task, while also inducing non-negligible hallucination effects on other out-of-domain test tasks. Importantly, we further find that when a knowledge type consists entirely of new knowledge, even a small amount of such data can markedly increase hallucination tendencies.

Through further interpretability analyses, we find that learning new knowledge significantly weakens the model’s attention to key entities in the question, thereby triggering factual hallucinations. In contrast, training on known knowledge strengthens the model’s attention to key entities. Motivated by this observation, we introduce a simple training method KnownPatch, which restores disrupted attention patterns by injecting a small amount of known knowledge during the later stages of training, and thus alleviates factual hallucinations. Finally, by constructing carefully designed variants of reasoning tasks, we demonstrate that lexical similarity (measured by token overlap between contexts), rather than semantic similarity of the contexts, is the primary driver of hallucination propagation across tasks. These findings are visually presented in Figure \ref{fig:intro}.

The main contributions of this paper are:
\begin{itemize}
    \item \textbf{Fine-Grained Analysis}: A detailed analysis across knowledge types and task types reveals the manifestations of new-knowledge-induced hallucinations, showing that when all knowledge within a specific type is entirely unknown, it is more likely to trigger severe hallucinations, even on unrelated QA test sets.
    \item \textbf{Mechanism Interpretability}: An analysis of attention mechanisms shows that learning new knowledge reduces attention to key question entities, causing hallucinations. In addition, lexically similar contexts facilitate the spread of these attention patterns, enabling cross-task hallucination effects.
\end{itemize}

\section{Related Work}

\paragraph{New Knowledge and Hallucinations}
Existing studies have indicated that introducing new knowledge into LLMs may trigger hallucinations \citep{ghosal2024understanding,lin2023ra,ovadia2023fine}. Subsequent works have provided deeper analyses of this phenomenon \citep{gekhman2024does,kang2024unfamiliar,sun2025new}. \citet{gekhman2024does} found that as the proportion of new knowledge in fine-tuning data increases, the model's hallucination tendency intensifies. \citet{kang2024unfamiliar} analyzed that when fine-tuned LLMs encounter unknown queries during testing, their responses imitate those associated with unknown examples in the fine-tuning data. From the perspective of token probabilities, \citet{sun2025new} show that after learning new knowledge, the generation probabilities of answer entity tokens increase significantly even in irrelevant contexts, suggesting that the model may over-generalize newly acquired knowledge and consequently produce hallucinations.
However, previous studies focus mainly on closed-book QA settings with mixed knowledge types during training, while our controlled setup disentangles them to provide a more detailed analysis of new knowledge-induced hallucinations across types and tasks. 
Furthermore, we also investigate the underlying mechanisms of these phenomenon through an analysis of attention weights. 


\paragraph{Reducing Hallucinations}
Numerous studies are currently exploring ways to mitigate model hallucinations. A common approach involves providing additional relevant context to the model to reduce hallucinations during generation, such as through retrieval from knowledge bases or leveraging other large models to generate context \citep{shuster2021retrieval,sun2022recitation,asai2024self,feng2023knowledge}. Additionally, some research explicitly avoids hallucination risks by refusing to answer uncertain or unfamiliar questions \citep{yadkori2024mitigating,zhu2025grait,duwal2025mka}. In another direction, many studies encourage the model to generate more known knowledge from pre-training, for example, by promoting factual outputs via reinforcement learning \citep{rafailov2023direct,kang2024unfamiliar,li2025hallucination,gu2025mask} or by training only on known knowledge during supervised fine-tuning \citep{lin2024flame,ghosal2024understanding,liu2024fictitious} to enhance the model. Our work builds on 
SFT with known knowledge approach, 
but rather than pursuing comprehensive filtering across all training data, KnownPatch only introduces a small number of known knowledge samples in the later stages of training, and  alleviates the model's tendency for hallucination.
\section{Methodology of Analyzing Hallucinations}

\label{sec:methodology}
We aim to systematically investigate factual hallucinations in LLMs caused by learning different knowledge-related tasks. However, in real-world datasets, most factual knowledge may have already been seen by LLMs during pre-training, making it difficult to precisely control whether the knowledge being learned is new to the model. To address this limitation, we construct a synthetic dataset named \textit{Biography-Reasoning}, which allows a controllable examination of hallucination behaviors under varying knowledge types and task types. 

\subsection{\textit{Biography-Reasoning} Dataset}
Following the data construction methodologies of \citet{allen-zhu2024physics,zheng2025spurious}, we design the \textit{Biography-Reasoning} dataset. The dataset centers on individuals as the key entities, with each person associated with four attributes: birth year, death year, major, and university. We refer to the same attribute of different individuals as a knowledge type. 

Our dataset includes two types of knowledge-related tasks: knowledge QA and knowledge-based reasoning tasks. For knowledge QA tasks, questions are formulated by directly querying one of the attributes given the person’s name. Each task consists of questions on a single type, resulting in four QA tasks (e.g., Major\_QA). 

For knowledge-based reasoning tasks, we design three types of chain-of-thought-requiring reasoning tasks. Specifically, these include: 
\begin{itemize}
    \item \textbf{Single Reasoning}: extracting one attribute from a single entity and performing a simple reasoning process; 
    \item \textbf{Comparative Reasoning} : extracting one attribute from each of two entities and performing comparative reasoning between them; 
    \item \textbf{Novel Reasoning}: extracting one attribute from a single entity and performing a newly defined reasoning task, such as mathematical or symbolic reasoning. 
\end{itemize}
Table \ref{tab:dataexample} presents examples of the constructed questions. The reasoning tasks are intentionally designed to be more complex than mere knowledge extraction as QA problems. Some of them require auxiliary knowledge (e.g., the major Dentistry belongs to the field Medicine), which the model is expected to contain. To further guarantee the model’s proficiency, we additionally collect and train on these auxiliary facts. 

For each knowledge type we construct one QA and three reasoning tasks, leading to a total of 4 QA and 12 reasoning tasks per individual. Further dataset details can be found in Appendix \ref{app:dataset_details}.

\begin{table}[ht]
    \centering
    \footnotesize
\begin{tabularx}{\linewidth}{>{\raggedright\arraybackslash}p{0.20\linewidth}X}
        \toprule
        \textbf{Category} & \textbf{Example} \\
        \midrule
        QA &
        Question: What major did Darreus Hsiao study?\newline
        Answer: Dentistry \\
        \midrule
        Single Reasoning &
        Question: What field does Darreus Hsiao's major belong to?\newline
        Answer: Darreus Hsiao's major is Dentistry. Dentistry belongs to Medicine.\newline
        The answer is: Medicine \\
        \midrule
        Comparative Reasoning &
        Question: Do Darreus Hsiao and Virgus Hong's majors belong to the same field?\newline
        Answer: Darreus Hsiao's major is Dentistry. Dentistry belongs to Medicine. Virgus Hong's major is Nursing. Nursing belongs to Medicine. Medicine and Medicine are the same.\newline
        The answer is: YES \\
        \midrule
        Novel Reasoning &
        Question: What is the sequence of odd-positioned letters in the first word of Darreus Hsiao's major name?\newline
        Answer: Darreus Hsiao's major is Dentistry. The first word of `Dentistry' is `Dentistry'. The spelling of Dentistry is D, E, N, T, I, S, T, R, Y. The sequence of odd-positioned letters in `Dentistry' is DNITY.\newline
        The answer is: DNITY \\
        \bottomrule
    \end{tabularx}
    \caption{Examples of the QA and reasoning tasks in \textit{Biography-Reasoning}, associated with the Major type.}
    \label{tab:dataexample}
\end{table}

\subsection{Controlled Study Design}

To examine factual hallucinations caused by training with tasks containing new knowledge, we need to discriminate \textbf{known} and \textbf{unknown} knowledge, control their usage during training, and evaluate related hallucinations.

Since initially the model has no exposure to any knowledge of our synthetic dataset, we prepare the study by continue pre-training the model with a subset of the knowledge, which becomes \textbf{known} to the model; and keep another subset of the knowledge as \textbf{unknown}. By mixing the constructed questions from known and unknown knowledge in varying proportions, we are able to create situations where different proportion of newly introduced knowledge participates in training.

To evaluate how training leads to hallucinations, we reserve another subset of knowledge as \textbf{test} knowledge. The test knowledge are continue pre-trained together with the known knowledge during the preparation, but are kept away from further training. 
Therefore, the difference in performance on test set with and without unknown knowledge in the training data indicates the influence of factual hallucinations induced by training new knowledge. In addition, we use the real-world ENTITYQUESTIONS dataset \citep{sciavolino-etal-2021-simple} derived from Wikidata \citep{10.1145/2629489} (denoted as Wiki) as an out-of-distribution (OOD) test set to provide a more robust evaluation. 

\subsection{Models and Setups}
\label{experimentalsetup}
We conduct experiments primarily using the Qwen2.5-1.5B model \citep{qwen2.5}. As supplementary validation, we also perform key experiments on Llama3.2-1B \citep{grattafiori2024llama3herdmodels}, Qwen3-8B-Base \citep{qwen3technicalreport} and Qwen2.5-32B \citep{qwen2.5} to assess generalization across model scales and architectures, with their results provided in Appendix \ref{app:different_models}. 

As our experiments are conducted on base models, we first apply SFT to endow them with the ability to answer questions in the evaluation sets. For QA analysis, SFT is conducted solely on knowledge QA data, whereas for reasoning, the model is trained jointly on both task types to ensure general reasoning competence. 

All experiments are performed with full-parameter fine-tuning. Detailed hyperparameters are provided in the Appendix \ref{app:training_details}. 
In the SFT phase, we default to training for 3 epochs, but we also provide results for training 1, 5, and 20 epochs in Appendix~\ref{app:different_epochs}. The settings of 1, 3, and 5 epochs simulate typical training schedules in practice, whereas 20 epochs allow the model to acquire most of the knowledge in the training set, even for previously unknown information. 

Following \citet{allen-zhu2024physics} and \citet{ gekhman2024does}, we adopt Exact Match (EM) as the metric for both knowledge QA tasks and reasoning tasks to assess the accuracy of the final answers. Given that all test knowledge are known to the model, and the training and testing formats are consistent, there are no cases where the answers are correct but incorrectly formatted. To ensure the generality of our conclusions, we report the standard deviation of accuracy where applicable.

\section{Hallucination Analysis}

Using the \textit{Biography-Reasoning} dataset, we conduct a systematic study on factual hallucinations induced by learning different tasks containing various types of new knowledge through SFT. We additionally report the impact of learning new knowledge during CPT on hallucinations in Appendix~\ref{app:cpt_results}.

\subsection{Knowledge QA}
\label{sec:qa_results}
In this section, we analyze the impact of training on new knowledge in QA tasks. A \emph{baseline model} is trained on samples constructed from the known knowledge of all four types. We then replace the knowledge of one entire type with unknown samples while keeping the other three types unchanged, resulting in four variant models. For each variant, we evaluate performance drop compared to the \emph{baseline model} on three groups of QA test sets: (1) \textbf{Same-Type QA (STQA)}: the QA test set whose knowledge type matches the type trained with unknown knowledge; (2) \textbf{Different-Type QA (DTQA)}: the QA test sets whose knowledge types differ from the type trained with unknown knowledge; (3) \textbf{Wiki}: the real-world QA test set used as OOD evaluation.

\begin{figure}[htb]
    \centering
    \includegraphics[width=0.5\textwidth]{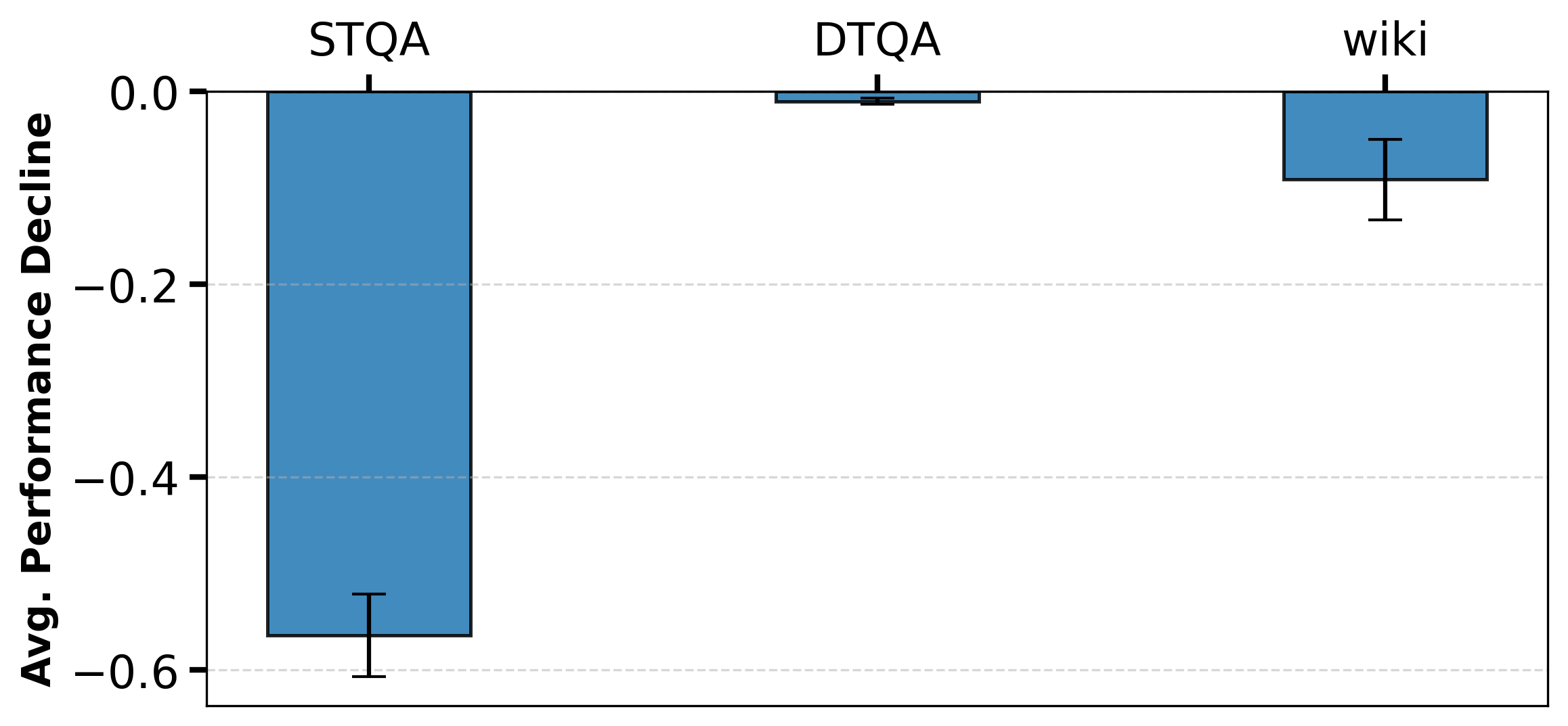}
    \caption{Average performance degradation (\%, mean with standard deviation) of four model variants. Detailed numerical results are reported in Appendix \ref{app:detailed_results}.}
    \label{tab:qa_results}
\end{figure}


\textbf{Learning new knowledge induces factual hallucinations within the same type, with some spillover effects to other types.} Figure~\ref{tab:qa_results} presents the performance drop averaged across the four variant models. Training on unknown knowledge leads to substantial performance drops on the STQA test set, reducing the accuracy by more than half. We also observe cross-type degradation, as training on one type negatively impacts average performance on others, including the real-world Wiki test set containing OOD knowledge. This confirms that learning new knowledge can induce hallucinations even on unrelated knowledge. Notably, the performance drop on DTQA is smaller than on Wiki, as the former consists entirely of known data in the training set, which greatly mitigates the effect. 

We further investigate how varying the proportion of unknown knowledge within a single type influences hallucination tendencies. Starting from the fully known-knowledge baseline, we progressively replace 5\%, 10\%, 20\%, 50\%, 80\%, and 100\% of the knowledge in one type with unknown knowledge, while still keeping the other three types entirely known. Two strategies are considered for handling the remaining known knowledge within the modified type: \textit{KeepKnown}, where the remaining known instances are retained, and \textit{RemoveKnown}, where they are excluded from training.

\begin{figure}[htb]
    \centering
    \includegraphics[width=0.5\textwidth]{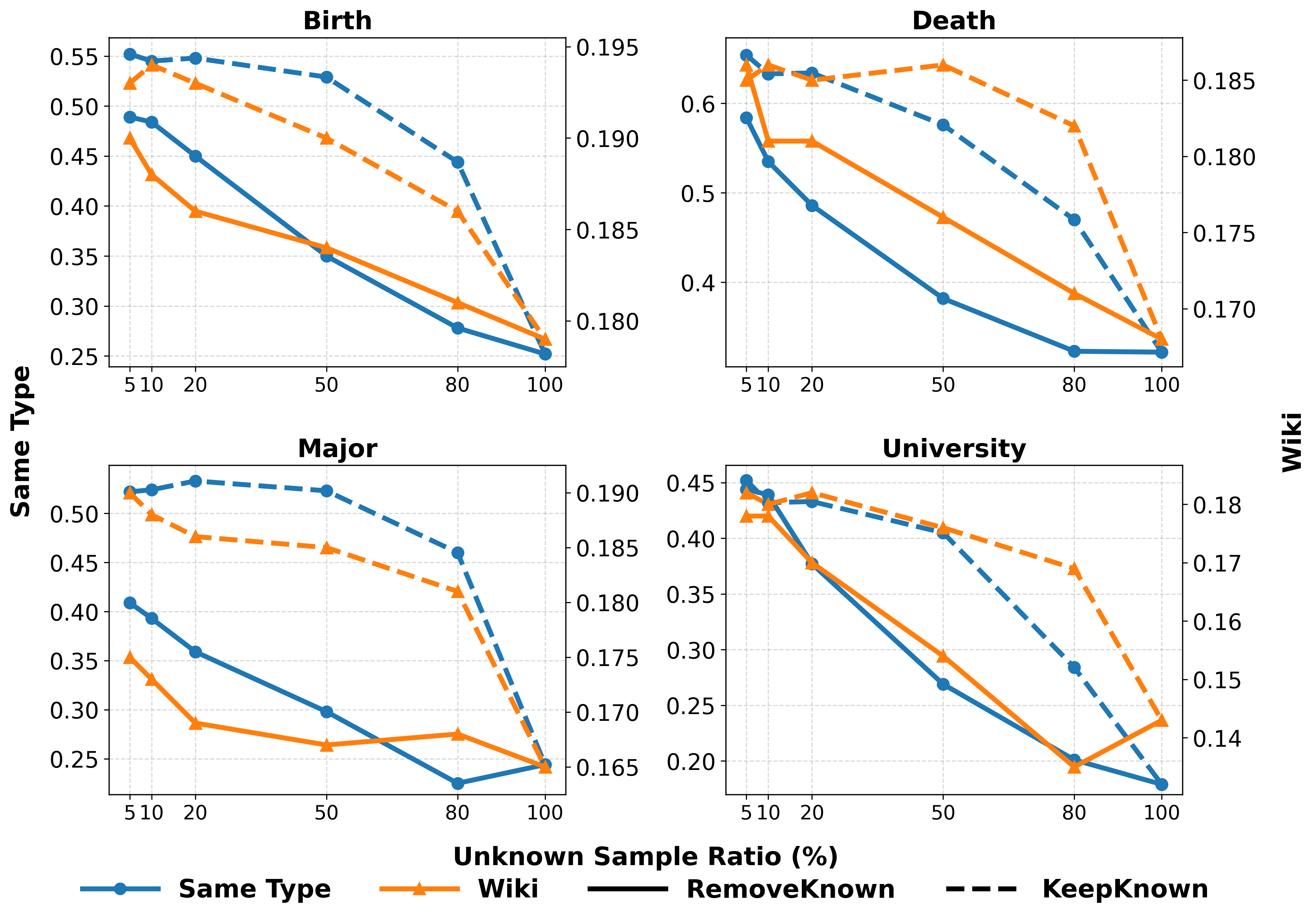}
    \caption{Performance under two settings with different proportions of unknown knowledge in the same type test set and Wiki test set.}
    \label{fig:qa_unknown_percentage}
\end{figure}

As shown in Figure~\ref{fig:qa_unknown_percentage}, the results across the four subplots are mutually corroborative, revealing a consistent pattern: \textbf{the higher the proportion of unknown knowledge, the more severe the hallucination.} In \textit{KeepKnown}, performance on both the same-type test set and the OOD Wiki test set degrades gradually at first, followed by a sharp decline as the unknown knowledge dominants. In contrast, \textit{RemoveKnown} exhibits a much faster degradation: even at low replacement ratios, the model already shows severe hallucination effects.

For the same replacement ratio, the key difference between \textit{KeepKnown} and \textit{RemoveKnown} lies in whether the knowledge type still contains any known instances.
We observe that this distinction has a substantial impact on model performance, with \textit{RemoveKnown} consistently underperforming \textit{KeepKnown}.
These observations suggest that \textbf{sparse but fully unknown knowledge types are more disruptive than those containing a mixture of known and unknown knowledge}, 
which differs from prior common understanding.

We additionally conduct the same set of experiments in this subsection using the real-world ENTITYQUESTIONS dataset. The results are reported in Appendix \ref{app:wiki_results}, and all findings are highly consistent with those obtained on the synthetic data.

\subsection{Knowledge-based Reasoning}
\label{sec:reasoning_results}

For reasoning-related experiments, we train the model on both reasoning and QA tasks to facilitate a more reliable evaluation across both test sets. The \emph{baseline model} is trained with all samples constructed from known knowledge. We then replace one reasoning task with instances derived from unknown knowledge and keep all other unchanged, resulting in 12 variant models.

We investigate how training on a knowledge-based reasoning task with unknown knowledge affects performance across different downstream tasks, including both knowledge-based reasoning and knowledge QA.
Specifically, we examine three groups of reasoning tasks:
(1) \textbf{Same-Type Same-Reasoning (STSR)}: the exact reasoning task that trained with unknown knowledge type;
(2) \textbf{Same-Type Different-Reasoning (STDR)}: different reasoning tasks within the unknown knowledge type; and
(3) \textbf{Different-Type Different-Reasoning (DTDR)}: all other reasoning tasks with different knowledge types. 
We also evaluate the model on the knowledge QA test groups defined in Section \ref{sec:qa_results}, namely STQA, DTQA, and Wiki.

\begin{figure}[htb]
    \centering
    \includegraphics[width=0.48\textwidth]{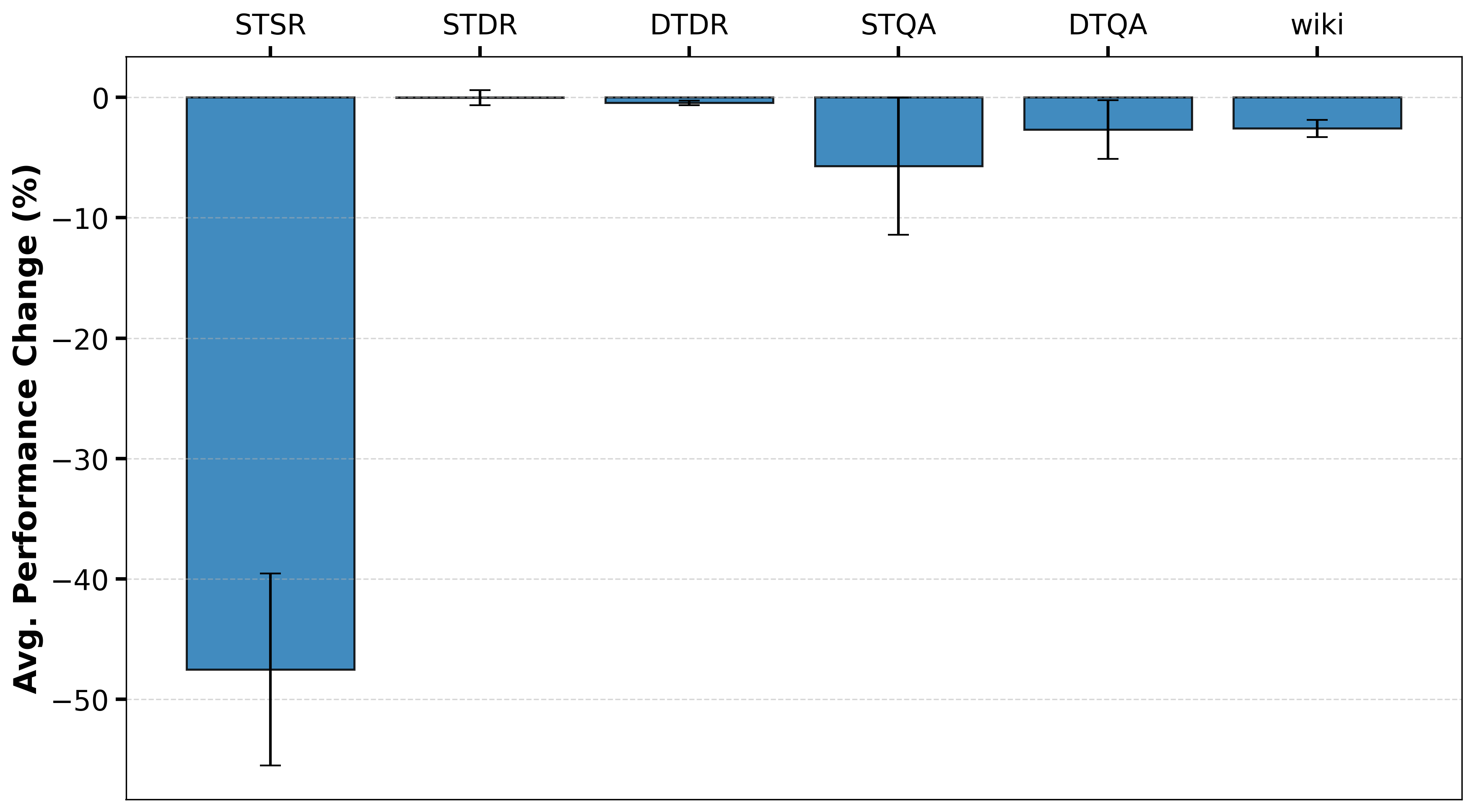}
    \caption{The impact of learning new knowledge in reasoning tasks on the average performance across different groups. Detailed results are presented in Appendix \ref{app:detailed_results}.}
    \label{fig:reasoning_big_table}
\end{figure}

We measure the relative performance change with respect to the \emph{baseline model} and compute the average difference within each of the six task groups. Results in Figure~\ref{fig:reasoning_big_table} show that learning reasoning tasks with new knowledge consistently induces performance degradation across all six groups. The overall trend aligns with previous findings: the most severe hallucinations occur in STSR, indicating strong intra-task interference.
Moreover, among other tested tasks, \textbf{QA test sets exhibit even stronger hallucinations than several seemingly more related reasoning tasks}: STQA, DTQA, and even the Wiki test set show greater performance degradation than STDR and DTDR.

\section{Interpretability Analysis}
\label{attentionanalysis}
In this section, we analyze the underlying mechanisms of new-knowledge-induced factual hallucinations by examining the relative changes in attention patterns and hallucination severity. Based on this analysis, we propose a simple training intervention, referred to as KnownPatch, which injects a small amount of known knowledge at the later training stage to restore attention patterns and alleviate hallucination behavior. Crucially, we also investigate the role of contextual similarity, providing evidence that hallucinations propagate primarily through shared lexical contexts.

\subsection{Attention Analysis Setup}
We measure how learning new knowledge alters the model's attention to key entities. In the \textit{Biography-Reasoning} dataset, the key entities are person names, so we quantify the model’s attention to the name tokens when generating the first token of the related knowledge.

Prior interpretability works suggest that knowledge retrieval and reasoning occur primarily in mid-to-late transformer layers \citep{wendler-etal-2024-llamas,zhao2024large}. We confirm this trend in our model by examining attention across layers in both QA and reasoning settings (as detailed in Appendix~\ref{app:attention_layer_details}). Figure~\ref{fig:layer_selection} shows that attention on key entities peaks in layers 12-24 (out of 28 layers), so we average attention over these layers in all subsequent analyses. We measure the relative change in entity attention by comparing models trained under new knowledge to the model trained entirely on known knowledge, i.e. the \emph{baseline model}. 


\begin{figure}[htb]
    \centering
    \includegraphics[width=0.48\textwidth]{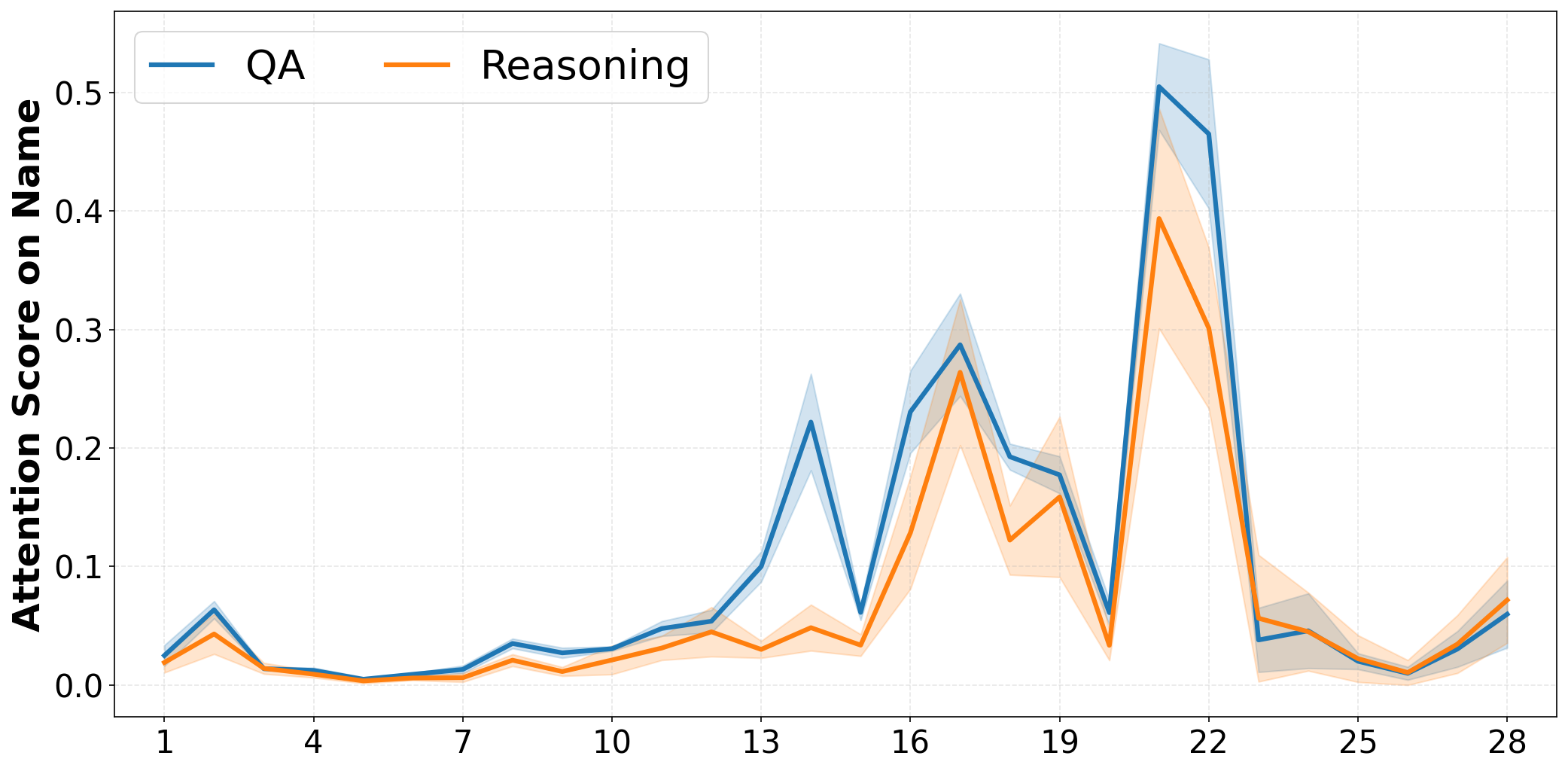}
    \caption{Attention score on the key entity name across layers in QA and reasoning training setups. The solid curves show the average attention score aggregated across all instances. The shaded regions represent the standard deviation.}
    \label{fig:layer_selection}
\end{figure}

\subsection{Correlation between Attention and Hallucination}
\label{sec:interpretability}

\begin{figure}[hb]
    \centering
    \includegraphics[width=0.5\textwidth]{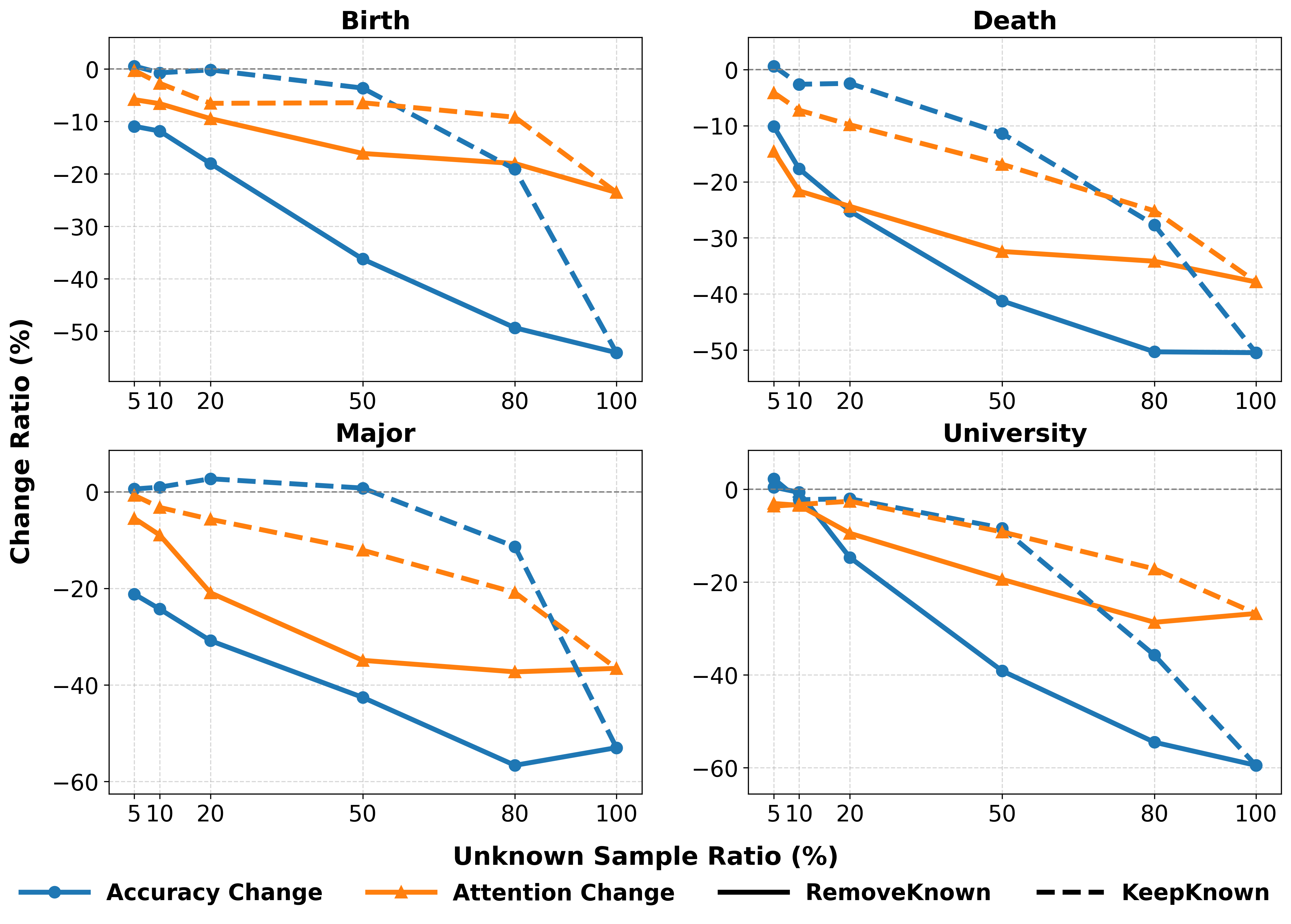}
    \caption{Accuracy and attention score changes with different unknown data ratio in certain type.}
    \label{fig:qa_unknown_percentage_interpretability_lines}
\end{figure}
\begin{figure}[hbt]
    \centering
    \includegraphics[width=0.48\textwidth]{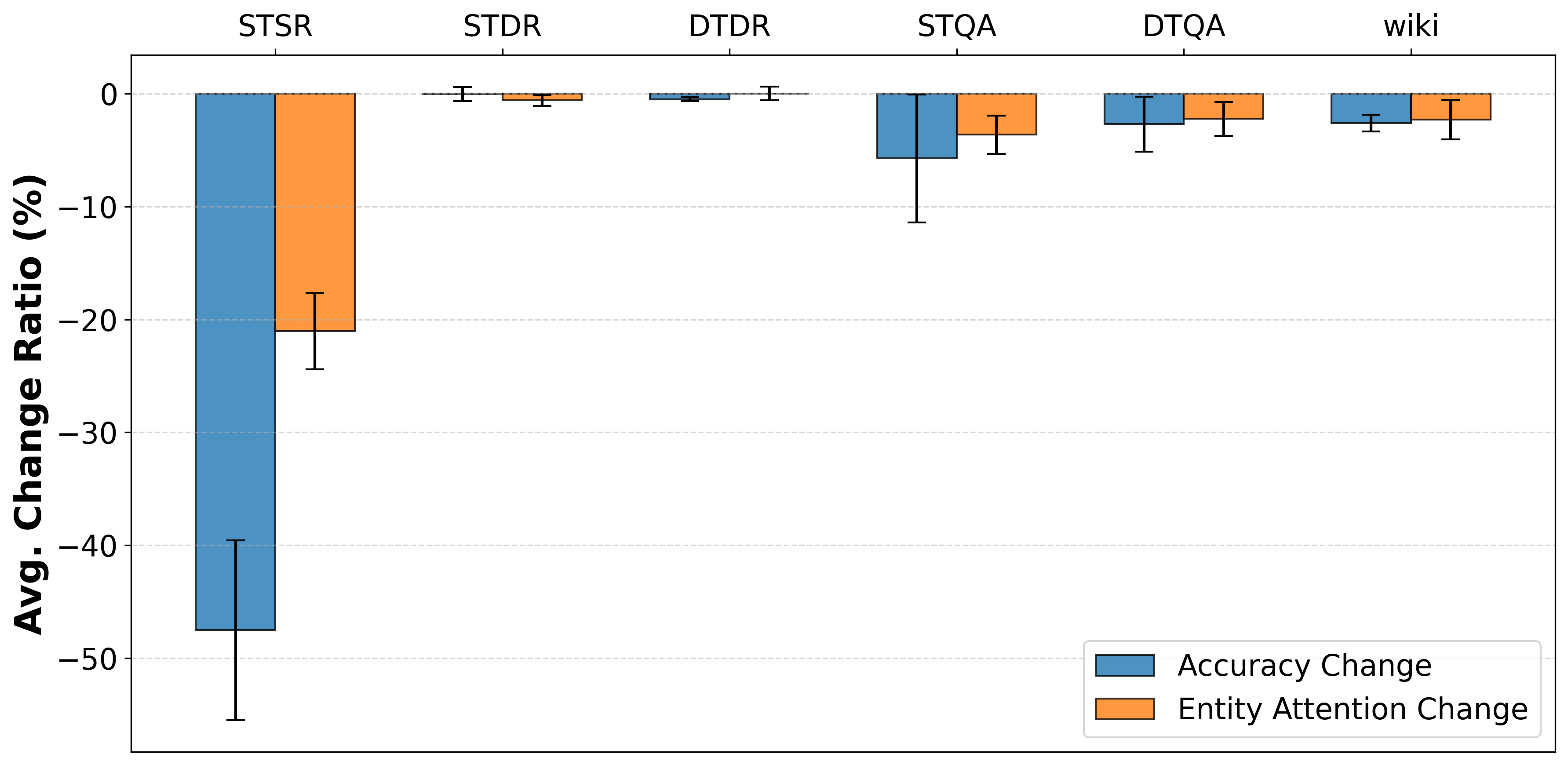}
    \caption{Accuracy and attention score changes when learning new knowledge in reasoning tasks.}
    \label{fig:reasoning_interpretability_bar}
\end{figure}

\textbf{Hallucinations correlate with declines in entity attention.} Figure~\ref{fig:qa_unknown_percentage_interpretability_lines} presents the interpretability analysis corresponding to Figure~\ref{fig:qa_unknown_percentage}. As the proportion of unknown instances within a knowledge type increases, the model’s attention to key entities gradually declines, accompanied by more severe hallucinations. Compared to \textit{KeepKnown}, \textit{RemoveKnown} exhibits a sharper attention decline and performance drop, indicating that the absence of known information accelerates attention decay and exacerbates performance degradation. Figure~\ref{fig:reasoning_interpretability_bar} shows the interpretability analysis for the reasoning tasks corresponding to Figure~\ref{fig:reasoning_big_table}, where attention patterns are also correlated with hallucinations.

Similar experiments are also conducted on the real-world ENTITYQUESTIONS dataset. The results are reported in Appendix \ref{app:wiki_results}.

We further control attention during new knowledge learning and examine the resulting changes in hallucination severity. We add a KL divergence loss in addition to the standard cross-entropy loss\footnote{$L = L_{\text{CE}} + \alpha L_{\text{KL}}$, where $\alpha = 25$, imposing a relatively strong constraint that keeps the changes in the attention module’s outputs minimal.} to enforce consistency between the attention outputs of the model before and after training across all attention layers. All other training details are in Appendix \ref{app:training_details}.

\begin{figure}[htb]
    \centering
    \includegraphics[width=0.48\textwidth]{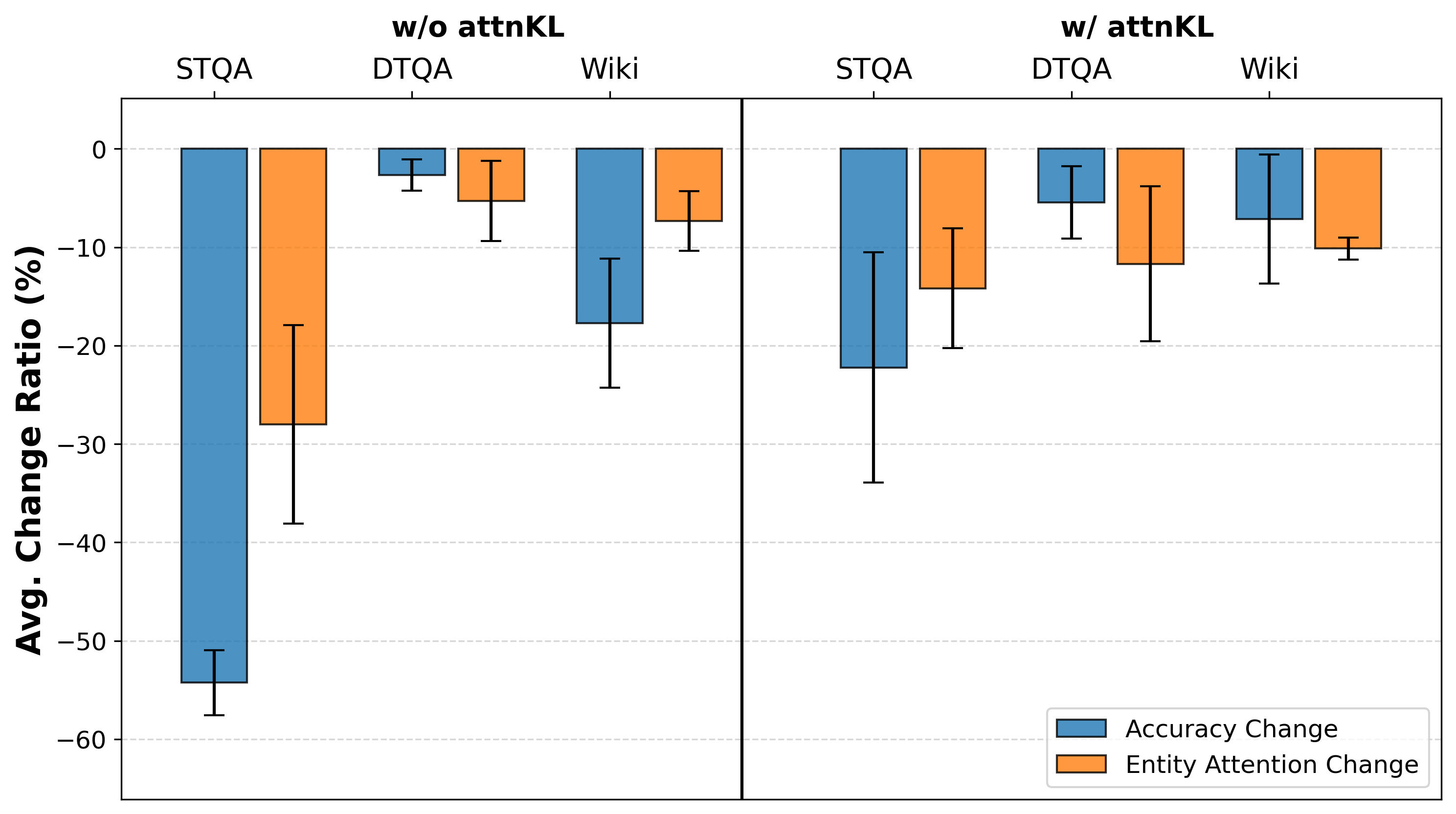}
    \caption{Results of adding KL loss. “Acc.” and “Attn.” denote the averaged performance drop and attention drop on key entities (\%, mean with standard deviation), respectively. The \emph{baseline model} is the same as Section \ref{sec:qa_results}.}
    \label{tab:KLattention}
\end{figure}

\textbf{Constraining attention alleviates factual hallucinations.} Figure \ref{tab:KLattention} presents a comparison with Table \ref{tab:qa_results} after introducing the KL constraint. We observe that, on the STQA test sets whose knowledge types are consistent with the unknown knowledge in the training data, as well as on the OOD Wiki test set, hallucinations are substantially alleviated.
However, we also find that for DTQA and the wiki test set, attention on key entities dropped more even after applying the KL constraint. To further investigate this phenomenon, we apply the same KL constraint to train another \emph{baseline model} (trained entirely on known knowledge). This results in an average accuracy drop of 3.67\% and a 9.21\% reduction in attention to key entities on the test sets, compared to the normally trained \emph{baseline model}.
\textbf{This indicates that training on known knowledge naturally encourages increased attention to key entities}, which we believe is beneficial for learning the task. Consequently, forcing attention to remain unchanged inevitably suppresses this, leading to a modest reduction in entity attention and a corresponding loss in accuracy. Overall, these results suggest that attention shifts during new knowledge training are a key contributing factor to the emergence of hallucinations.

\subsection{KnownPatch: Late-stage Injection of Known Knowledge}

Building on the above analysis, which shows that training on known knowledge promotes increased attention to key entities, we further explore whether injecting a small amount of \emph{known} knowledge at the \emph{final stage} of training can help alleviate factual hallucinations induced by learning new knowledge. We refer to this training method as \textbf{KnownPatch}. The underlying intuition is that exposure to unfamiliar knowledge can disrupt the model’s attention patterns, whereas re-introducing known knowledge encourages the restoration of entity-centric attention and leads to more stable model behavior.

The training data used before applying the fully known injection patch consist entirely of unknown knowledge across all types, simulating a worst-case scenario. We define the injection ratio as the proportion of injected known samples relative to the total training data, and we experiment with injection ratios of 5\%, 10\%, and 20\%. The \emph{baseline model} is trained on known knowledge of the full training data size, serving as an upper bound. To control for the effect of training order, we additionally compare against a model trained on the same mixed data after shuffling. Our analysis focuses on the relative performance and attention changes of the models trained under the KnownPatch and Shuffled settings compared to the \emph{baseline model}. 

\begin{figure}[htb]
    \centering
    \includegraphics[width=0.48\textwidth]{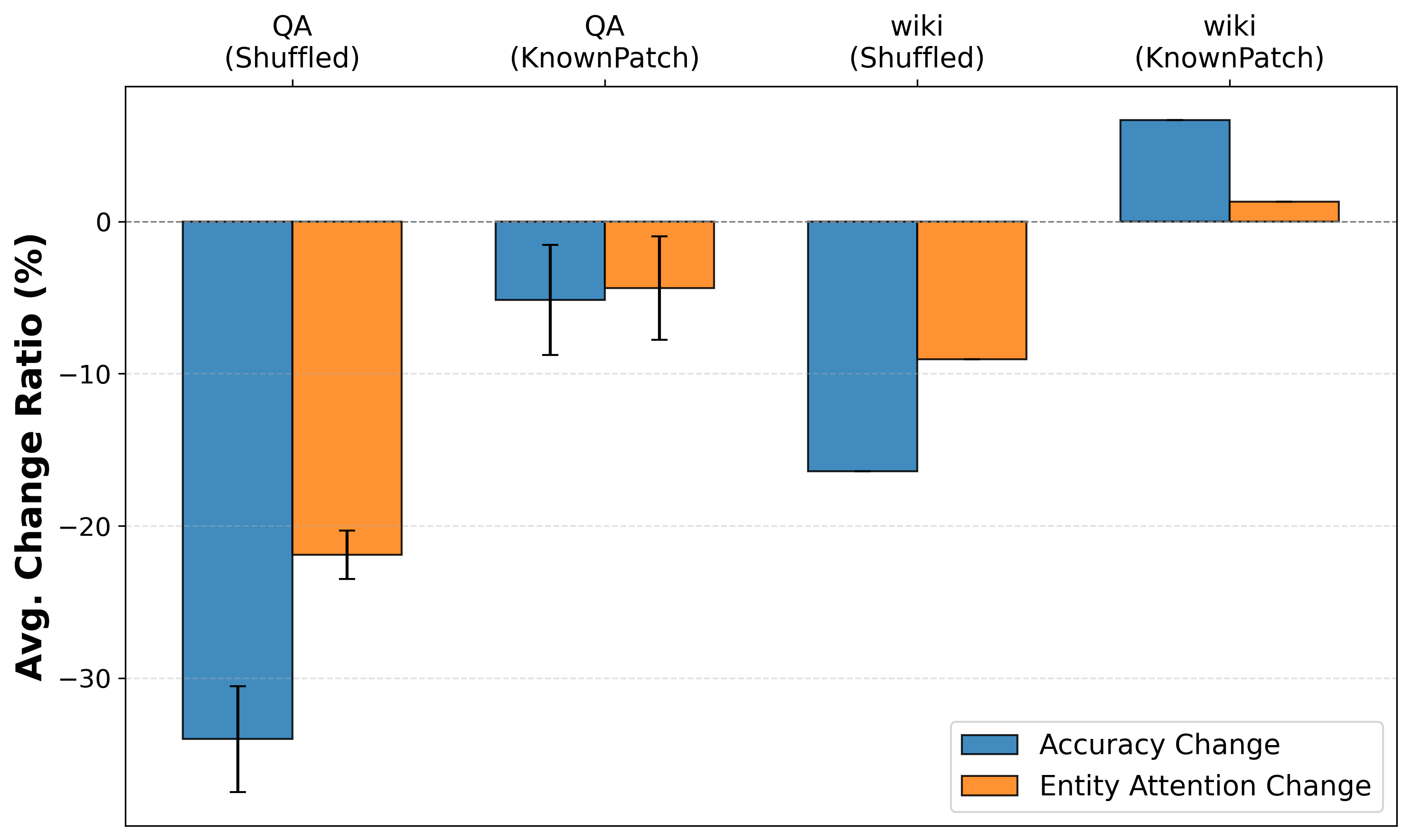}
    \caption{Performance and attention score changes under Shuffled and KnownPatch (with 20\% injection ratio) settings. QA represents the average across the four QA test sets, and error bars indicate standard deviations.}
    \label{fig:qa_mitigating_interpretability_bar}
\end{figure}

\textbf{KnownPatch stabilizes the model’s attention patterns and mitigates hallucinations.}
Figure~\ref{fig:qa_mitigating_interpretability_bar} reports the performance drop and the change in entity attention on the QA test sets under a 20\% injection ratio.  Across both in-domain QA and the OOD wiki set, KnownPatch consistently outperforms the shuffled baseline.
With 20\% injection, QA performance approaches the all-known upper bound, and performance on the OOD Wiki set even slightly exceeds it. Results under different injection ratios and reasoning tasks are provided in Appendix~\ref{app:KnownPatch_results}, 
and we find that even with only 5\% injection, KnownPatch already yields substantially higher performance than the Shuffled setting on both QA and Wiki test sets. 
These gains are accompanied by increased attention to key entities in the questions, suggesting that KnownPatch restores entity-centric attention patterns disrupted by training on unknown knowledge and yields a robust mitigation effect that generalizes to OOD data.

In Appendix \ref{app:KnownPatch_results}, we report the performance when the injected known knowledge in KnownPatch does not cover all unknown knowledge types. We observe that even for the uncovered knowledge types, factual hallucinations are still effectively mitigated. This suggests that KnownPatch is not merely performing knowledge replay, but instead alleviates hallucinations by reshaping the model’s key entity-centered attention patterns.

\subsection{Mechanism of Hallucination Propagation}

We further aim to investigate how new-knowledge-induced factual hallucinations propagate across different tasks.

\begin{table}[hb]
    \centering
    \footnotesize
    \begin{tabular}{c c c c c c c}
        \toprule
        STSR & STDR & DTDR & STQA & DTQA & wiki \\
        \midrule
        1.00 & 0.62 & 0.59 & 0.73 & 0.70 & 0.72 \\
        \bottomrule
    \end{tabular}
    \caption{Averaged token-level overlap between STSR and other test groups. Each value represents the mean token overlap ratio, defined as the proportion of tokens in a task context that also appear in the STSR context, averaged over all tasks in the corresponding group.}
    \label{tab:context_similarity}
\end{table}

Since attention weights form a normalized distribution that sums to one across all input tokens, a reduction in attention to key entities implies a corresponding shift of attention toward the remaining contextual tokens. 
Figure~\ref{fig:reasoning_big_table} shows that learning new knowledge in reasoning tasks has a more substantial impact on QA test sets than on other reasoning test sets. This is consistent with the lexical similarity (measured by token-level overlap) analysis in Table~\ref{tab:context_similarity}: QA test sets are generally more lexically similar to the STSR tasks that trained with unknown knowledge, and this higher lexical similarity aligns with the observed trend of stronger hallucination propagation. This can be attributed to the fact that reasoning tasks typically involve relatively long and diverse input contexts, making them lexically less similar to one another, whereas many knowledge QA contexts share substrings with reasoning trajectories.

\begin{table}[htb]
    \centering
    \footnotesize
    \setlength{\tabcolsep}{4pt}
    \begin{tabular}{l | c | c c | c c}
        \toprule
        \textbf{Lexical}
        & \textbf{Same}
        & \multicolumn{2}{c|}{\textbf{Similar}}
        & \multicolumn{2}{c}{\textbf{Different}} \\
        \midrule
        \textbf{Semantic}
        & \textbf{Same}
        & Same & Diff.
        & Same & Diff. \\
        \midrule
        Token Overlap    & 1.00 & 0.97 & 0.89 & 0.62 & 0.52 \\
        Hallucination & 25.57 & 18.80 & 16.61 & 6.65 & 5.90 \\
        \bottomrule
    \end{tabular}
    \caption{Averaged contextual similarity (token overlap) and performance drop (hallucination, \%).}
    \label{tab:context_similarity_extend}
\end{table}

To disentangle the effects of lexical similarity and semantic similarity in contexts on hallucination propagation, we construct extended variants of the reasoning tasks. Specifically, for each of the 12 reasoning tasks in \textit{Biography-Reasoning}, we construct four variant tasks under the same knowledge type, defined by lexical similarity (similar vs.\ different) and semantic similarity (same vs.\ different). An example for each variant task is provided in Appendix \ref{app:dataset_details}. We then examine, across the four variant tasks, the extent of performance degradation when the Origin task is trained on unknown knowledge compared to training on known knowledge. Results in Table~\ref{tab:context_similarity_extend} show that learning new knowledge induces stronger hallucinations on tasks with higher lexical similarity. This confirms that \textbf{hallucinations propagate primarily through lexical similarity rather than semantic relatedness}.


\section{Conclusion}

In this work, we present a systematic study on new-knowledge-induced factual hallucinations in LLMs, examining their behavior across knowledge types and task types. Our experiments reveal that even a small number of fully unknown facts can trigger severe hallucinations and can propagate to other tasks. Our analysis reveals that this behavior is closely associated with shifts in attention: learning new knowledge reduces attention to key entities, whereas training on known knowledge reinforces entity-centric attention. Motivated by this observation, we explore a simple training intervention, KnownPatch, which injects a small amount of known knowledge at the late stage of training to restore attention patterns and mitigate hallucinations. 
Finally, we show that the extent of hallucination propagation increases with lexical similarity between contexts, rather than semantic relatedness, highlighting contextual similarity as a key driver of cross-task hallucination transfer.

\section*{Limitations}
While our work offers a in-depth analysis of new-knowledge-induced factual hallucinations, there are several boundaries to our current study.

\paragraph{Regarding the synthetic data:} To strictly disentangle new knowledge from facts already internalized during pre-training, we constructed the synthetic Biography-Reasoning dataset. While this controlled environment was necessary, it may not fully capture the noise and semantic complexity of naturally occurring text. We partially mitigated this by validating our findings on the real-world ENTITYQUESTIONS dataset, yet exploring more complex linguistic structures remains a direction for future work.

\paragraph{Regarding model scale:} Our analysis primarily relies on open-source models (e.g., Qwen2.5-1.5B) to enable detailed and fine-grained investigation. While we verify the key findings on larger variants (e.g., Qwen3-8B, Qwen2.5-32B) in the appendix, we acknowledge that extending our experiments to extremely large-scale models (e.g., 70B+ or even larger models) is left for future work due to computational constraints.

\section*{Acknowledgements}
We would like to thank the anonymous reviewers for their insightful comments. Shujian Huang is the corresponding author. This work is supported by National Science Foundation of China (No. 62376116), research project of Nanjing University-China Mobile Joint Institute (NJ20250038), the Fundamental Research Funds for the Central Universities (No. 2024300507), Fundamental and Interdisciplinary Disciplines Breakthrough Plan of the Ministry of Education of China (No. JYB2025XDXM118).

\bibliography{ref}

\clearpage
\appendix
\renewcommand{\topfraction}{0.95}
\renewcommand{\bottomfraction}{0.95}
\renewcommand{\textfraction}{0.05}
\renewcommand{\floatpagefraction}{0.9}
\setcounter{topnumber}{5}
\setcounter{bottomnumber}{5}
\setcounter{totalnumber}{10}

\section{Dataset Details}
\label{app:dataset_details}
\subsection{Wiki Details}
\label{app:wikidetails}
The ENTITYQUESTIONS dataset from Wikidata is divided into multiple subsets, such as P17, P20, etc., with each subset containing questions of the same format. For example, an instance from P17 is ``Which country is Juniper Bank located in?", and an instance from P20 is ``Where did Connee Boswell die?". Based on \citet{gekhman2024does}'s classification of knowledge, we categorize Wikipedia knowledge into four levels: HighlyKnown, MaybeKnown, WeaklyKnown, and Unknown. We construct the test set using subsets of HighlyKnown and MaybeKnown instances. To ensure the balanced distribution of the test set, we sample approximately the same number of questions from each subset, resulting in a final test set of 1,000 questions.

\subsection{\textit{Biography-Reasoning} Details}
Each individual in the dataset is assigned four attributes: birth year, death year, major, and university. The dataset contains 3,000 individuals in total. Among them, 1,000 are kept as the \textbf{unknown} subset, while the remaining 2,000 individuals are trained during a CPT stage. Within the CPT subset, 1,000 individuals are reserved for building the test sets, and the other 1,000 are used as \textbf{known} knowledge to construct the training data. The detailed procedures for constructing names, attributes, and reasoning tasks are described below.

\paragraph{Names} 
The first name and last name of each individual are selected from separate pools and are ensured to be unique. For first names, we use 3,000 English names 
from the UCI Machine Learning Repository dataset\footnote{\url{https://archive.ics.uci.edu/dataset/591/gender+by+name}, which is under CC BY 4.0 license and could be used for any purpose.}, 
with an equal split between male and female names (this affects the use of gendered pronouns in reasoning tasks). For last names, we select 250 Chinese surnames 
from a GitHub repository\footnote{\url{https://github.com/smashew/NameDatabases/}, which is under `The Unlicense' that allows anyone to use for free.}
, which are then randomly paired with the first names in a balanced manner. This random combination of English first names and Chinese last names is designed to generate synthetic individuals that minimize overlap with real-world knowledge already known to language models.

\paragraph{Attributes}
The birth year of each synthetic individual is a random integer between 1800 and 1980. The death year is randomly assigned within the range of $birth\_year + 30$ to $min(2020, birth\_year + 100)$, ensuring realistic lifespans. The major and university attributes are based on real-world entities. There are 50 universities in total, distributed across 10 countries (5 universities per country). There are also 50 majors, categorized into 10 broad fields (5 majors per field), e.g., Computer Science → Engineering.

We refer to the four attributes Birth, Death, Major and University as B, D, M and U, respectively.

\paragraph{CPT Data}
The CPT data are mainly constructed in the form of biography texts. Here is an example of a biography:
\begin{quote}
\begin{footnotesize}
Hannalee Sui was registered as born in 1974. Hannalee Sui brought her life to a close in 2015. Hannalee Sui participated in Accounting-related research. Hannalee Sui was officially registered at University of Alberta.
\end{footnotesize}
\end{quote}
For the biographies used to construct \textbf{known} knowledge, each biography is rephrased 50 times to ensure consistent exposure. For those used to construct the test set, the biographies are divided into 10 subgroups, each rephrased 5, 10, …, up to 50 times, respectively. This design simulates a more realistic and diverse distribution of knowledge familiarity, reflecting varying degrees of knowledge internalization in practice.

\paragraph{Auxiliary Knowledge}
To construct knowledge reasoning tasks, we introduce a set of auxiliary knowledge. Specifically, our dataset involves relations such as major → field (e.g., Computer Science belongs to Engineering) and university → country (e.g., Stanford University belongs to the United States). These auxiliary facts already exist in the model’s pre-trained knowledge base. To ensure that the model reliably retains them, we also rephrase each auxiliary fact 50 times and include them in the CPT data. All auxiliary knowledge is provided in Tables~\ref{tab:major2category} and~\ref{tab:university2country}.

\paragraph{Reasoning Tasks}
For each attribute of a synthetic individual, we construct three types of reasoning questions. We refer to Single Reasoning, Comparative Reasoning, and Novel Reasoning as SR, CR, and NR, respectively. In Table \ref{tab:dataset_examples}, we provide an example for each category of QA and reasoning questions in the dataset.

Each CR task involves two attributes: the primary attribute of interest and another randomly selected one. For major- and university-related CR tasks, which take a binary (Yes/No) form, we further constrain the sampling process to maintain an approximately balanced ratio of positive and negative instances (50\% each).


To investigate the effect of context similarity on hallucination propagation, we construct variant tasks for each reasoning problem under the same knowledge type: (a) Same Meaning, Similar Lexical (SMSL); (b) Same Meaning, Different Lexical (SMDL); and (c) Different Meaning, Similar Lexical (DMSL). Table~\ref{tab:variant_reasoning_for_context_similarity} presents representative examples of these variants. For the Different Meaning, Different Lexical setting, we directly adopt the STDR tasks defined in Section \ref{sec:reasoning_results}.

\begin{table*}[ht]
    \centering
    \footnotesize
    \begin{tabularx}{\linewidth}{lX}    
    \toprule
    \textbf{Category} & \textbf{Example} \\
    \midrule
    B\_QA & Question: When was Darreus Hsiao born?\newline
Answer: 1974 \\ \midrule
    D\_QA & Question: When did Darreus Hsiao die?\newline
Answer: 2017 \\ \midrule
    M\_QA & Question: What major did Darreus Hsiao study?\newline
Answer: Dentistry \\ \midrule
    U\_QA & Question: Which university did Darreus Hsiao graduate from?\newline
Answer: Zhejiang University \\ \midrule
    B\_SR & Question: Is the number of Darreus Hsiao's birth year an odd number?\newline
Answer: Darreus Hsiao was born in 1974. 1974 \% 2 = 0. So 1974 is not an odd number.
The answer is: NO \\ \midrule
    B\_CR & Question: How many years apart is the birth year between Darreus Hsiao and Aydn Cheung?\newline
Answer: Darreus Hsiao was born in 1974. Aydn Cheung was born in 1858. The difference is abs(1974 - 1858) = 116.\newline
The answer is: 116 \\ \midrule
    B\_NR & Question: What is the MScore of Darreus Hsiao's birth year?\newline
Answer: Darreus Hsiao was born in 1974. The four numbers are 1, 9, 7 and 4. So the MScore of it is 1 * 9 * 7 * 4 = 252.\newline
The answer is: 252 \\ \midrule
    D\_SR & Question: What year is the 10th anniversary of Darreus Hsiao's death?\newline
Answer: Darreus Hsiao died in 2017. 10 years after it should be 2017 + 10 = 2027.\newline
The answer is: 2027 \\ \midrule
    D\_CR & Question: Who died first, Darreus Hsiao or Aydn Cheung?\newline
Answer: Darreus Hsiao died in 2017. Aydn Cheung died in 1919. 1919 is earlier than 2017. So Aydn Cheung died first.\newline
The answer is: Aydn Cheung \\ \midrule
    D\_NR & Question: What is the AScore of Darreus Hsiao's death year?\newline
Answer: Darreus Hsiao died in 2017. The four numbers are 2, 0, 1 and 7. So the AScore of it is 2 + 0 + 1 + 7 = 10.\newline
The answer is: 10 \\ \midrule
    M\_SR & Question: What field does Darreus Hsiao's major belong to?\newline
Answer: Darreus Hsiao's major is Dentistry. Dentistry belongs to Medicine.\newline
The answer is: Medicine \\ \midrule
    M\_CR & Question: Do Darreus Hsiao and Virgus Hong's majors belong to the same field?\newline
Answer: Darreus Hsiao's major is Dentistry. Dentistry belongs to Medicine. Virgus Hong's major is Nursing. Nursing belongs to Medicine. Medicine and Medicine are the same.\newline
The answer is: YES \\ \midrule
    M\_NR & Question: What is the sequence of odd-positioned letters in the first word of Darreus Hsiao's major name?\newline
Answer: Darreus Hsiao's major is Dentistry. The first word of `Dentistry' is `Dentistry'. The spelling of Dentistry is D, E, N, T, I, S, T, R, Y. The sequence of odd-positioned letters in `Dentistry' is DNITY.\newline
The answer is: DNITY \\ \midrule
    U\_SR & Question: In which country did Darreus Hsiao attend university?\newline
Answer: Darreus Hsiao was graduated from Zhejiang University. Zhejiang University is located in China.\newline
The answer is: China \\ \midrule
    U\_CR & Question: Are Darreus Hsiao and Angee Fung college alumni?\newline
Answer: Darreus Hsiao was graduated from Zhejiang University. Saritha Tong was graduated from Kyoto University. Zhejiang University and Kyoto University are not the same.\newline
The answer is: NO \\ \midrule
    U\_NR & Question: What is the sequence of the first and last letters of each word in Darreus Hsiao's university name?\newline
Answer: Darreus Hsiao was graduated from Zhejiang University, which can be splitted into words: Zhejiang, University. The first and last letters of `Zhejiang' are ZG. The first and last letters of `University' are UY. So, the whole sequence is ZGUY.\newline
The answer is: ZGUY \\

    \bottomrule
    \end{tabularx}
    \caption{Examples of each QA and reasoning tasks in \textit{Biography-Reasoning}. B, D, M, and U denote birth year, death year, major, and university, respectively; 
SR, CR, and NR denote Single Reasoning, Comparative Reasoning, and Novel Reasoning, respectively.}
    \label{tab:dataset_examples}
\end{table*}

\begin{table*}[ht]
    \centering
    \footnotesize
    \begin{tabularx}{\linewidth}{lX}    
    \toprule
    \textbf{Category} & \textbf{Example} \\
    \midrule
    B\_SR & SMSL: Is Hakam Cheng's birth year an odd number?\newline
SMDL: Can Hakam Cheng's year of birth, when considered as an integer, be classified under the category of odd values?\newline
DMSL: Is the number of Hakam Cheng's birth year an even number?\\ \midrule
    B\_CR & SMSL: How many years is the birth year between Hakam Cheng and Graicyn Xian apart?\newline
SMDL: By what number of years are Hakam Cheng and Graicyn Xian separated in terms of their birth years?\newline
DMSL: How many months apart is the birth year between Hakam Cheng and Graicyn Xian?\\ \midrule
    B\_NR & SMSL: What's the MScore of Hakam Cheng's birth year?\newline
SMDL: Determine the specific MScore value attributed to the calendar year during which Hakam Cheng was born.\newline
DMSL: What is the AScore of Hakam Cheng's birth year?\\ \midrule
    D\_SR & SMSL: What is the 10th anniversary of Hakam Cheng's death year?\newline
SMDL: After 10 years from Hakam Cheng passed away, what year does that correspond to?\newline
DMSL: What is the year before the 10th anniversary of Hakam Cheng's death?\\ \midrule
    D\_CR & SMSL: Hakam Cheng or Graicyn Xian, who died first?\newline
SMDL: Identify which individual—Hakam Cheng or Graicyn Xian—died at an earlier date.\newline
DMSL: Who died later, Hakam Cheng or Graicyn Xian?\\ \midrule
    D\_NR & SMSL: What's the AScore of Hakam Cheng's death year?\newline
SMDL: Which AScore value corresponds to the calendar year in which Hakam Cheng died?\newline
DMSL: What is the MScore of Hakam Cheng's death year?\\ \midrule
    M\_SR & SMSL: Which field does Hakam Cheng's major belong to?\newline
SMDL: To which academic discipline can the major pursued by Hakam Cheng be categorized?\newline
DMSL: What is the first letter of the field that Hakam Cheng's major belongs to?\\ \midrule
    M\_CR & SMSL: Do Hakam Cheng's and Graicyn Xian's majors belong to the same field?\newline
SMDL: Are Hakam Cheng and Graicyn Xian's areas of academic specialization considered part of the same disciplinary category?\newline
DMSL: Do Hakam Cheng and Graicyn Xian's majors belong to different field?\\ \midrule
    M\_NR & SMSL: What is the sequence of odd-positioned letters within the first word of Hakam Cheng's major name?\newline
SMDL: From the initial word of Hakam Cheng's major, extract the characters occupying positions with odd indices and present them in order.\newline
DMSL: What is the sequence of even-positioned letters in the first word of Hakam Cheng's major name?\\ \midrule
    U\_SR & SMSL: Hakam Cheng attend university in which country?\newline
SMDL: Identify the nation within whose borders Hakam Cheng pursued university-level studies.\newline
DMSL: What is the last letter of the country in which did Hakam Cheng attended university?\\ \midrule
    U\_CR & SMSL: Hakam Cheng and Graicyn Xian are college alumni?\newline
SMDL: Have Hakam Cheng and Graicyn Xian both completed their higher education at the same university?\newline
DMSL: Are Hakam Cheng and Graicyn Xian not college alumni?\\ \midrule
    U\_NR & SMSL: What's the sequence of the first and last letters of each word in Hakam Cheng's university name?\newline
SMDL: Provide the ordered sequence formed by the initial and final characters of every word appearing in Hakam Cheng's university title.\newline
DMSL: What is the sequence of the last and first letters of each word in Hakam Cheng's university name?\\

    \bottomrule
    \end{tabularx}
    \caption{Examples of the variant tasks for reasoning tasks.}
    \label{tab:variant_reasoning_for_context_similarity}
\end{table*}

\section{Training Details}
\label{app:training_details}
In all CPT experiments, unless otherwise specified, we use a batch size of 16, a learning rate of 1e-5, a cutoff length of 512, and train for 1 epoch. 

In all SFT experiments (including knowledge QA and knowledge-based reasoning tasks), unless otherwise specified, we use a batch size of 32, a learning rate of 1e-5, and train for 3 epochs. We also did experiments of training 1 or 5 epochs, the results are presented in Appendix \ref{app:different_epochs}.

In the experiment that uses an auxiliary KL loss in Section \ref{sec:interpretability}, due to this additional constraint, training becomes more difficult, so we set the number of training epochs to at most ten and apply early stopping when the training accuracy exceeds 95\%. All other training setups are the same as above.

Most of the experiments are conducted on up to four NVIDIA A6000 GPUs for models with fewer than 8B parameters. For training models with 8B parameters or more, up to eight NVIDIA H20 GPUs are used. The CPT stage is performed using LLaMA-Factory \citep{zheng2024llamafactory}. 

\clearpage

\begin{table*}[ht]
    \centering
    \begin{tabular}{l|l}
    \toprule
    \textbf{Field} & \textbf{Major} \\
    \midrule
Economics & Finance, Investment, Taxation, Insurance, Digital Economy \\ \midrule
Law & Intellectual Property, Criminal Justice, Sociology, International Politics, Diplomacy \\ \midrule
Literature & Journalism, Advertising, English, French, Russian \\ \midrule
History & Chinese History, World History, Museum Studies, Science History, Historical Geography \\ \midrule
Science & Mathematics, Physics, Chemistry, Biology, Geology \\ \midrule
Engineering & Computer Science, Software Engineering, Automation, Architecture, Electrical Engineering \\ \midrule
Medicine & Clinical Medicine, Dentistry, Pharmacy, Nursing, Public Health \\ \midrule
Agriculture & Agronomy, Horticulture, Plant Protection, Animal Science, Forestry \\ \midrule
Management & Accounting, Finance Management, Library Science, Tourism Management, Logistics Management \\ \midrule
Art & Fine Arts, Music, Dance, Art Theory, Environmental Design \\
    \bottomrule
    \end{tabular}
    \caption{Auxiliary knowledge related to majors.}
    \label{tab:major2category}
\end{table*}

\begin{table*}[ht]
    \centering
    \begin{tabularx}{\linewidth}{lX}
\toprule
\textbf{Country} & \textbf{Universities} \\
\midrule
United States & Harvard University, Stanford University, Princeton University,\newline Yale University, Columbia University \\ \midrule
United Kingdom & University of Oxford, University of Cambridge, Imperial College London,\newline University College London, University of Manchester \\ \midrule
Canada & University of Toronto, McGill University, University of Alberta,\newline McMaster University, University of Waterloo \\ \midrule
Australia & University of Melbourne, University of Sydney, University of Queensland,\newline Monash University, Macquarie University \\ \midrule
Germany & Heidelberg University, RWTH Aachen University, University of Freiburg,\newline University of Hamburg, University of Tübingen \\ \midrule
France & Sorbonne University, University of Paris, University of Strasbourg,\newline University of Lyon, University of Bordeaux \\ \midrule
China & Tsinghua University, Peking University, Fudan University,\newline Zhejiang University, Nanjing University \\ \midrule
Japan & Kyoto University, Osaka University, Tohoku University,\newline Nagoya University, Hokkaido University \\ \midrule
Singapore & Nanyang Technological University, Singapore Management University, Temasek Polytechnic,\newline Republic Polytechnic, Singapore Polytechnic \\ \midrule
South Korea & Seoul National University, Korea University, Yonsei University,\newline Sungkyunkwan University, Hanyang University \\
\bottomrule
    \end{tabularx}
    \caption{Auxiliary knowledge related to universities.}
    \label{tab:university2country}
\end{table*}

\begin{table*}[ht]
    \centering
    \begin{tabular}{l|ccc|ccc|ccc|ccc}
    \toprule
    \textbf{Dataset} &
      \multicolumn{3}{c|}{\textbf{Birth}} &
      \multicolumn{3}{c|}{\textbf{Death}} &
      \multicolumn{3}{c|}{\textbf{Major}} &
      \multicolumn{3}{c}{\textbf{University}} \\
    \cmidrule(lr){2-4} \cmidrule(lr){5-7} \cmidrule(lr){8-10} \cmidrule(lr){11-13}
     & SR & CR & NR & SR & CR & NR & SR & CR & NR & SR & CR & NR \\
    \midrule
    $\text{All}_\text{k}$ & 0.777 & 0.335 & 0.611 & 0.677 & 0.914 & 0.710 & 0.773 & 0.776 & 0.707 & 0.724 & 0.777 & 0.653 \\
    $\text{B\_SR}_\text{unk}$ & \textbf{0.643} & 0.321 & 0.589 & 0.665 & 0.908 & 0.707 & 0.777 & 0.784 & 0.688 & 0.728 & 0.777 & 0.636 \\
    $\text{B\_CR}_\text{unk}$ & 0.797 & \textbf{0.088} & 0.618 & 0.663 & 0.913 & 0.694 & 0.786 & 0.788 & 0.695 & 0.718 & 0.768 & 0.635   \\
    $\text{B\_NR}_\text{unk}$ & 0.781 & 0.329 & \textbf{0.367} & 0.663 & 0.913 & 0.694 & 0.780 & 0.794 & 0.702 & 0.726 & 0.774 & 0.647 \\
    $\text{D\_SR}_\text{unk}$ & 0.785 & 0.331 & 0.609 & \textbf{0.091} & 0.909 & 0.692 & 0.785 & 0.793 & 0.709 & 0.723 & 0.772 & 0.649 \\
    $\text{D\_CR}_\text{unk}$ & 0.781 & 0.320 & 0.592 & 0.656 & \textbf{0.849} & 0.691 & 0.772 & 0.787 & 0.701 & 0.718 & 0.760 & 0.645  \\
    $\text{D\_NR}_\text{unk}$ & 0.779 & 0.327 & 0.598 & 0.657 & 0.904 & \textbf{0.330} & 0.785 & 0.790 & 0.707 & 0.726 & 0.772 & 0.633 \\
    $\text{M\_SR}_\text{unk}$ & 0.773 & 0.342 & 0.601 & 0.671 & 0.912 & 0.701 & \textbf{0.603} & 0.799 & 0.698 & 0.722 & 0.766 & 0.637 \\
    $\text{M\_CR}_\text{unk}$ & 0.769 & 0.324 & 0.606 & 0.667 & 0.915 & 0.703 & 0.793 & \textbf{0.573} & 0.722 & 0.730 & 0.765 & 0.607  \\
    $\text{M\_NR}_\text{unk}$ & 0.788 & 0.332 & 0.614 & 0.664 & 0.914 & 0.709 & 0.790 & 0.797 & \textbf{0.141} & 0.725 & 0.766 & 0.631  \\
    $\text{U\_SR}_\text{unk}$ & 0.798 & 0.330 & 0.616 & 0.670 & 0.905 & 0.707 & 0.780 & 0.790 & 0.703 & \textbf{0.288} & 0.782 & 0.650   \\
    $\text{U\_CR}_\text{unk}$ & 0.789 & 0.329 & 0.611 & 0.663 & 0.915 & 0.704 & 0.784 & 0.796 & 0.706 & 0.742 & \textbf{0.562} & 0.663  \\
    $\text{U\_NR}_\text{unk}$ & 0.793 & 0.344 & 0.618 & 0.669 & 0.908 & 0.701 & 0.781 & 0.784 & 0.701 & 0.731 & 0.784 & \textbf{0.156} \\
    \bottomrule
    \end{tabular}
    \caption{Impact on other reasoning test sets when training new knowledge in reasoning tasks.}
    \label{tab:reasoning2reasoning}
\end{table*}

\begin{table*}[ht]
    \centering
    \label{tab:aggregated_metrics_simplified}
    \begin{tabular}{l|ccccc}
    \toprule
    \textbf{Dataset} & \textbf{B\_QA} & \textbf{D\_QA} & \textbf{M\_QA} & \textbf{U\_QA} & \textbf{wiki} \\
    \midrule
    $\text{All}_{\text{k}}$ & 0.578 & 0.665 & 0.297 & 0.673 & 0.286 \\
    $\text{B\_SR}_{\text{unk}}$ & 0.562 & 0.651 & 0.316 & 0.668 & 0.289 \\
    $\text{B\_CR}_{\text{unk}}$ & 0.581 & 0.639 & 0.328 & 0.684 & 0.275 \\
    $\text{B\_NR}_{\text{unk}}$ & 0.568 & 0.616 & 0.165 & 0.671 & 0.279 \\
    $\text{D\_SR}_{\text{unk}}$ & 0.569 & 0.669 & 0.279 & 0.670 & 0.274 \\
    $\text{D\_CR}_{\text{unk}}$ & 0.552 & 0.627 & 0.293 & 0.670 & 0.273 \\
    $\text{D\_NR}_{\text{unk}}$ & 0.563 & 0.658 & 0.318 & 0.681 & 0.283 \\
    $\text{M\_SR}_{\text{unk}}$ & 0.573 & 0.663 & 0.319 & 0.669 & 0.282 \\
    $\text{M\_CR}_{\text{unk}}$ & 0.566 & 0.603 & 0.157 & 0.598 & 0.272 \\
    $\text{M\_NR}_{\text{unk}}$ & 0.577 & 0.545 & 0.190 & 0.655 & 0.266 \\
    $\text{U\_SR}_{\text{unk}}$ & 0.578 & 0.571 & 0.210 & 0.606 & 0.290 \\
    $\text{U\_CR}_{\text{unk}}$ & 0.564 & 0.657 & 0.355 & 0.685 & 0.276 \\
    $\text{U\_NR}_{\text{unk}}$ & 0.565 & 0.619 & 0.299 & 0.683 & 0.284 \\
    \bottomrule
    \end{tabular}
    \caption{Impact on other QA test sets when training new knowledge in reasoning tasks.}
    \label{tab:reasoning2qa}
\end{table*}
\clearpage

\section{More Results for Main Text}

\subsection{Detailed Results for Main Text}
\label{app:detailed_results}
In this section, we detail the test results of each model on each dataset as shown in Table \ref{tab:qa_results} and Figure \ref{fig:reasoning_big_table} in the main text. In all settings, $\text{All}_\text{k}$ denotes the baseline model trained on data constructed entirely from known knowledge, while $\text{X}_\text{unk}$ refers to the variant where the subset $\text{X}$ is replaced with unknown knowledge. Table \ref{tab:qa_results_detail} present the detailed results of the four variants of Table \ref{tab:qa_results}. Table \ref{tab:reasoning2reasoning} and Table \ref{tab:reasoning2qa} presents the twelve variants of reasoning tasks.

\begin{table}[htb]
    \centering
    \footnotesize
    \begin{tabular}{l c c c c c}
    \toprule
    Model & B\_QA & D\_QA & M\_QA & U\_QA & wiki \\
    \midrule
    $\text{All}_\text{k}$ & 0.549 & 0.650 & 0.519 & 0.442 & 0.199 \\
    $\text{B}_{\text{unk}}$ & \textbf{0.252} & 0.618 & 0.513 & 0.430 & 0.179 \\
    $\text{D}_{\text{unk}}$ & 0.521 & \textbf{0.322} & 0.511 & 0.426 & 0.168 \\
    $\text{M}_{\text{unk}}$ & 0.539 & 0.627 & \textbf{0.244} & 0.445 & 0.165 \\
    $\text{U}_{\text{unk}}$ & 0.528 & 0.635 & 0.508 & \textbf{0.179} & 0.143 \\
    \bottomrule
    \end{tabular}
    \caption{Hallucination induced by SFT on different unknown knowledge types.}
    \label{tab:qa_results_detail}
\end{table}

\subsection{Results on Real-World Dataset}
\label{app:wiki_results}
To examine whether our findings hold on real-world data, we additionally conduct experiments on the ENTITYQUESTIONS dataset \citep{sciavolino-etal-2021-simple}. A key difference between real-world data and our synthetic dataset lies in whether the corresponding knowledge has already been observed by the model during its initial pre-training stage.

Following the procedure described in Appendix \ref{app:wikidetails}, we categorize knowledge for the Qwen2.5-1.5B model into four groups: HighlyKnown, MaybeKnown, WeaklyKnown, and Unknown. We select four knowledge types that contain a relatively large proportion of HighlyKnown and Unknown instances: P36, P106, P159, and P407, and use them as training knowledge types. 

Following the experimental setups in Section \ref{sec:qa_results}, we sample 500 instances for each knowledge type. A model trained on the fully known dataset serves as the baseline model, while variant models are obtained by replacing one knowledge type with unknown instances. We then evaluate these models on the corresponding test sets of the four knowledge types, as well as on the same Wiki test set used in the main text. The results are shown in Table \ref{tab:wiki_qa_results}, which shows exactly the same trends as Table \ref{tab:qa_results}.

\begin{table}[htb]
    \centering
    \begin{tabular}{c c c}
    \toprule
    STQA & DTQA & wiki \\
    \midrule
    -36.94 {\footnotesize($\pm$ 11.30)} & -3.29 {\footnotesize($\pm$ 3.35)} & -6.86 {\footnotesize($\pm$ 2.57)} \\
    \bottomrule
    \end{tabular}
    \caption{Average performance degradation (\%, mean ± std) of four model variants.}
    \label{tab:wiki_qa_results}
\end{table}

\begin{figure}[htb]
    \centering
    \includegraphics[width=0.5\textwidth]{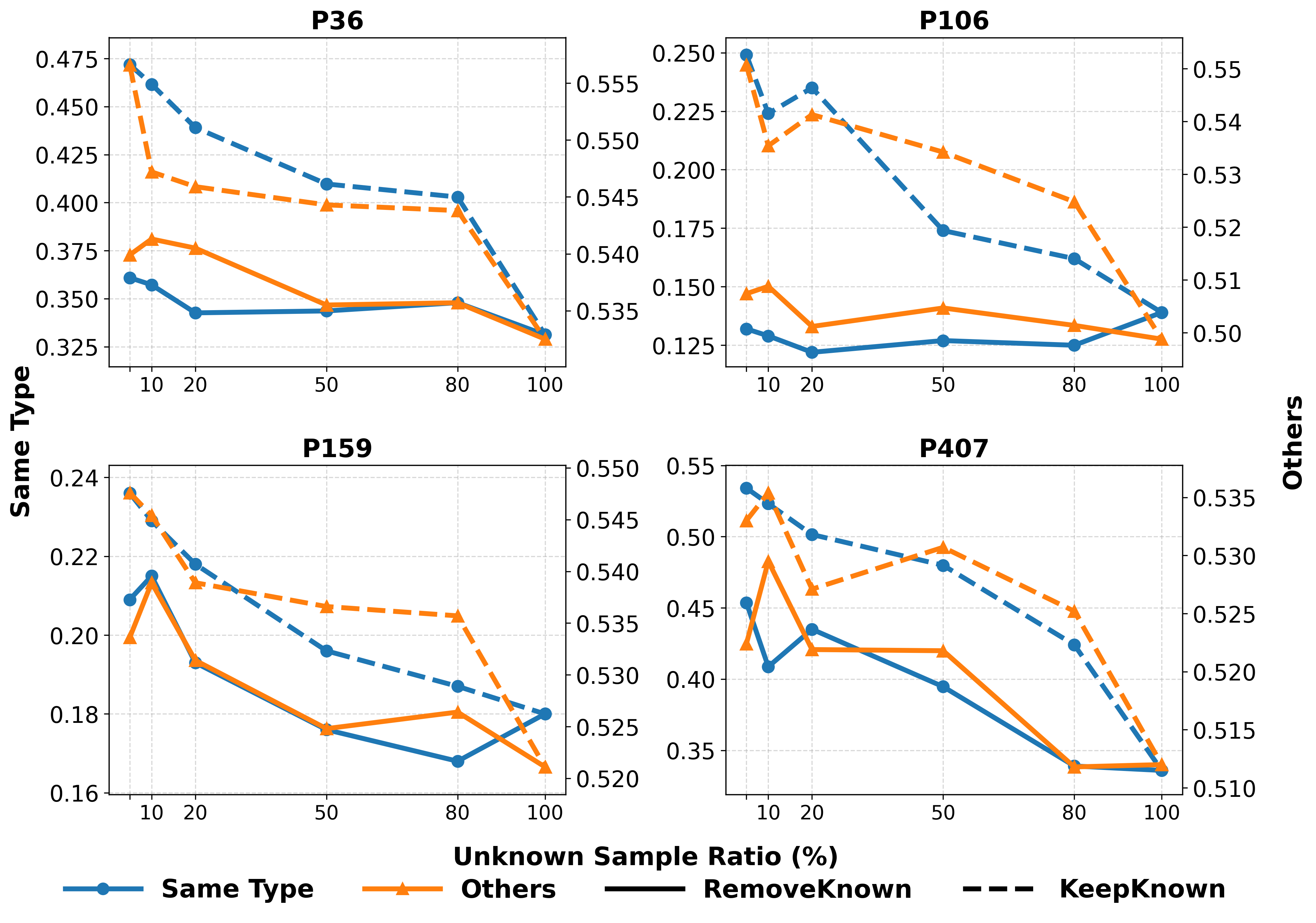}
    \caption{Performance with different proportions of unknown knowledge in the same type and Wiki test set.}
    \label{fig:wiki_qa_unknown_percentage}
\end{figure}

We conduct experiments to examine how the proportion of unknown knowledge affects the severity of hallucinations. The results are shown in Figure \ref{fig:wiki_qa_unknown_percentage}, which shows similar trends as Figure \ref{fig:qa_unknown_percentage}.

\begin{figure}[hb]
    \centering
    \includegraphics[width=0.5\textwidth]{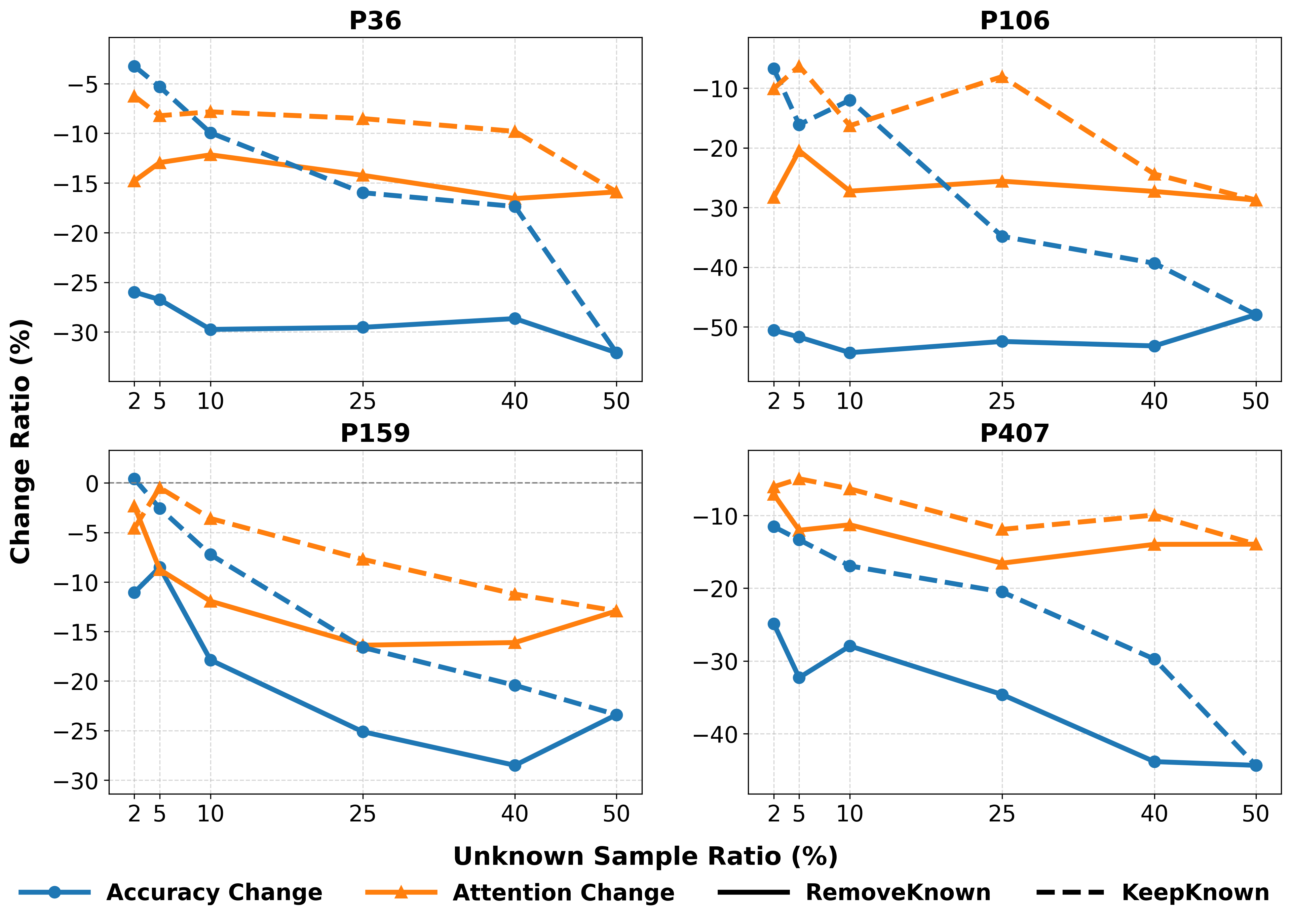}
    \caption{Accuracy and attention score changes with different unknown data ratio in certain type.}
    \label{fig:wiki_qa_unknown_percentage_interpretability_lines}
\end{figure}

Following the experiments in Section \ref{sec:interpretability}, in Figure \ref{fig:wiki_qa_unknown_percentage_interpretability_lines} we also show that the attention to key entities is correlated to performance drop when unknown knowledge proportion increases.

\begin{figure}[hb]
    \centering
    \includegraphics[width=0.5\textwidth]{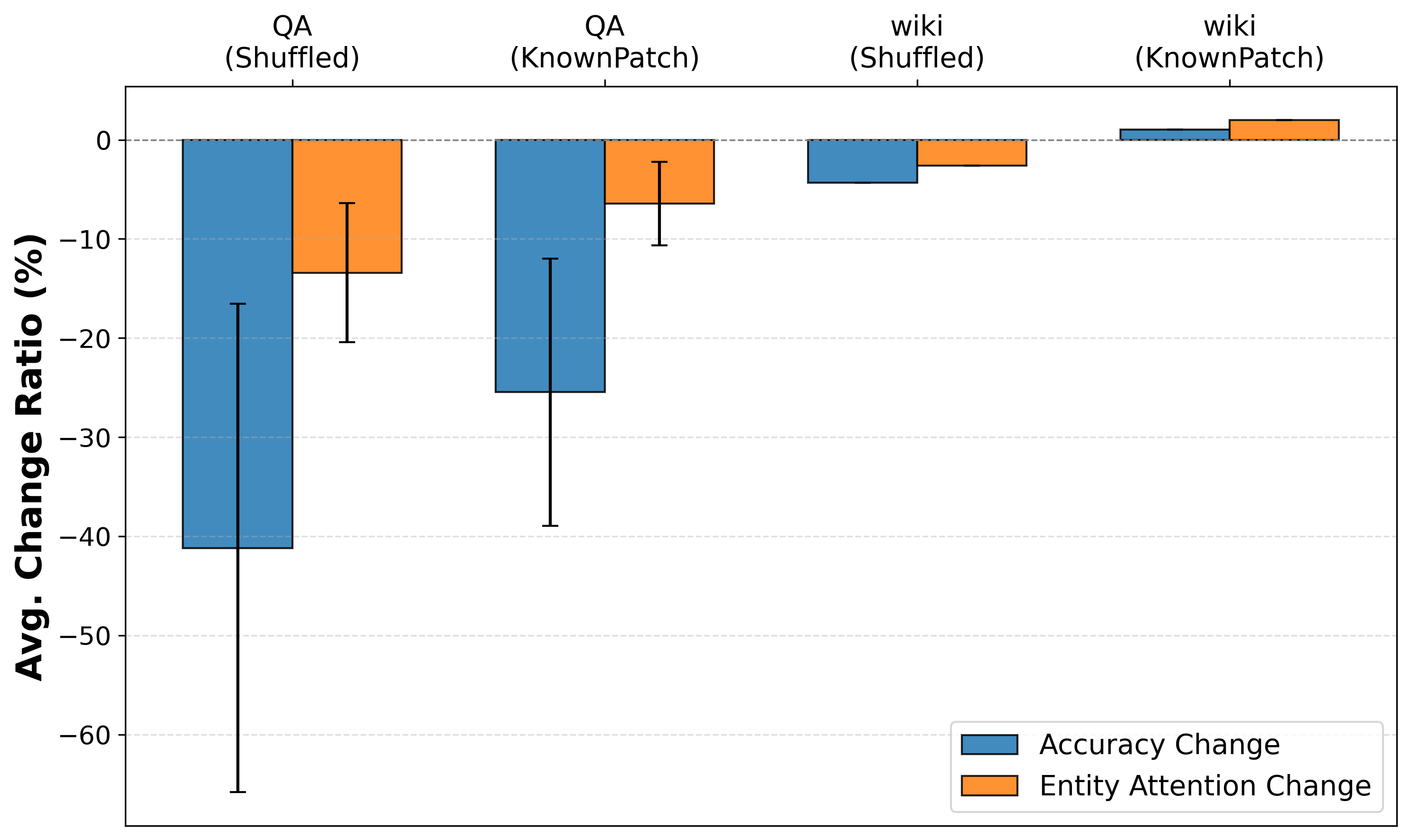}
    \caption{Performance and attention score changes under Shuffled and KnownPatch (with 20\% injection ratio) settings. QA represents the average across the four QA test sets, and error bars indicate standard deviations.}
    \label{fig:wiki_qa_mitigating}
\end{figure}

Figure \ref{fig:wiki_qa_mitigating} presents the results of KnownPatch (with injection ratio 20\%) on real-world data, along with its attention analysis.

\section{Attention Layer Selection}
\label{app:attention_layer_details}
In Figure \ref{fig:layer_selection}, the two lines represent the layer-wise entity attention patterns of two models across multiple datasets. The ``QA'' line corresponds to a model trained on all known QA questions 
(the baseline model in Figure \ref{fig:qa_mitigating_interpretability_bar})
, with attention averaged over entity tokens in five QA test sets. The ``Reasoning'' line represents a model trained on a mixture of all 12 known reasoning tasks and QA questions 
(the baseline model in Figure \ref{fig:reasoning_interpretability_bar})
, with attention averaged over entity tokens in the reasoning test sets across all reasoning types. 

\section{CPT Results}
\label{app:cpt_results}
We investigate hallucination in models during the CPT phase. Using QA questions constructed from the \textit{Biography-Reasoning} dataset, we construct CPT data concatenated via the EOS token. Note that this is the second CPT, because the first injection of known knowledge have undergone one CPT process, as described in Section \ref{sec:methodology}. Building on the experimental setup described in the Appendix \ref{app:training_details}, we conduct the following ablation studies: (1) the original experimental setting; (2) reducing the batch size from 16 to 1; (3) shortening the cutoff length from 512 to 32; and (4) increasing the total training data volume by a factor of 10. 

The models are then evaluated with 5-shot QA format. We adapt the knowledge categorization method from \citet{gekhman2024does}, with minor modifications. Specifically, we prompt the model with 5 different 5-shot samples. If the model answers correctly in at least one case, it is classified as \textbf{Known}; if all answers are incorrect, it is classified as \textbf{Unknown}. This is because the selection and order of few shots can significantly affect the model's performance \citep{lu-etal-2022-fantastically,pmlr-v139-zhao21c}, and we need to rule out this influence. The results are shown in Tables \ref{tab:qa_results_cpt_1}, \ref{tab:qa_results_cpt_2}, \ref{tab:qa_results_cpt_3} and \ref{tab:qa_results_cpt_4}.

Among all the results, except for Table \ref{tab:qa_results_cpt_1}, there are quite serious hallucination phenomena. By varying different experimental settings, we rule out all interference factors and found that the number of steps for parameter updates may be the only variable that influences the degree of hallucination when LLM learns new knowledge. Specifically, due to the limited size of our dataset, the model is updated for only a small number of steps, resulting in relatively mild hallucinations as shown in Table \ref{tab:qa_results_cpt_1}. However, no matter whether we reduce the batch size, decrease the cutoff length, or increase the data volume, as long as the number of update steps increases, the hallucination phenomenon will become more serious.

\begin{table}[!htbp]
    \centering
    \begin{tabular}{l c c c c}
    \toprule
    Model & B\_QA & D\_QA & M\_QA & U\_QA \\
    \midrule
    $\text{All}_\text{k}$ & 0.655 & 0.715 & 0.429 & 0.430 \\
    $\text{B}_\text{unk}$ & \textbf{0.560} & 0.706 & 0.326 & 0.416 \\
    $\text{D}_\text{unk}$ & 0.621 & \textbf{0.684} & 0.346 & 0.421 \\
    $\text{M}_\text{unk}$ & 0.652 & 0.712 & \textbf{0.368} & 0.422 \\
    $\text{U}_\text{unk}$ & 0.646 & 0.713 & 0.366 & \textbf{0.426} \\
    \bottomrule
    \end{tabular}
    \caption{Accuracy on test sets of models trained on different unknown knowledge types during CPT with the original setting.}
    \label{tab:qa_results_cpt_1}
\end{table}

\begin{table}[!htbp]
    \centering
    \begin{tabular}{l c c c c}
    \toprule
    Model & B\_QA & D\_QA & M\_QA & U\_QA \\
    \midrule
    $\text{All}_\text{k}$ & 0.419 & 0.477 & 0.356 & 0.417 \\
    $\text{B}_\text{unk}$ & \textbf{0.019} & 0.463 & 0.255 & 0.368 \\
    $\text{D}_\text{unk}$ & 0.402 & \textbf{0.070} & 0.292 & 0.375 \\
    $\text{M}_\text{unk}$ & 0.398 & 0.485 & \textbf{0.018} & 0.364 \\
    $\text{U}_\text{unk}$ & 0.422 & 0.505 & 0.422 & \textbf{0.084} \\
    \bottomrule
    \end{tabular}
    \caption{Accuracy on test sets of models trained on different unknown knowledge types during CPT with setting (2): batch size reduced to 1.}
    \label{tab:qa_results_cpt_2}
\end{table}

\begin{table}[!htbp]
    \centering
    \begin{tabular}{l c c c c}
    \toprule
    Model & B\_QA & D\_QA & M\_QA & U\_QA \\
    \midrule
    $\text{All}_\text{k}$ & 0.477 & 0.522 & 0.520 & 0.439 \\
    $\text{B}_\text{unk}$ & \textbf{0.081} & 0.538 & 0.523 & 0.464 \\
    $\text{D}_\text{unk}$ & 0.464 & \textbf{0.114} & 0.562 & 0.416 \\
    $\text{M}_\text{unk}$ & 0.453 & 0.539 & \textbf{0.027} & 0.407 \\
    $\text{U}_\text{unk}$ & 0.460 & 0.565 & 0.408 & \textbf{0.162} \\
    \bottomrule
    \end{tabular}
    \caption{Accuracy on test sets of models trained on different unknown knowledge types during CPT with setting (3): cutoff length reduced to 32.}
    \label{tab:qa_results_cpt_3}
\end{table}

\begin{table}[!htbp]
    \centering
    \begin{tabular}{l c c c c}
    \toprule
    Model & B\_QA & D\_QA & M\_QA & U\_QA \\
    \midrule
    $\text{All}_\text{k}$ & 0.817 & 0.843 & 0.545 & 0.664 \\
    $\text{B}_\text{unk}$ & \textbf{0.130} & 0.810 & 0.535 & 0.657 \\
    $\text{D}_\text{unk}$ & 0.792 & \textbf{0.191} & 0.589 & 0.704 \\
    $\text{M}_\text{unk}$ & 0.801 & 0.831 & \textbf{0.013} & 0.488 \\
    $\text{U}_\text{unk}$ & 0.800 & 0.828 & 0.333 & \textbf{0.014} \\
    \bottomrule
    \end{tabular}
    \caption{Accuracy on test sets of models trained on different unknown knowledge types during CPT with setting (4) dataset increased by a factor of 10.}
    \label{tab:qa_results_cpt_4}
\end{table}

\section{Supplementary Results of KnownPatch}
\label{app:KnownPatch_results}

\begin{figure}[ht]
    \centering
    \includegraphics[width=0.44\textwidth]{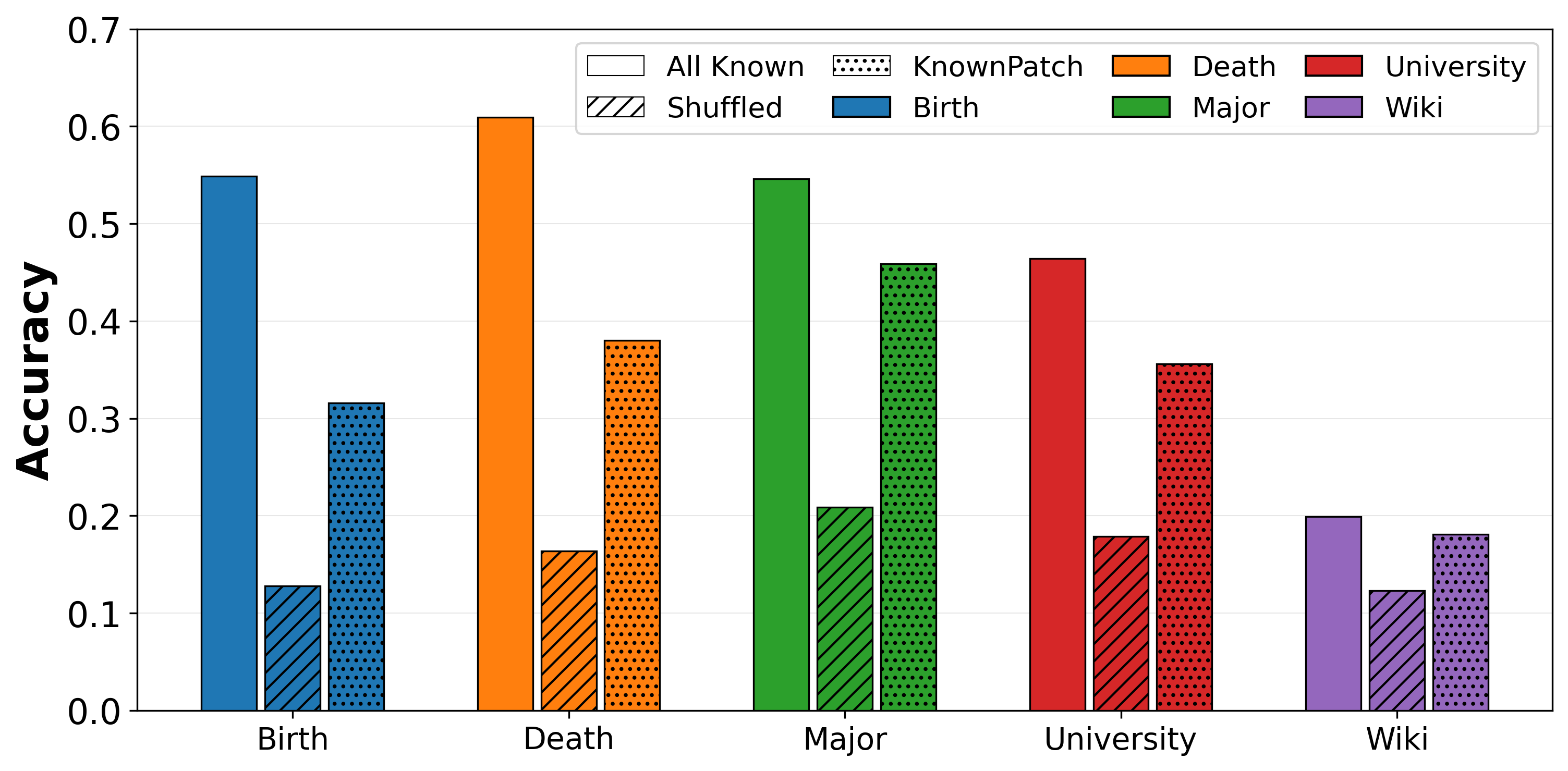}
    \includegraphics[width=0.44\textwidth]{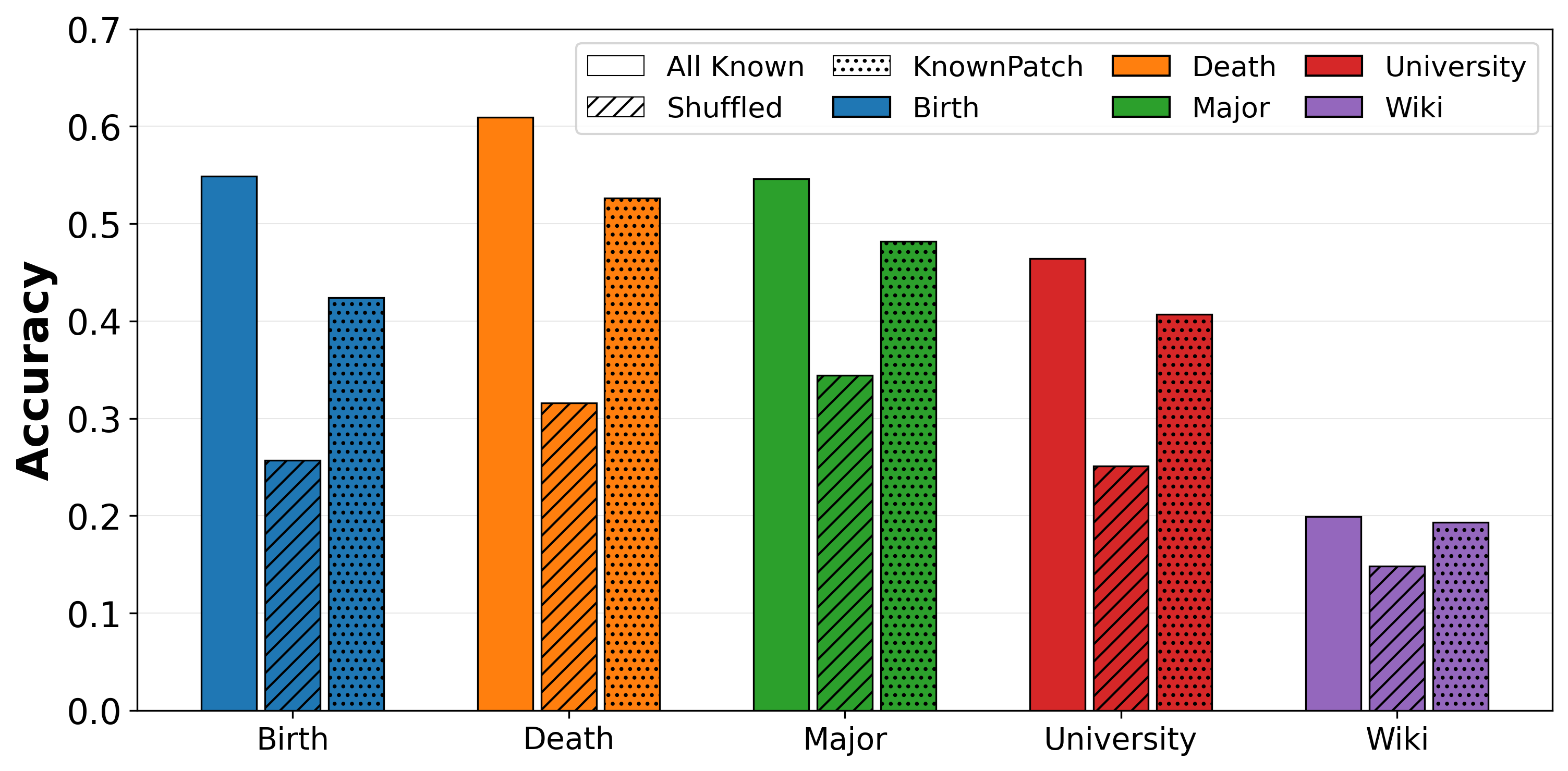}
    \includegraphics[width=0.44\textwidth]{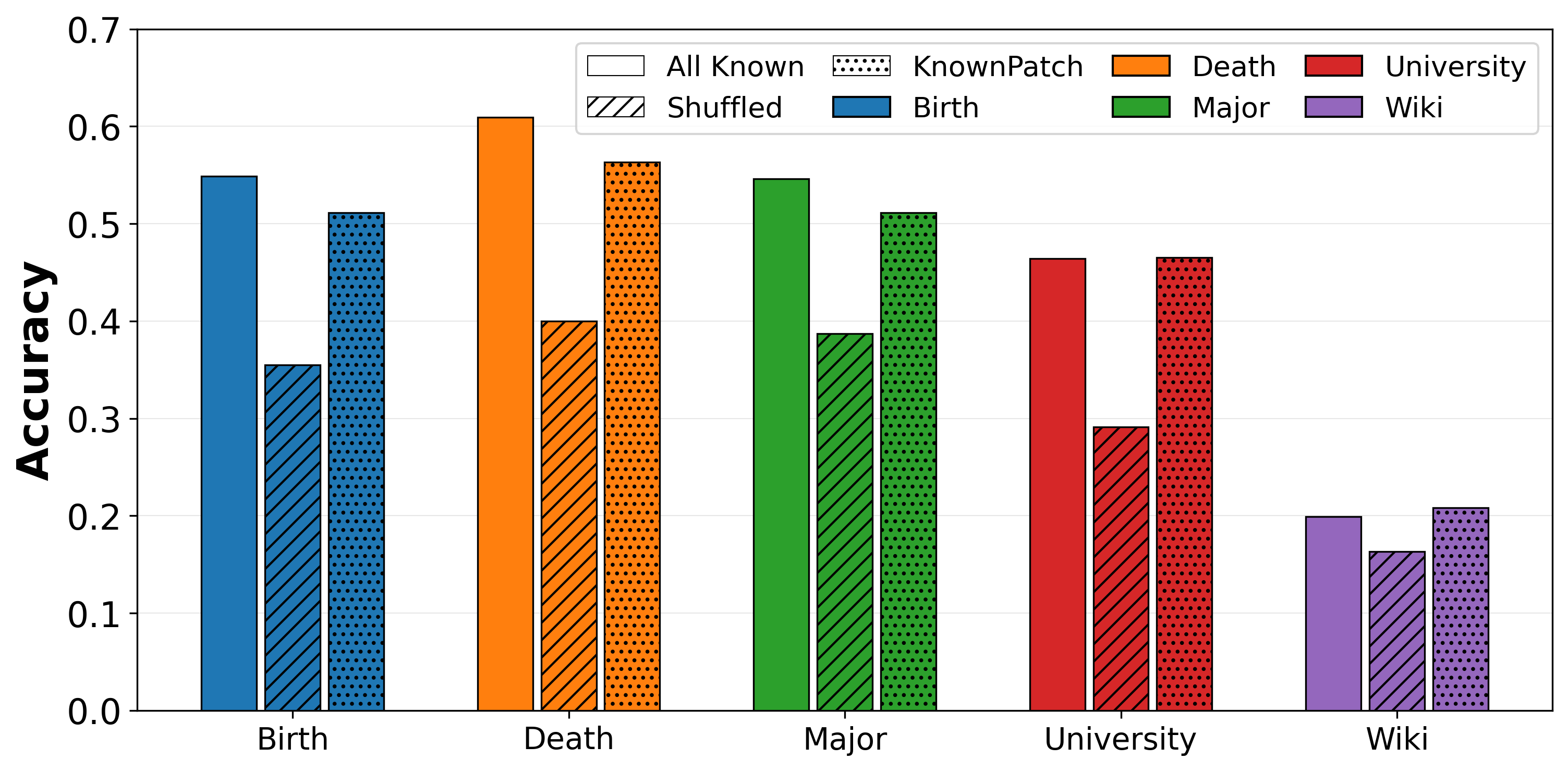}
    \caption{Performance of KnownPatch on QA task when injecting 5\% (upper), 10\% (middle) and 20\% (lower) known data. All experiments trained for 3 epoch.}
    \label{fig:KnownPatch_qa_5_10_20_3ep}
\end{figure}

\begin{figure}[!htbp]
    \centering
    \includegraphics[width=0.48\textwidth]{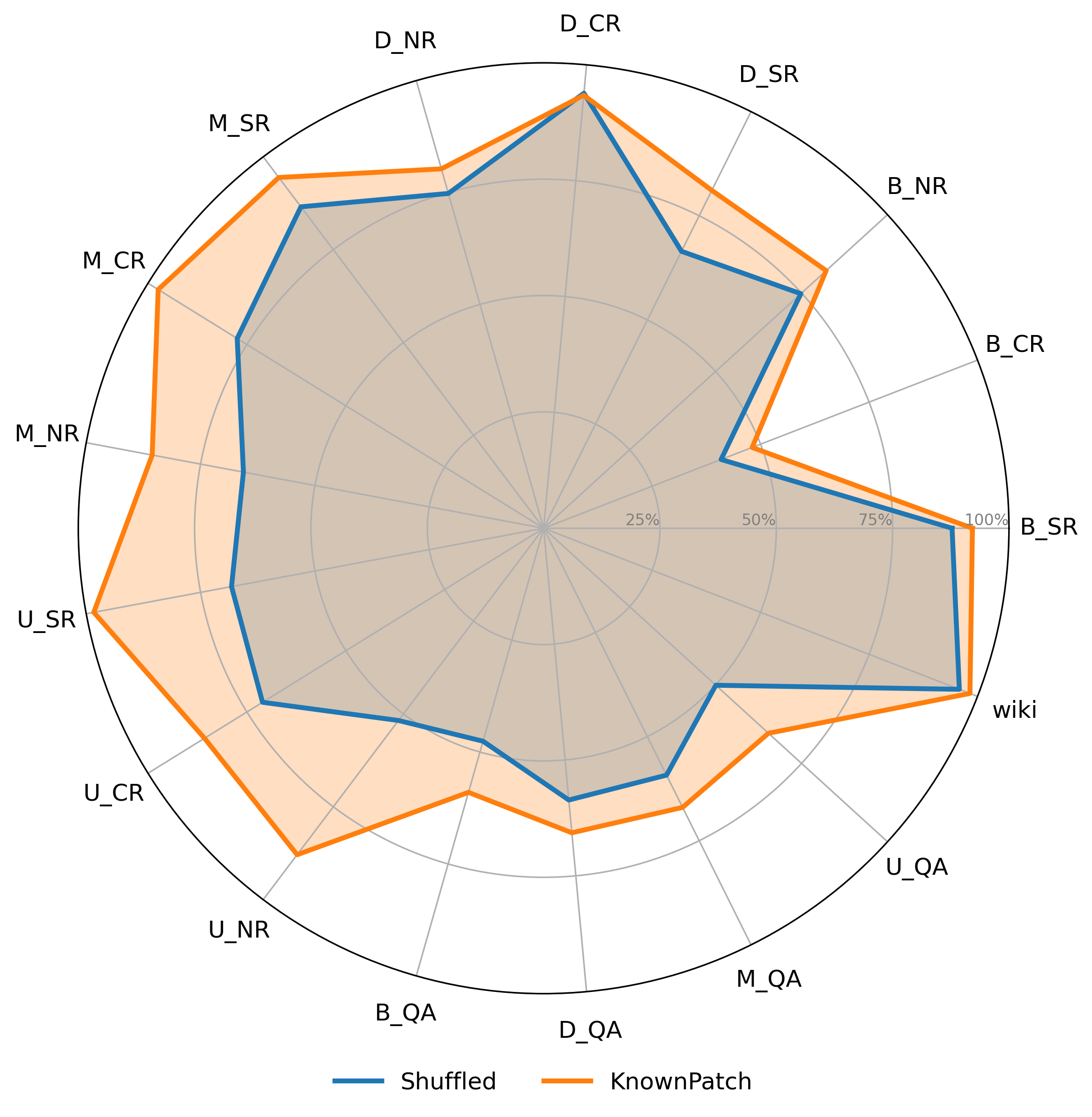}
    \caption{KnownPatch on reasoning tasks with 5\% injection ratio. All experiments trained for 3 epoch.}
    \label{fig:KnownPatch_reasoning_5}
\end{figure}

\begin{figure}[!htbp]
    \centering
    \includegraphics[width=0.48\textwidth]{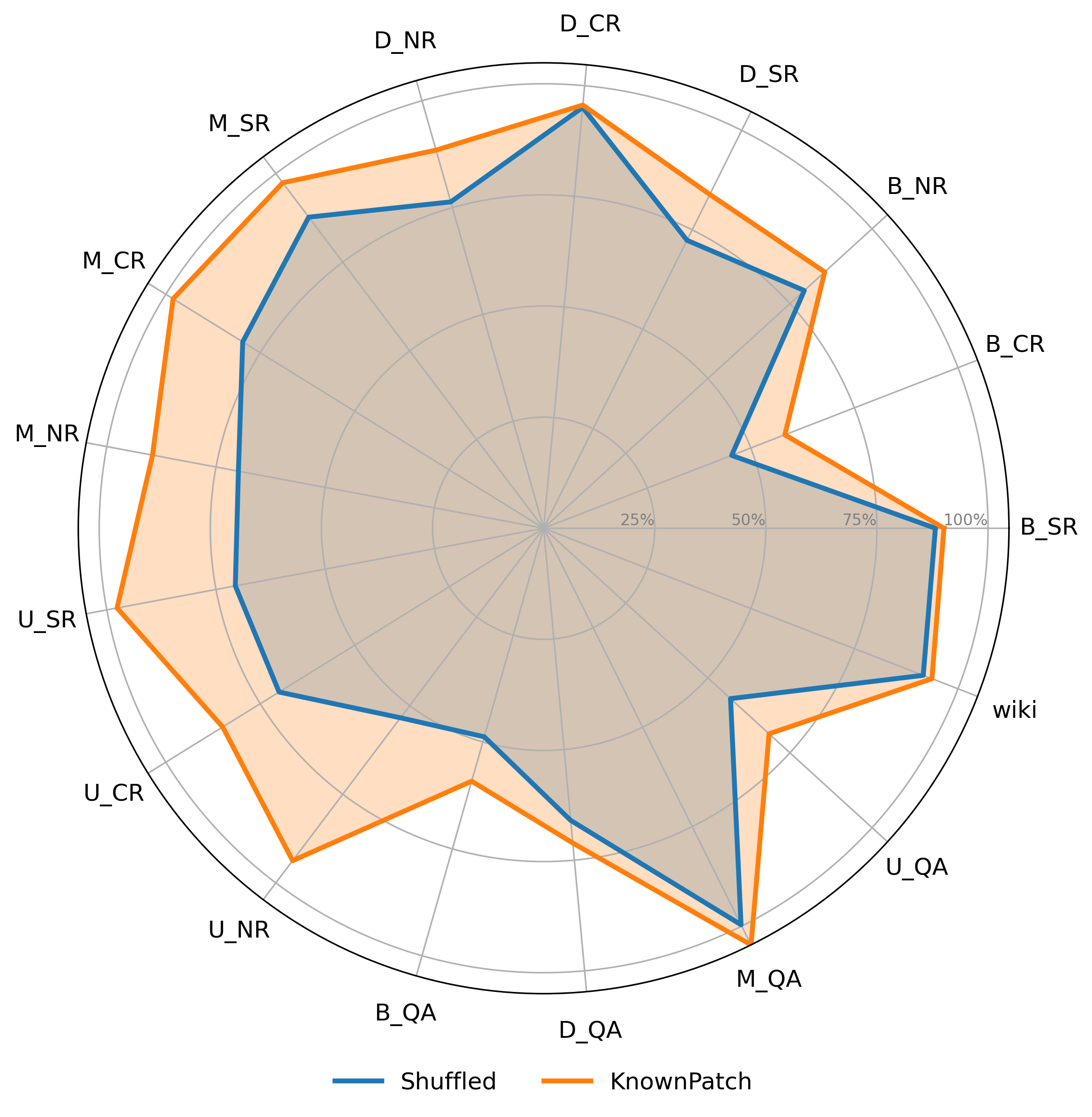}
    \caption{KnownPatch on reasoning tasks with 10\% injection ratio. All experiments trained for 3 epoch.}
    \label{fig:KnownPatch_reasoning_10}
\end{figure}

\begin{figure}[!htbp]
    \centering
    \includegraphics[width=0.48\textwidth]{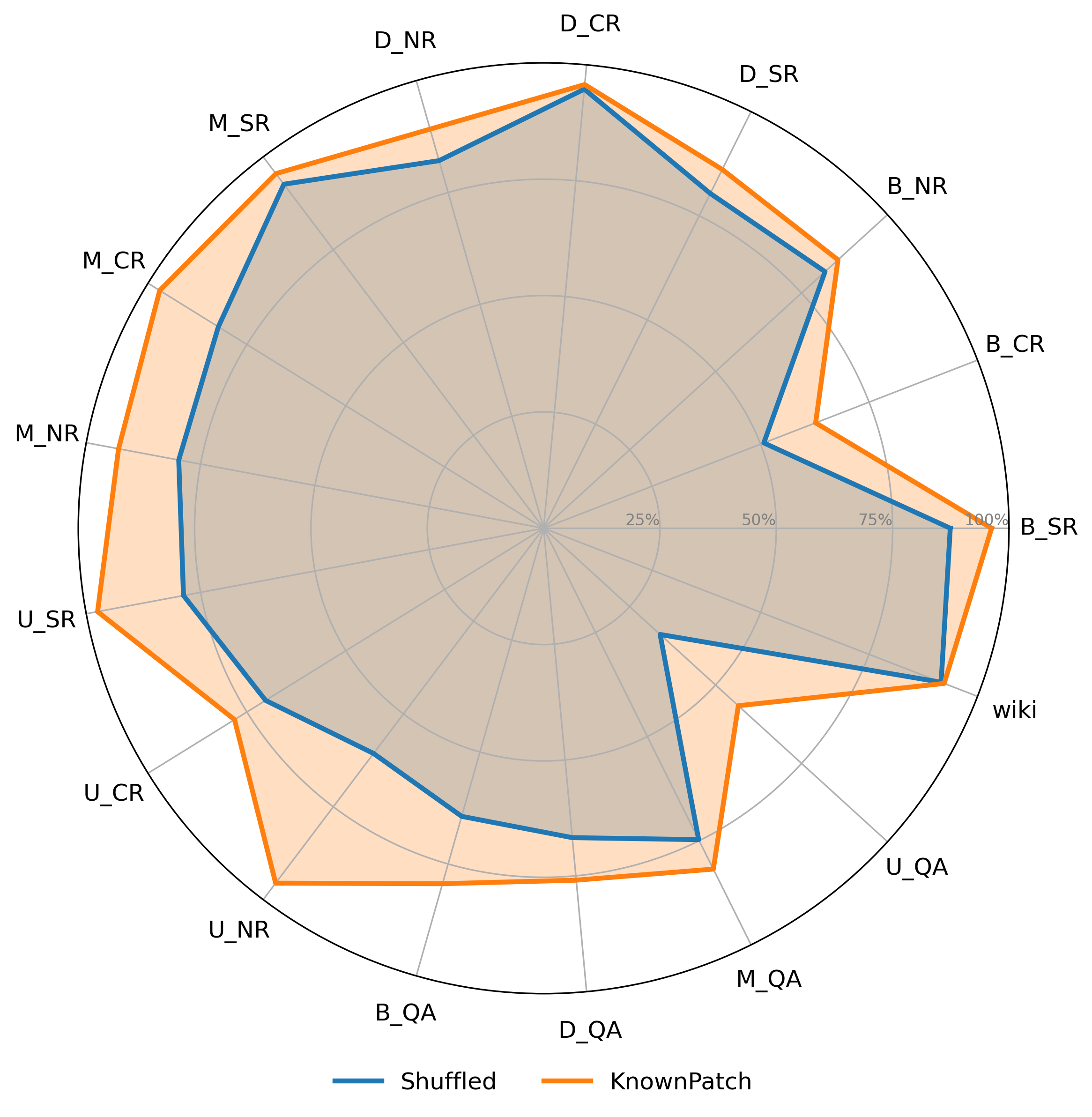}
    \caption{KnownPatch on reasoning tasks with 20\% injection ratio. All experiments trained for 3 epoch.}
    \label{fig:KnownPatch_reasoning_20}
\end{figure}

\begin{figure}[!htbp]
    \centering
    \includegraphics[width=0.48\textwidth]{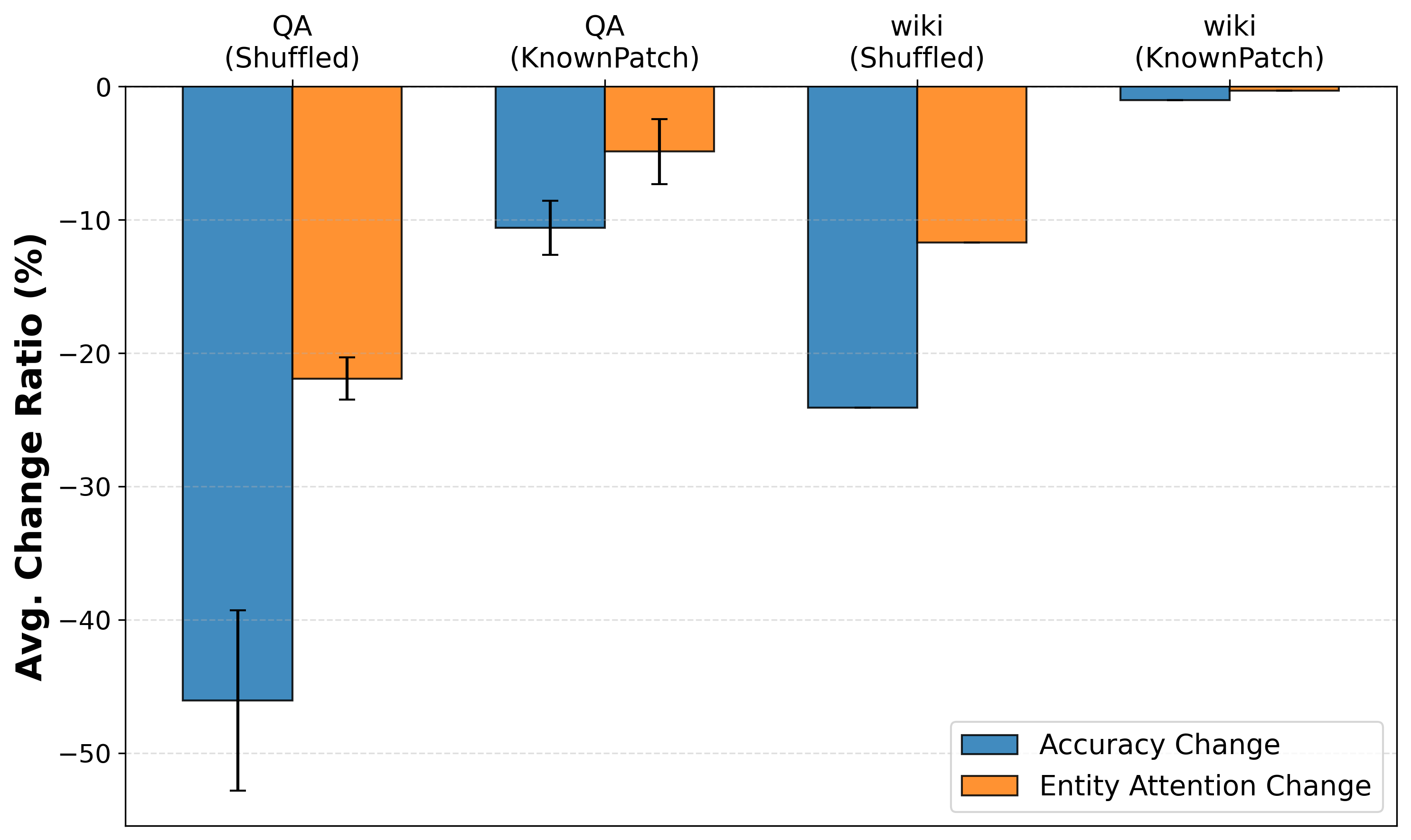}
    \caption{Performance and attention score changes when learning new knowledge in QA tasks, and after applying KnownPatch (with 10\% known data). QA represents the average across the four QA test sets, and error bars indicate standard deviations.}
    \label{fig:qa_mitigating_interpretability_bar_10}
\end{figure}

\begin{figure}[!htbp]
    \centering
    \includegraphics[width=0.48\textwidth]{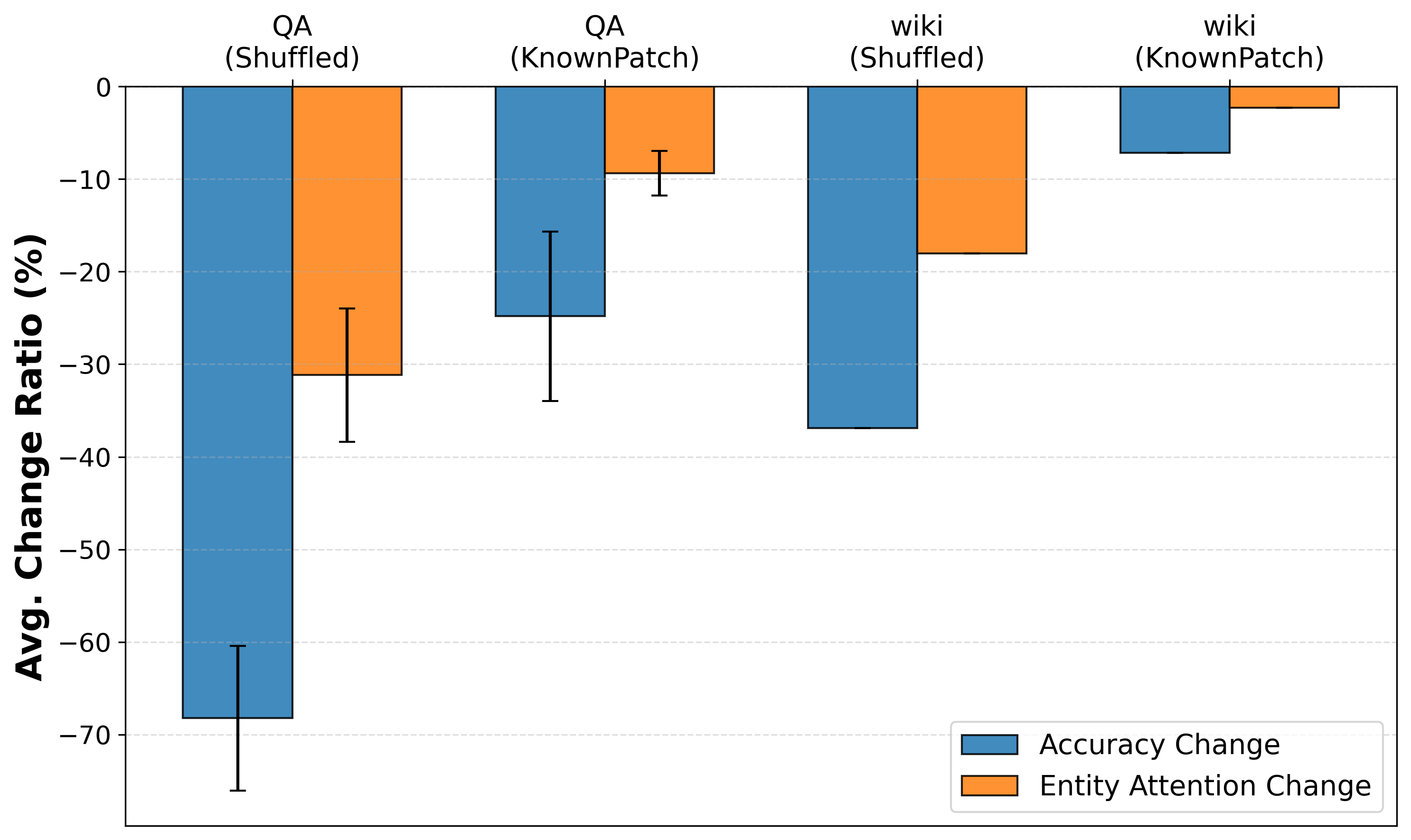}
    \caption{Performance and attention score changes when learning new knowledge in QA tasks, and after applying KnownPatch (with 5\% known data). QA represents the average across the four QA test sets, and error bars indicate standard deviations.}
    \label{fig:qa_mitigating_interpretability_bar_5}
\end{figure}

Figure \ref{fig:KnownPatch_qa_5_10_20_3ep} is the result of KnownPatch on QA tasks with different injection ratios; Figures \ref{fig:KnownPatch_reasoning_5}, \ref{fig:KnownPatch_reasoning_10} and \ref{fig:KnownPatch_reasoning_20} are results of KnownPatch on reasoning tasks with injection ratios of 5\%, 10\% and 20\%; Figures \ref{fig:qa_mitigating_interpretability_bar_10} and \ref{fig:qa_mitigating_interpretability_bar_5} are interpretability results of injecting 10\% and 5\% known data in KnownPatch.

\begin{figure}[htb]
    \centering
    \includegraphics[width=0.44\textwidth]{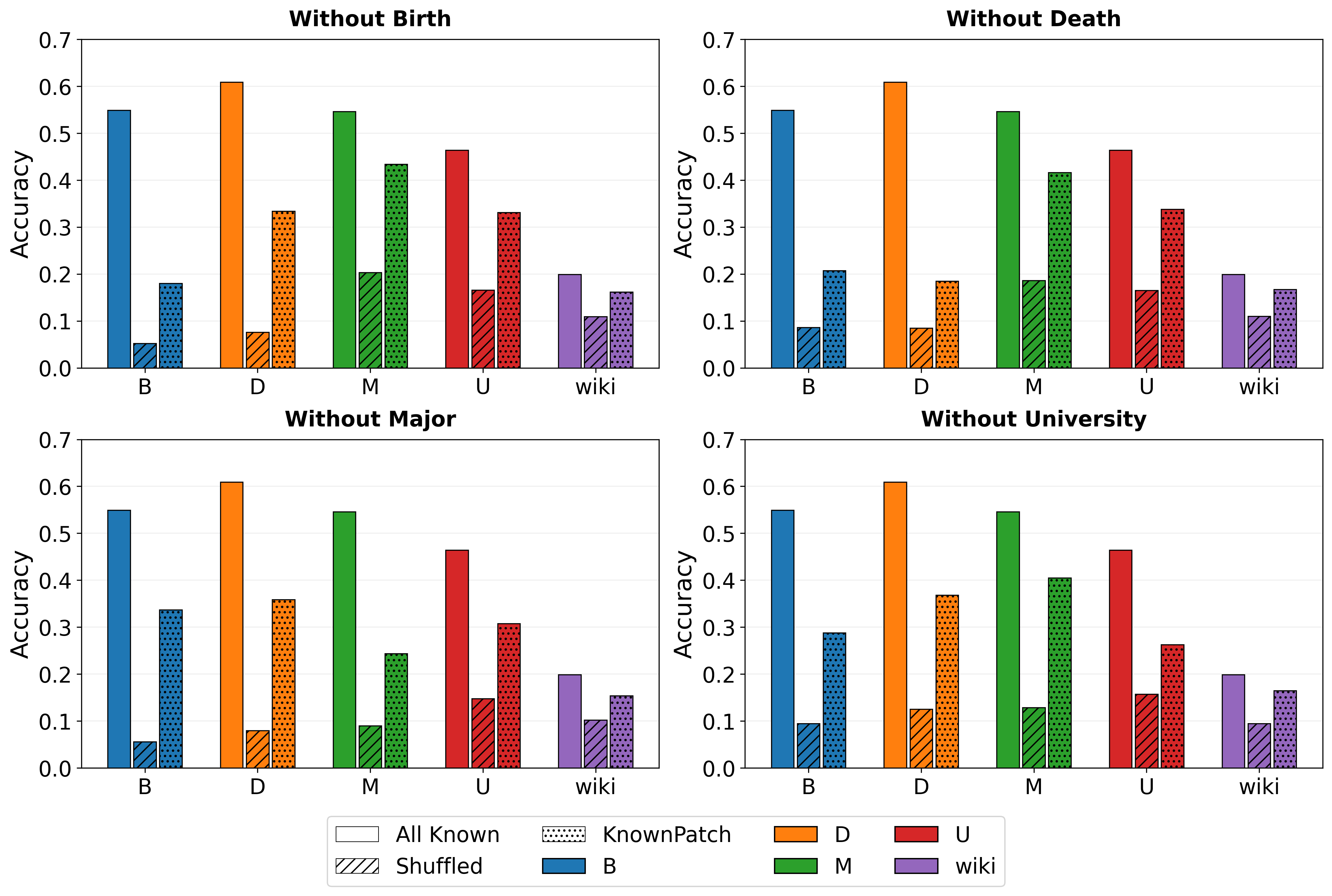}
    \caption{KnownPatch (missing one knowledge type) on QA tasks with an injection ratio of 5\%. All experiments trained for 1 epoch.}
    \label{fig:qa_mitigating_missing_cate_u950k50-3ep}
\end{figure}

\begin{figure}[htb]
    \centering
    \includegraphics[width=0.44\textwidth]{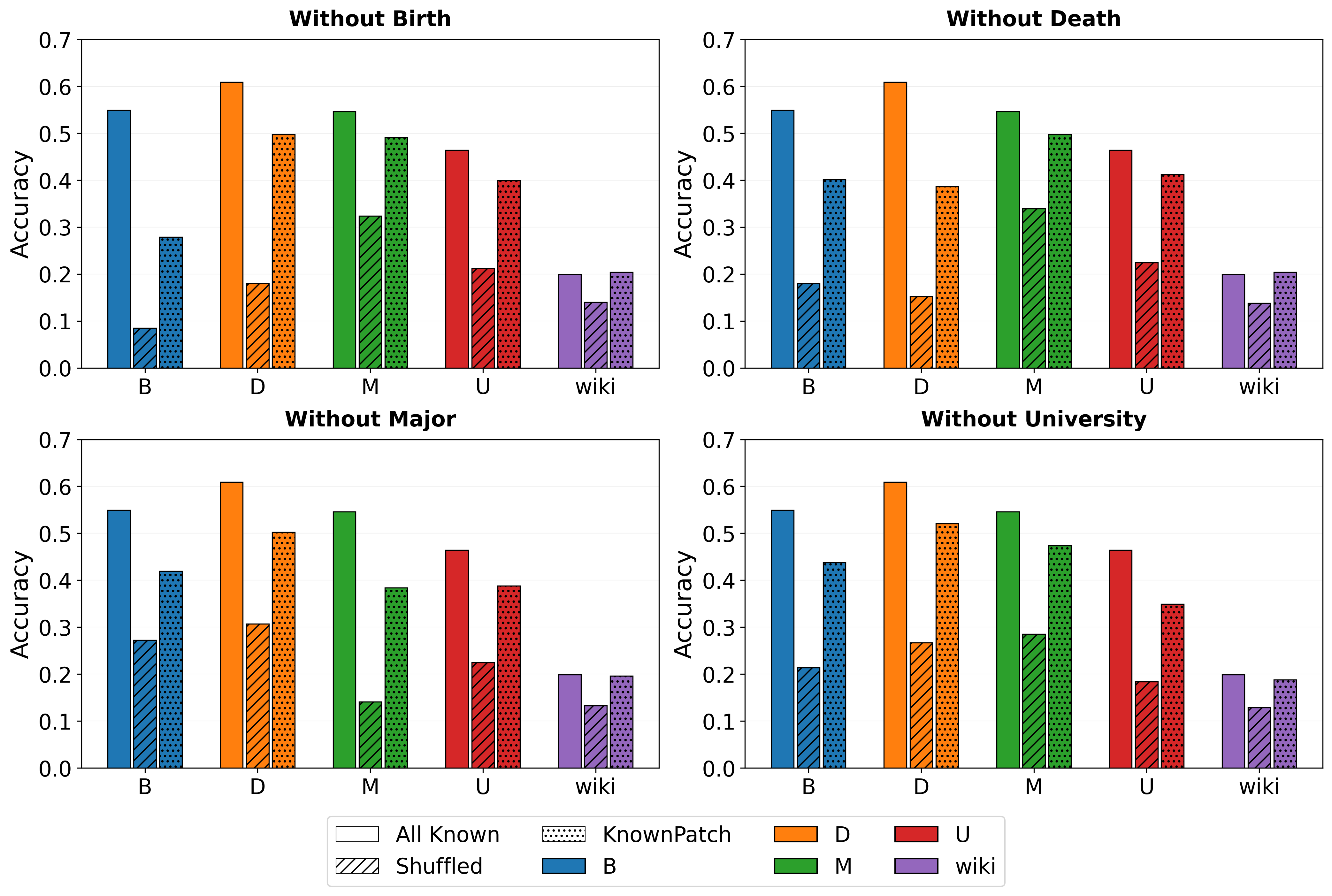}
    \caption{KnownPatch (missing one knowledge type) on QA tasks with an injection ratio of 10\%. All experiments trained for 1 epoch.}
    \label{fig:qa_mitigating_missing_cate_u900k100-3ep}
\end{figure}

\begin{figure}[htb]
    \centering
    \includegraphics[width=0.44\textwidth]{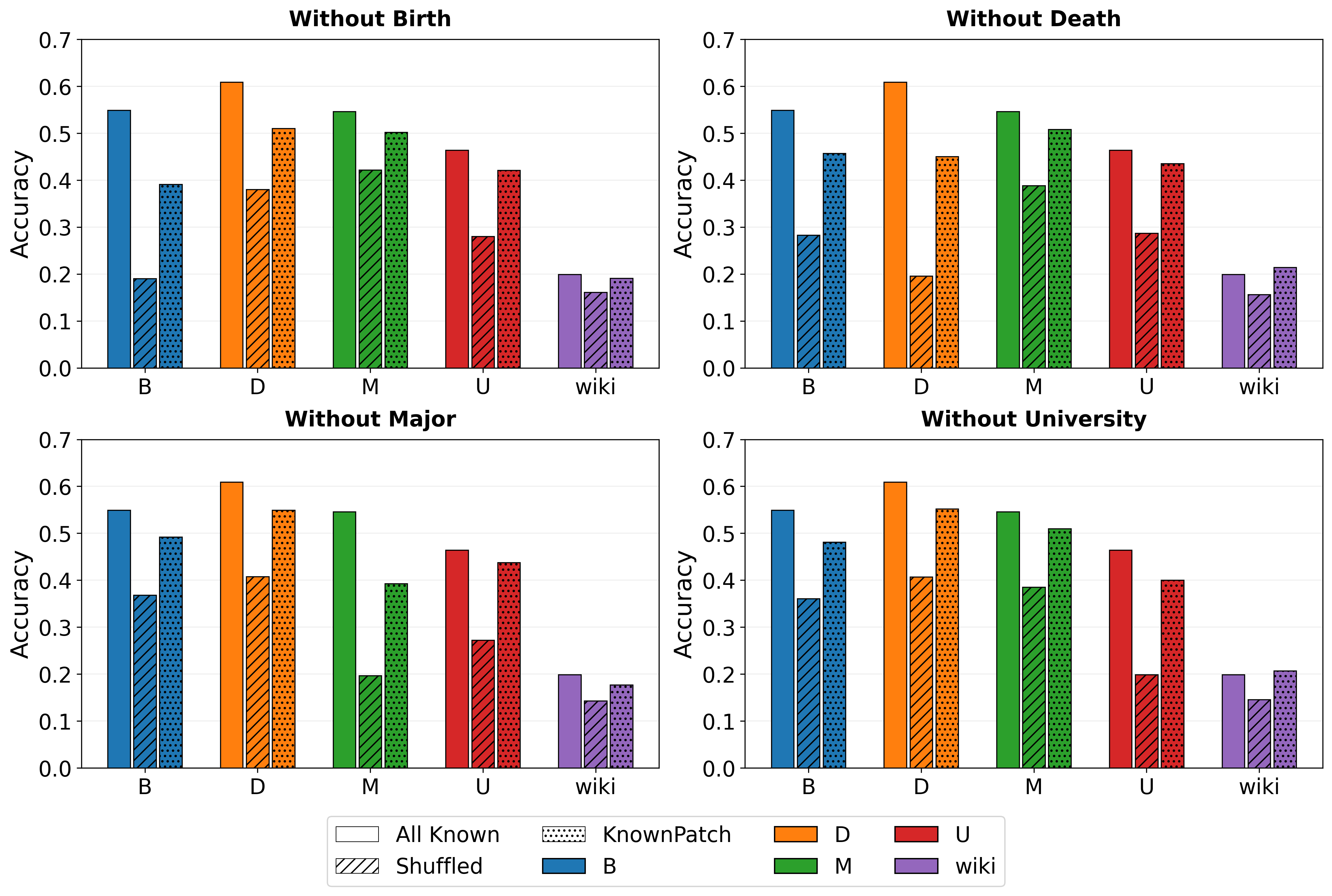}
    \caption{KnownPatch (missing one knowledge type) on QA tasks with an injection ratio of 20\%. All experiments trained for 1 epoch.}
    \label{fig:qa_mitigating_missing_cate_u800k200-3ep}
\end{figure}

In a more realistic scenario, the injected known knowledge does not cover all knowledge types. We therefore specifically examine the case where known data from one type is missing. For a knowledge type that has only been observed in the unknown data, where the known data used by KnownPatch comes solely from other knowledge types, Figures~\ref{fig:qa_mitigating_missing_cate_u950k50-3ep}, \ref{fig:qa_mitigating_missing_cate_u900k100-3ep} and \ref{fig:qa_mitigating_missing_cate_u800k200-3ep} shows that the method can still substantially mitigate factual hallucinations under injection ratios 5\%, 10\% and 20\%, respectively.

\FloatBarrier

\section{Results on Different Models}
\label{app:different_models}
We used Qwen2.5-1.5B (main text) and Qwen3-8B, Llama3.2-1B (appendix), spanning architectures and sizes, all supporting our conclusions. Due to resource limitations, larger-scale training was infeasible. For larger models, prior work \citep{AL2024-knowledge3} shows they retain factual knowledge better. 

\subsection{Results on Llama-3.2-1B}
\begin{figure}[ht]
    \centering
    \includegraphics[width=0.48\textwidth]{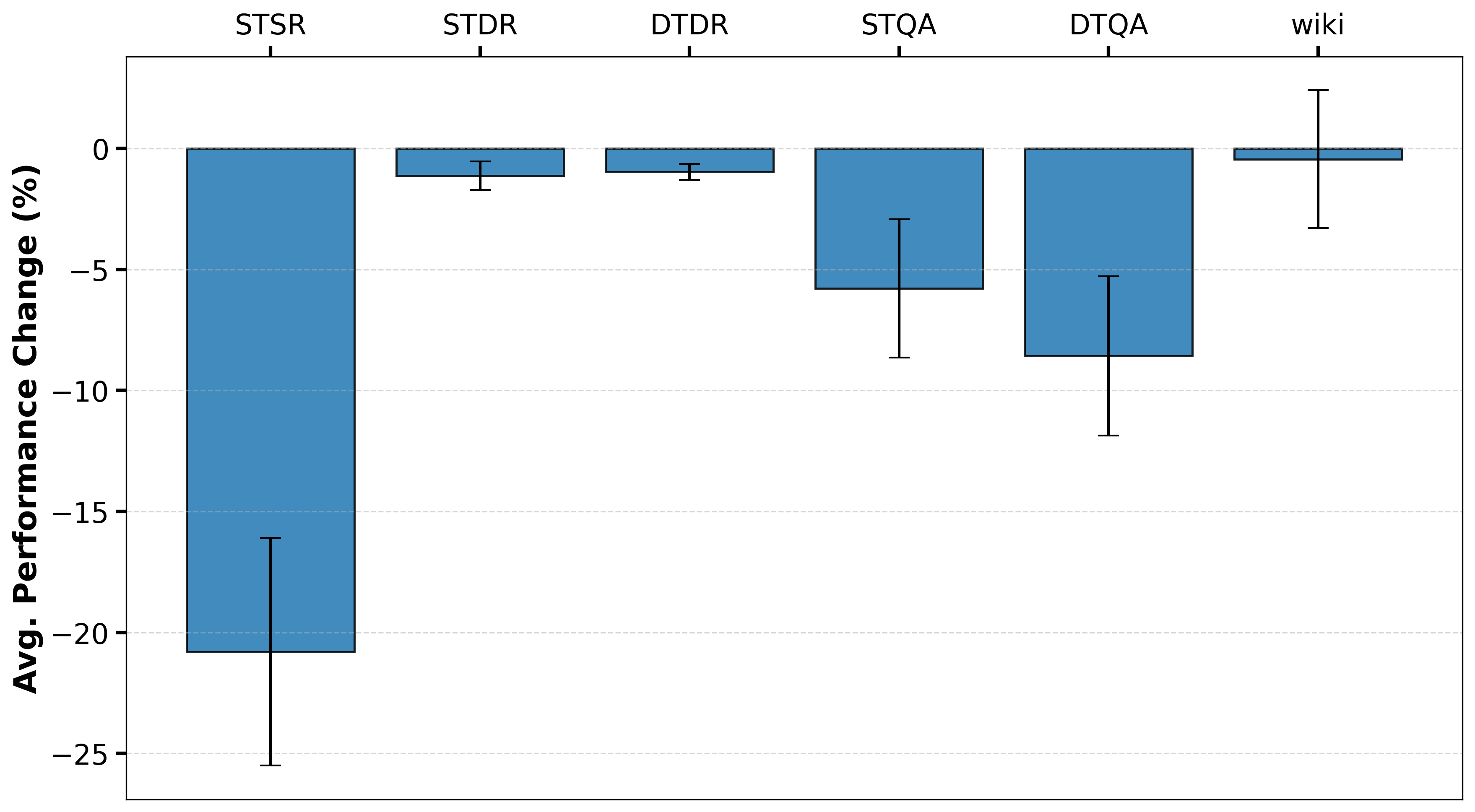}
    \caption{The impact of learning new knowledge in reasoning tasks on the average performance across different groups (on the Llama-3.2-1B model).}
    \label{fig:llama_reasoning_big_table}
\end{figure}

In this section we provide results of Llama-3.2-1B. Table \ref{tab:llama_qa_results} (similar to Table \ref{tab:qa_results}) provides the hallucination results in QA tasks when learning new knowledge; Figure \ref{fig:llama_reasoning_big_table} (similar to Figure \ref{fig:reasoning_big_table}) shows the impact of new knowledge in reasoning tasks on different groups.

\begin{table}[ht]
    \centering
    \begin{tabular}{c c c}
    \toprule
    STQA & DTQA & Wiki \\
    \midrule
    -49.80 {\footnotesize($\pm$ 11.74)} 
    & -1.04 {\footnotesize($\pm$ 1.03)} 
    & -10.65 {\footnotesize($\pm$ 10.11)} \\
    \bottomrule
    \end{tabular}
    \caption{Llama-3.2-1B model's hallucination induced by training on different unknown knowledge types in QA tasks.}
    \label{tab:llama_qa_results}
\end{table}


\begin{figure}[ht]
    \centering
    \includegraphics[width=0.48\textwidth]{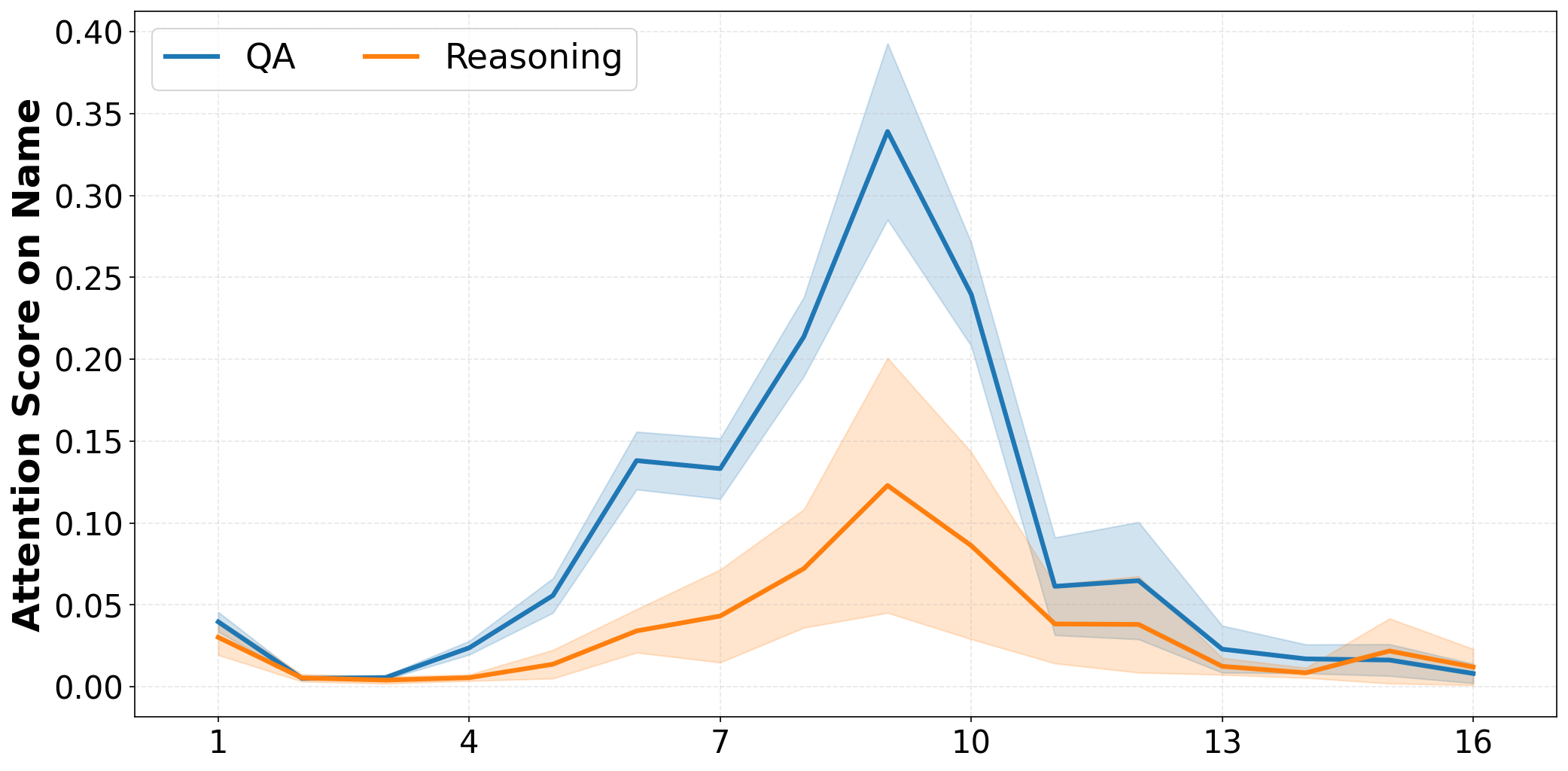}
    \caption{Llama-3.2-1B model's attention score on the target name across layers in QA and reasoning training setups. The solid curves show the average attention score at each layer, aggregated across all datasets and instances. The shaded regions represent the standard deviation.}
    \label{fig:llama_layer_selection}
\end{figure}

\begin{figure}[ht]
    \centering
    \includegraphics[width=0.48\textwidth]{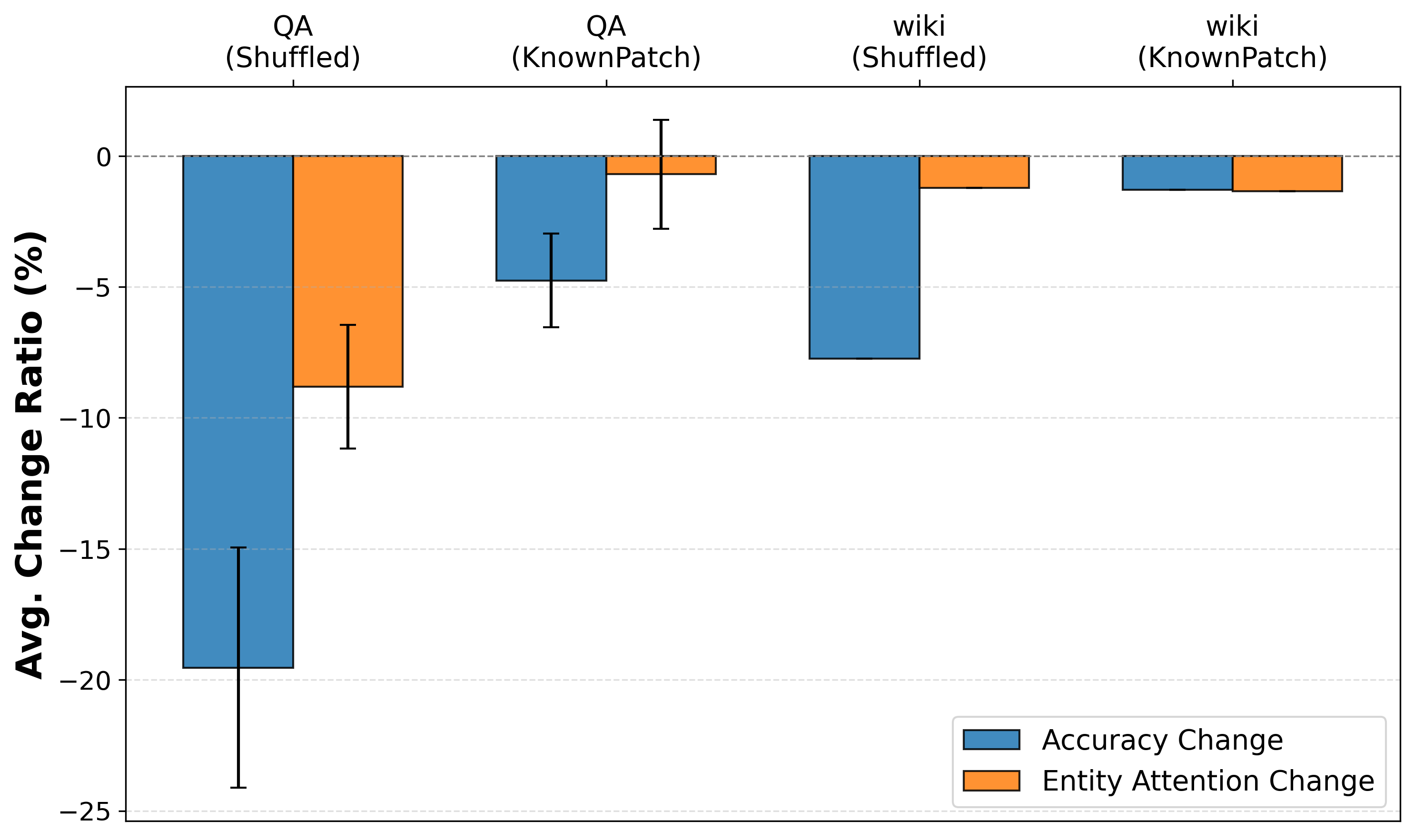}
    \caption{Llama-3.2-1B model's performance and attention score changes when learning new knowledge in QA tasks, and after applying KnownPatch (with 20\% known data). QA represents the average across the four QA test sets, and error bars indicate standard deviations.}
    \label{fig:llama_qa_mitigating_interpretability_bar}
\end{figure}

We also perform the same interpretability analysis as Section \ref{attentionanalysis} and Appendix \ref{app:attention_layer_details} on the Llama-3.2-1B model. Based on the results of Figure \ref{fig:llama_layer_selection} (similar to Figure \ref{fig:layer_selection}), we chose its layers 4-14 for further interpretability analysis, and Figure \ref{fig:llama_qa_mitigating_interpretability_bar} (similar to Figure \ref{fig:qa_mitigating_interpretability_bar}) shows the results.

\subsection{Results on Qwen3-8B-Base}
Due to the large scale of model parameters, our training setting differ from the default one. We only fine-tune for 1 epoch with a learning rate 5e-6 in all the SFT experiments.

\begin{figure}[ht]
    \centering
    \includegraphics[width=0.48\textwidth]{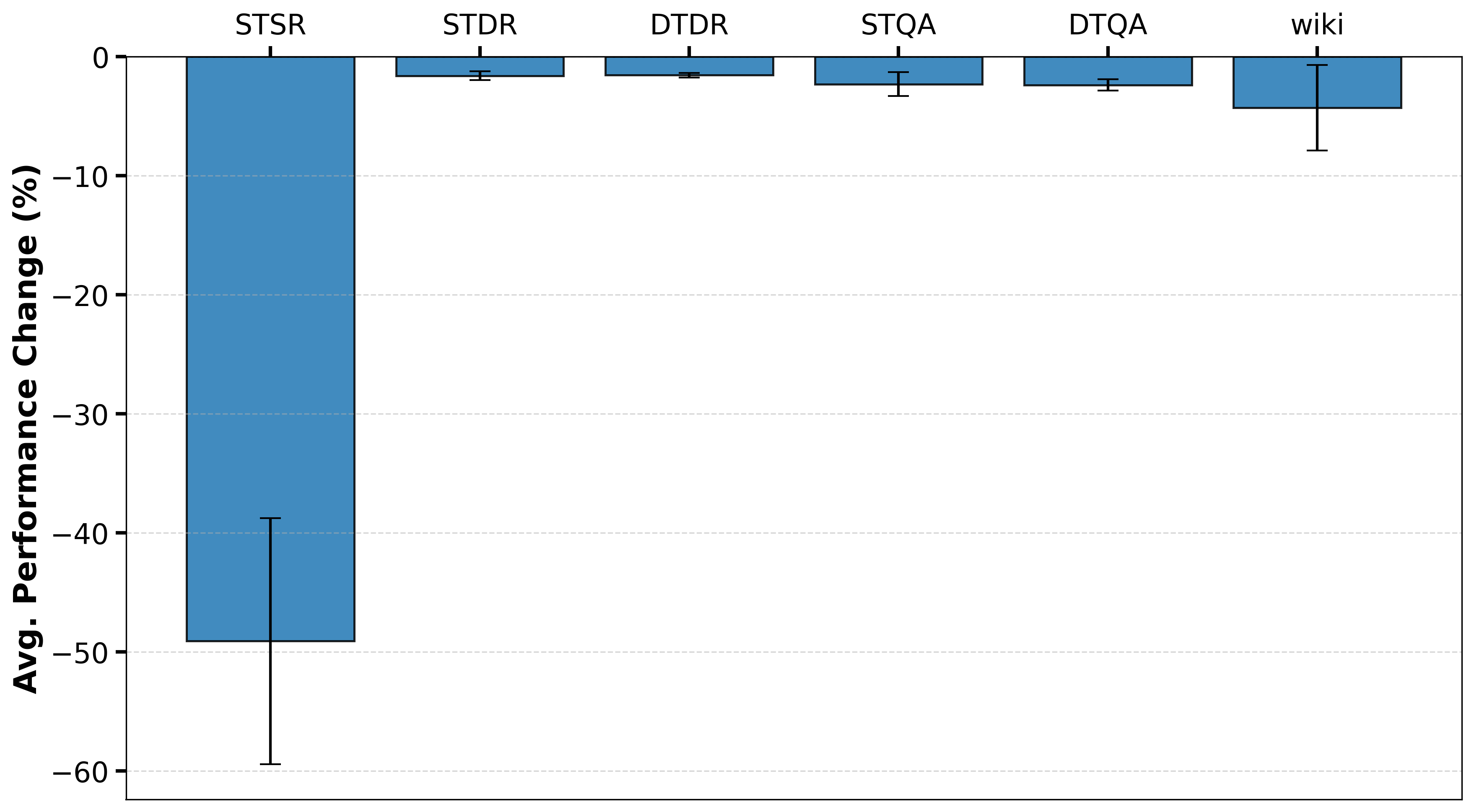}
    \caption{The impact of learning new knowledge in reasoning tasks on the average performance of different groups (on the Qwen3-8B-Base model).}
    \label{fig:qwen3_reasoning_big_table}
\end{figure}

Table \ref{tab:qwen3_qa_results} (similar to Table \ref{tab:qa_results}) provides the hallucination results in QA tasks when learning new knowledge; Figure \ref{fig:qwen3_reasoning_big_table} (similar to Figure \ref{fig:reasoning_big_table}) shows the impact of new knowledge in reasoning tasks on different groups.

\begin{table}[ht]
    \centering
    \begin{tabular}{c c c}
    \toprule
    STQA & DTQA & Wiki \\
    \midrule
    -80.64 {\footnotesize($\pm$ 1.23)} 
    & -2.33 {\footnotesize($\pm$ 1.42)} 
    & -13.29 {\footnotesize($\pm$ 8.40)} \\
    \bottomrule
    \end{tabular}
    \caption{Qwen3-8B-Base model's hallucination induced by training on different unknown knowledge types in QA tasks.}
    \label{tab:qwen3_qa_results}
\end{table}


\begin{figure}[ht]
    \centering
    \includegraphics[width=0.48\textwidth]{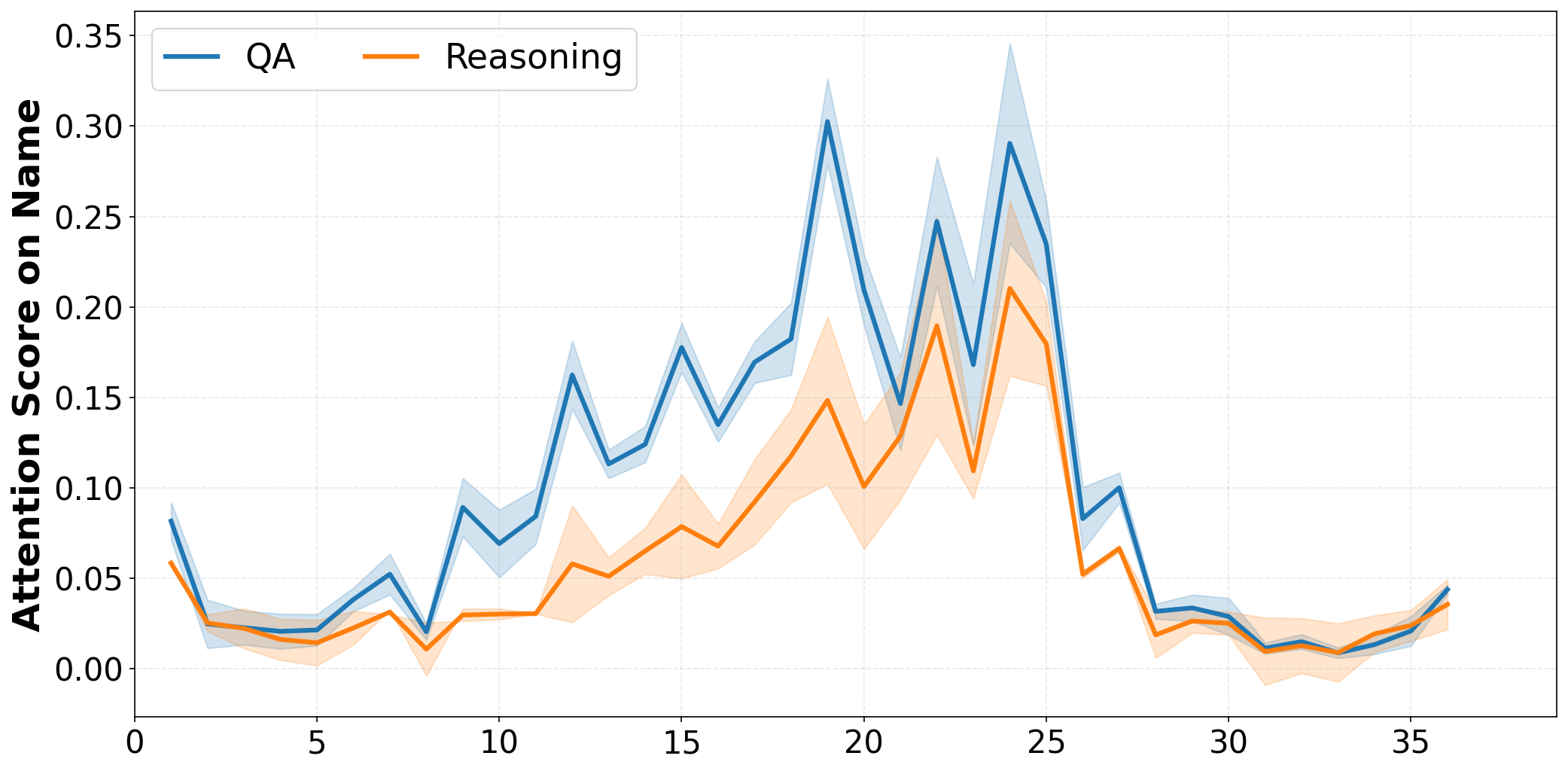}
    \caption{Qwen3-8B-Base model's attention score on the target name across layers in QA and reasoning training setups. The solid curves show the average attention score at each layer, aggregated across all datasets and instances. The shaded regions represent the standard deviation.}
    \label{fig:qwen3_layer_selection}
\end{figure}

\begin{figure}[ht]
    \centering
    \includegraphics[width=0.48\textwidth]{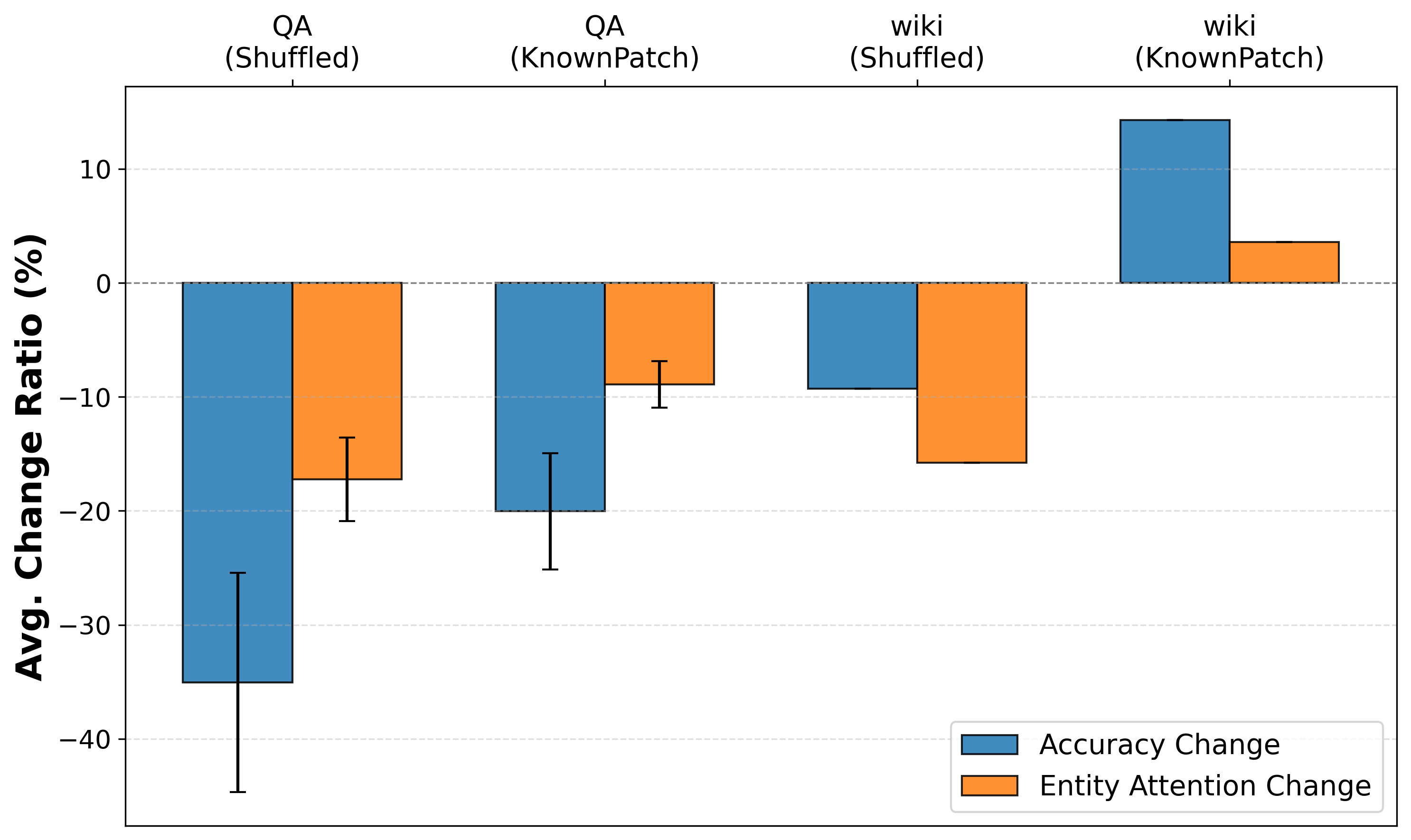}
    \caption{Qwen3-8B-Base model's performance and attention score changes when learning new knowledge in QA tasks, and after applying KnownPatch (with 20\% known data). QA represents the average across the four QA test sets, and error bars indicate standard deviations.}
    \label{fig:qwen3_qa_mitigating_interpretability_bar}
\end{figure}

We perform the same analysis as Section \ref{attentionanalysis} and Appendix \ref{app:attention_layer_details} on the Qwen3-8B-Base model. Based on the results of Figure \ref{fig:qwen3_layer_selection} (similar to Figure \ref{fig:layer_selection}), we chose its layers 9-27 for further interpretability analysis, and Figure \ref{fig:qwen3_qa_mitigating_interpretability_bar} (similar to Figure \ref{fig:qa_mitigating_interpretability_bar}) shows the results.

\subsection{Results on Qwen2.5-32B}
Due to the large scale of model parameters, our training setting differ from the default one. We only fine-tune for 1 epoch with a learning rate 5e-6 in all the SFT experiments.

\begin{figure}[ht]
    \centering
    \includegraphics[width=0.48\textwidth]{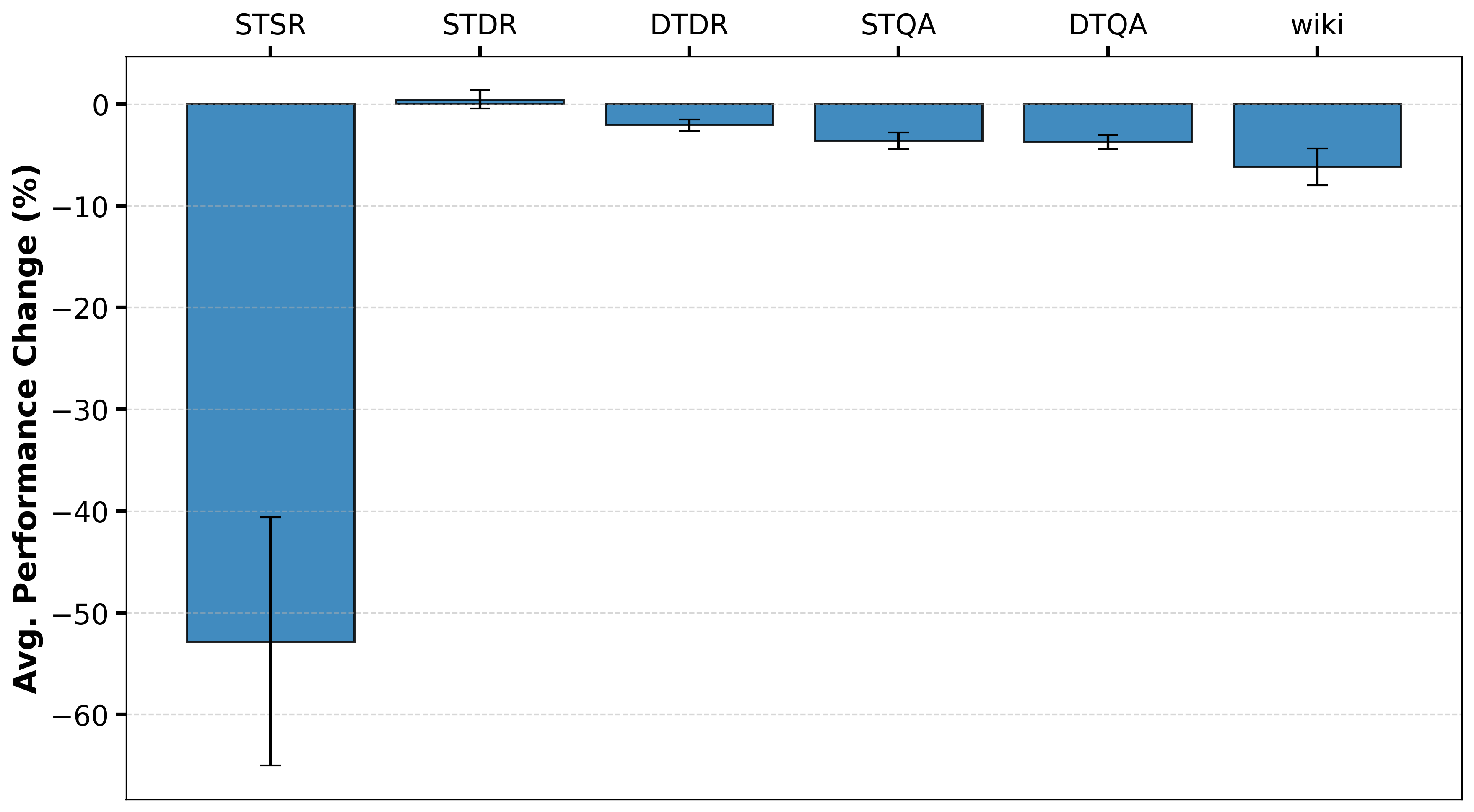}
    \caption{The impact of learning new knowledge in reasoning tasks on the average performance of different groups (on the Qwen2.5-32B model).}
    \label{fig:qwen_32B_reasoning_big_table}
\end{figure}

Table \ref{tab:qwen_32B_qa_results} (similar to Table \ref{tab:qa_results}) provides the hallucination results in QA tasks when learning new knowledge; Figure \ref{fig:qwen_32B_reasoning_big_table} (similar to Figure \ref{fig:reasoning_big_table}) shows the impact of new knowledge in reasoning tasks on different groups.

\begin{table}[ht]
    \centering
    \begin{tabular}{c c c}
    \toprule
    STQA & DTQA & Wiki \\
    \midrule
    -74.00 {\footnotesize($\pm$ 5.29)} 
    & -2.59 {\footnotesize($\pm$ 3.57)} 
    & -9.69 {\footnotesize($\pm$ 7.88)} \\
    \bottomrule
    \end{tabular}
    \caption{Qwen2.5-32B model's hallucination induced by training on different unknown knowledge types in QA tasks.}
    \label{tab:qwen_32B_qa_results}
\end{table}

\begin{figure}[ht]
    \centering
    \includegraphics[width=0.48\textwidth]{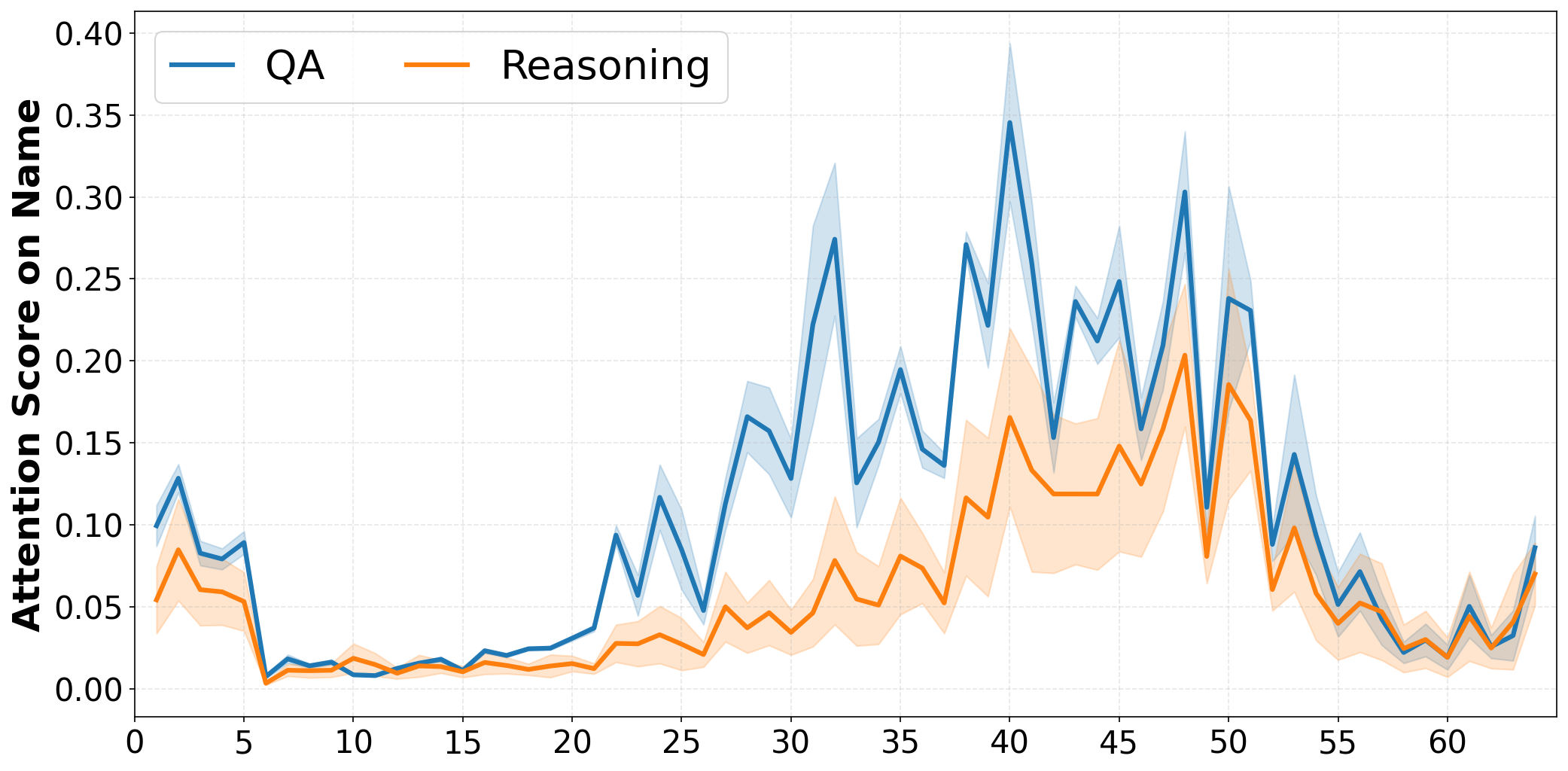}
    \caption{Qwen2.5-32B model's attention score on the target name across layers in QA and reasoning training setups. The solid curves show the average attention score at each layer, aggregated across all datasets and instances. The shaded regions represent the standard deviation.}
    \label{fig:qwen_32B_layer_selection}
\end{figure}

\begin{figure}[ht]
    \centering
    \includegraphics[width=0.48\textwidth]{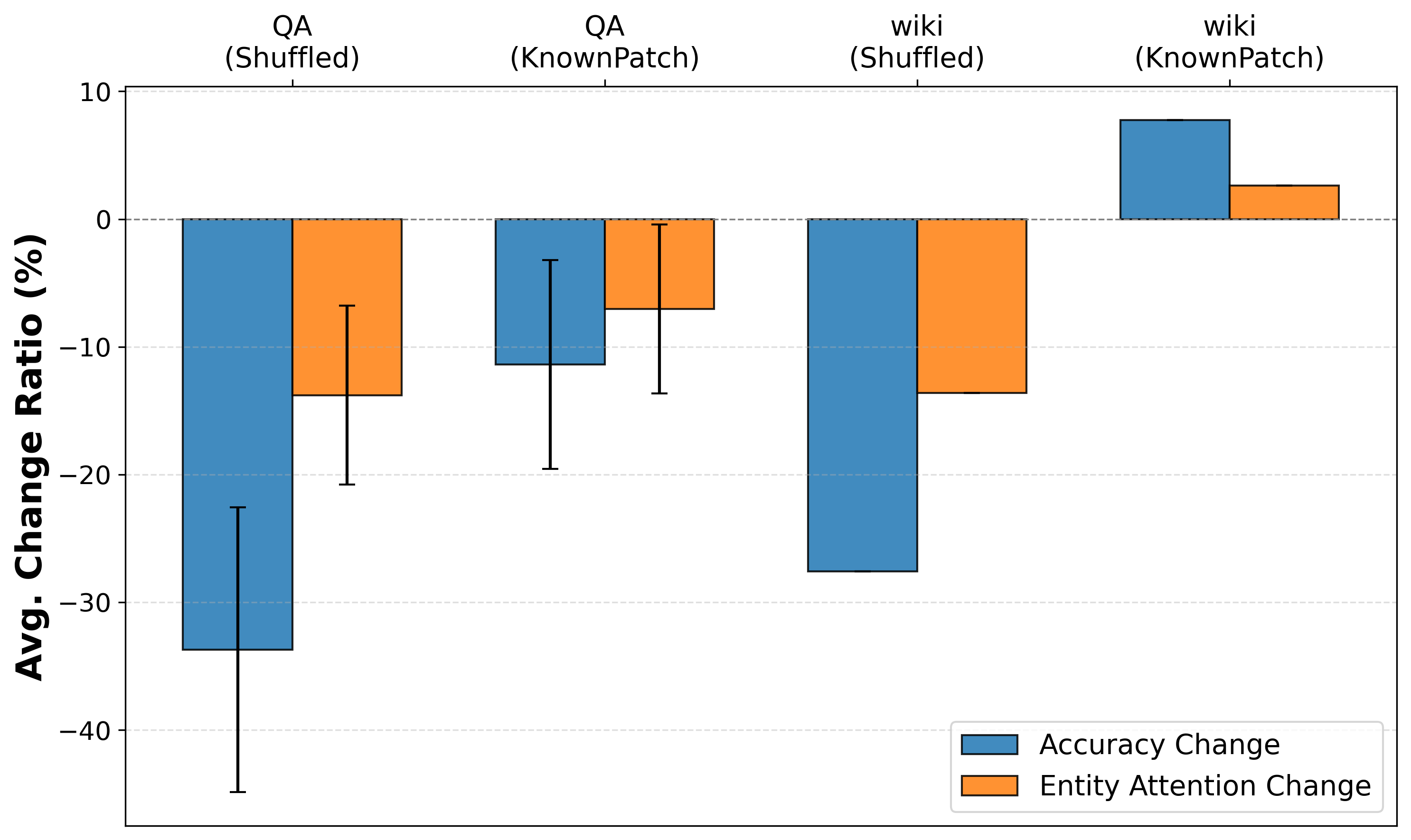}
    \caption{Qwen2.5-32B model's performance and attention score changes when learning new knowledge in QA tasks, and after applying KnownPatch (with 20\% known data). QA represents the average across the four QA test sets, and error bars indicate standard deviations.}
    \label{fig:qwen_32B_qa_mitigating_interpretability_bar}
\end{figure}

We perform the same analysis as Section \ref{attentionanalysis} and Appendix \ref{app:attention_layer_details} on the Qwen2.5-32B model. Based on the results of Figure \ref{fig:qwen_32B_layer_selection} (similar to Figure \ref{fig:layer_selection}), we chose its layers 25-55 for further interpretability analysis, and Figure \ref{fig:qwen_32B_qa_mitigating_interpretability_bar} (similar to Figure \ref{fig:qa_mitigating_interpretability_bar}) shows the results.

\section{Results on Different Training Epochs}
\label{app:different_epochs}
All results presented in the main text are obtained after SFT for 3 epochs. In this section, we report the results of the Qwen2.5-1.5B model under the same experimental configurations, with the number of training epochs adjusted to 1, 5 and 20. 
Notably, after 3 epochs of training, the model already achieves over 95\% accuracy on questions derived from known knowledge in the training set, and about 50\% accuracy on those constructed from unknown knowledge. When training is extended to 20 epochs, the model reaches over 95\% accuracy on unknown knowledge questions in the training set, and further training brings little additional improvement. We observe that the overall trends and results remain consistent across different numbers of training epochs.

\subsection{1 Epoch}
\begin{figure}[ht]
    \centering
    \includegraphics[width=0.48\textwidth]{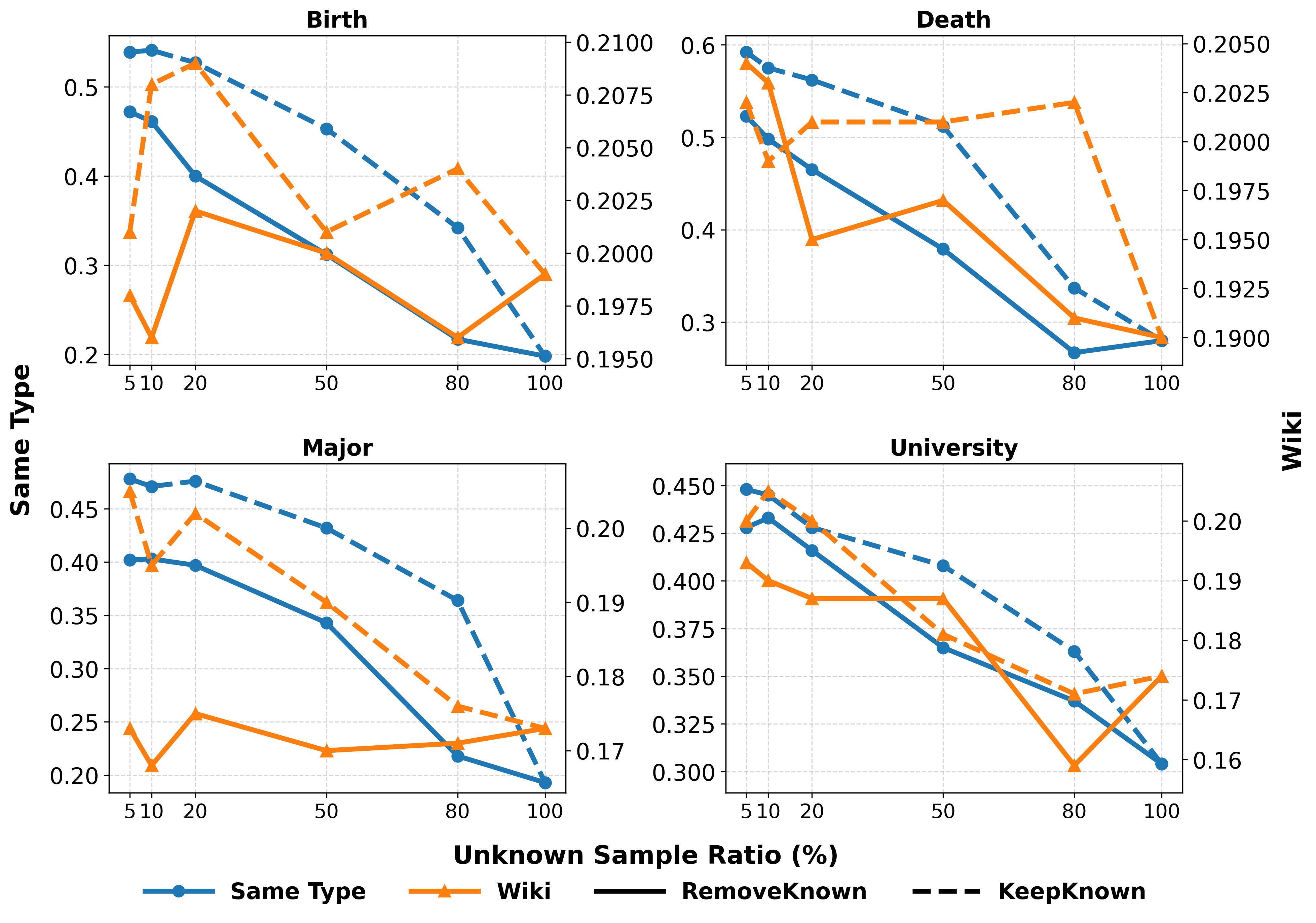}
    \caption{Performance in QA tasks under two settings with different proportions of unknown knowledge in the same type and wiki test set. All experiments trained for 1 epoch.}
    \label{fig:qa_unknown_percentage_1ep}
\end{figure}

\begin{figure}[ht]
    \centering
    \includegraphics[width=0.48\textwidth]{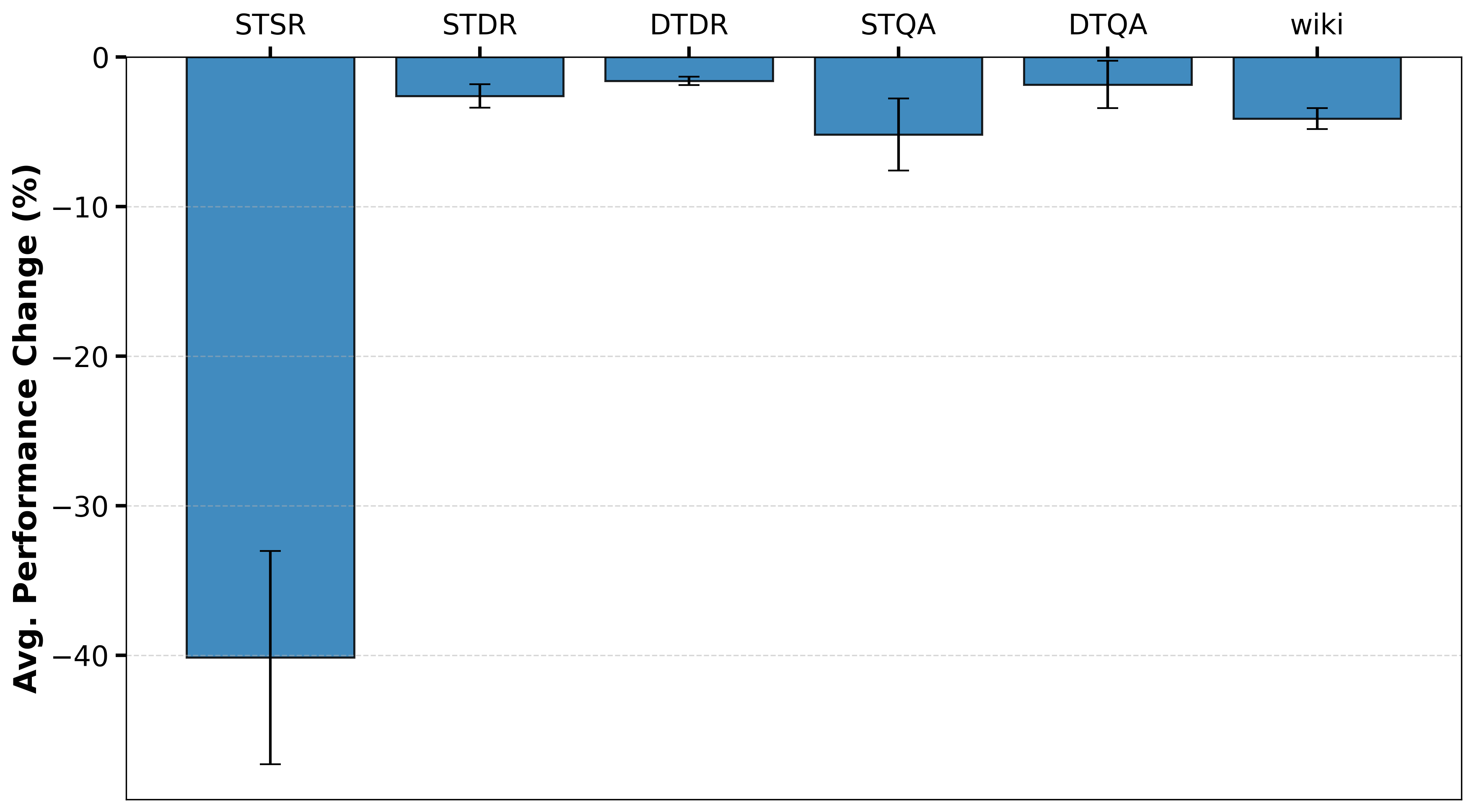}
    \caption{The impact of learning new knowledge in reasoning tasks on the average performance of different groups. All experiments trained for 1 epoch.}
    \label{fig:reasoning_big_table_1ep}
\end{figure}

\begin{figure}[ht]
    \centering
    \includegraphics[width=0.48\textwidth]{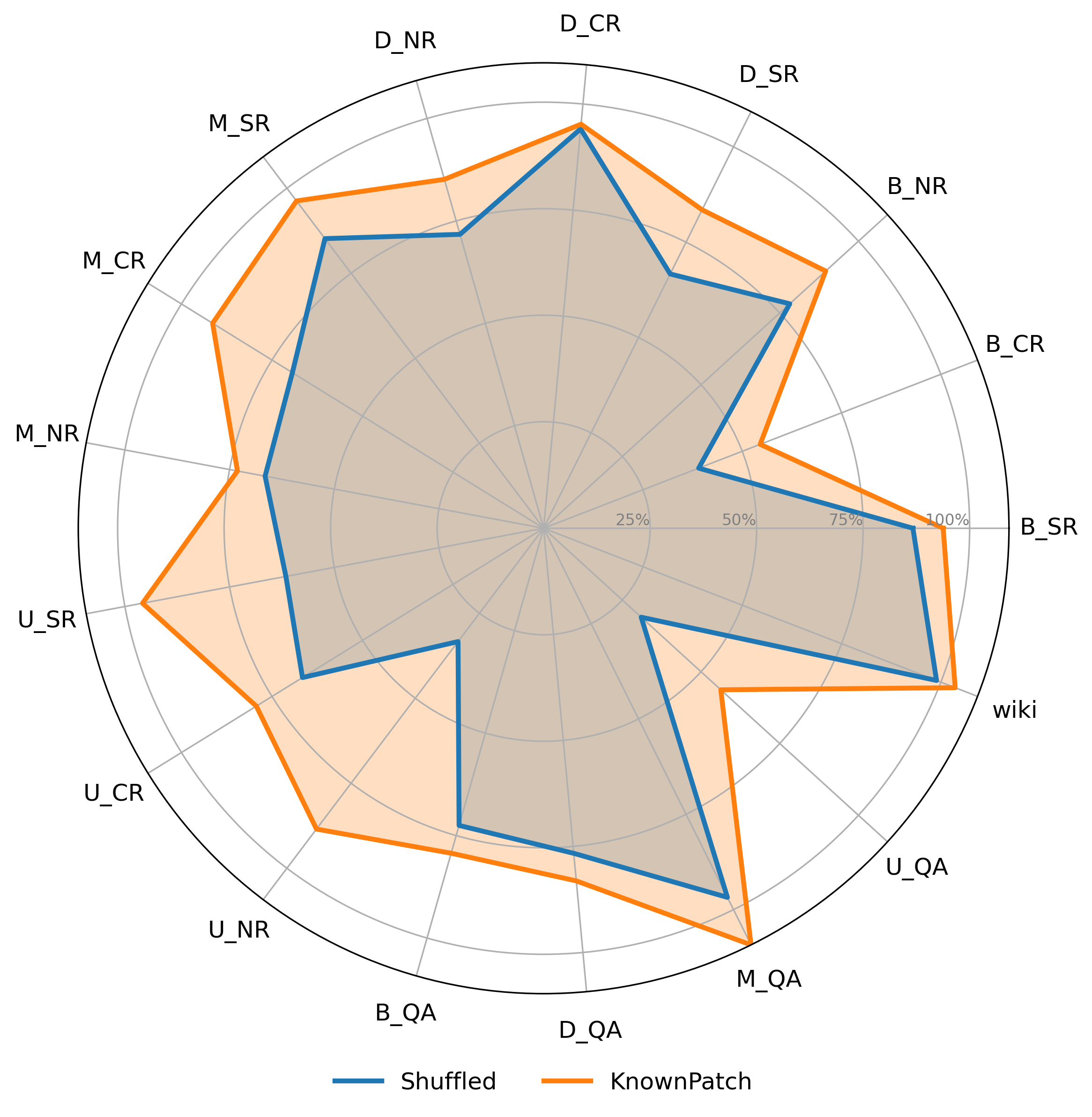}
    \caption{Performance of KnownPatch on reasoning task when injecting 20\% known data. The value here represents the accuracy percentage of this model compared to the fully known baseline model. All experiments trained for 1 epoch.}
    \label{fig:KnownPatch_reasoning_20_1ep}
\end{figure}

\begin{figure}[ht]
    \centering
    \includegraphics[width=0.48\textwidth]{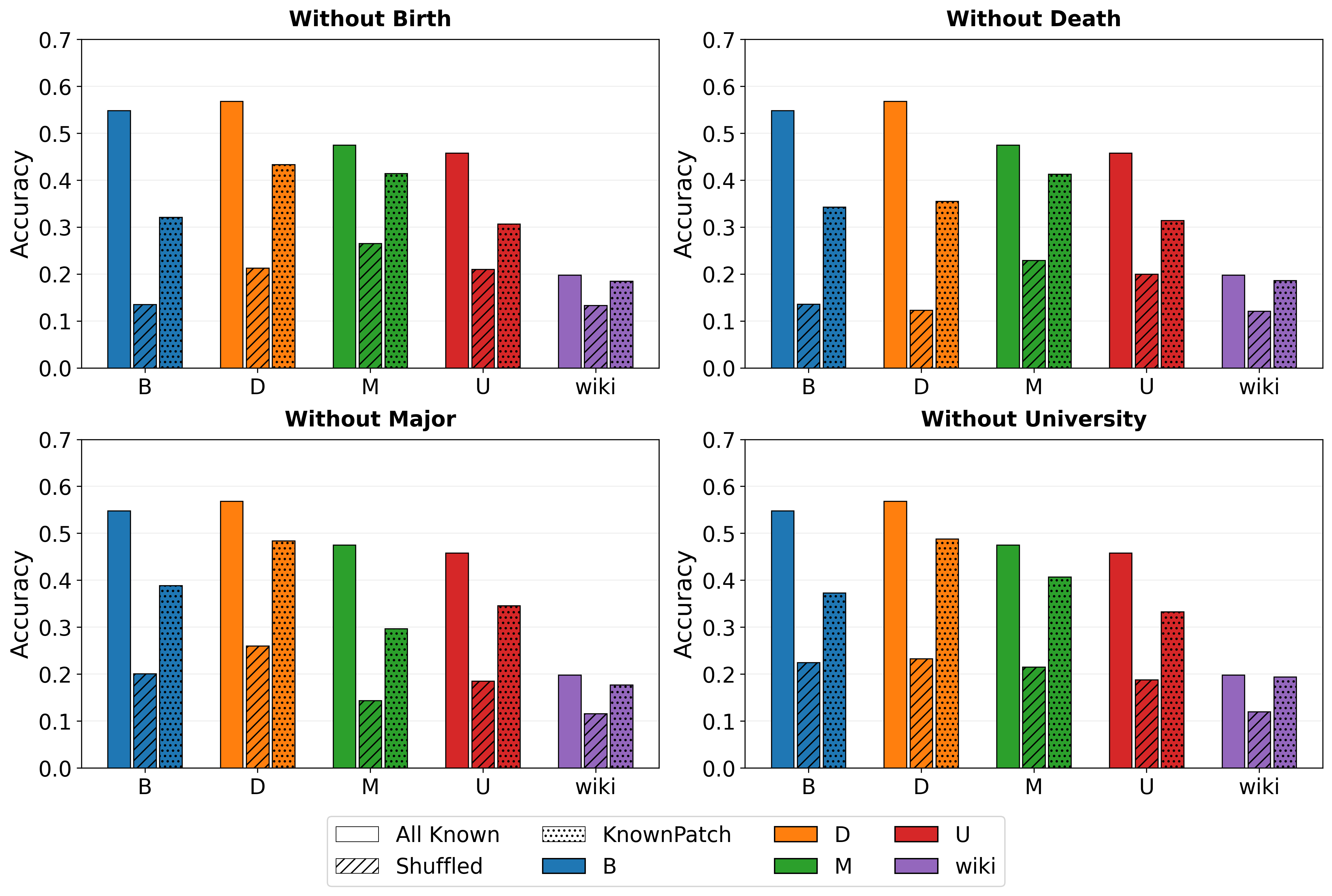}
    \caption{KnownPatch (missing one knowledge type) on QA tasks with an injection ratio of 20\%. All experiments trained for 1 epoch.}
    \label{fig:qa_mitigating_missing_cate_20_1ep}
\end{figure}

\begin{figure}[ht]
    \centering
    \includegraphics[width=0.48\textwidth]{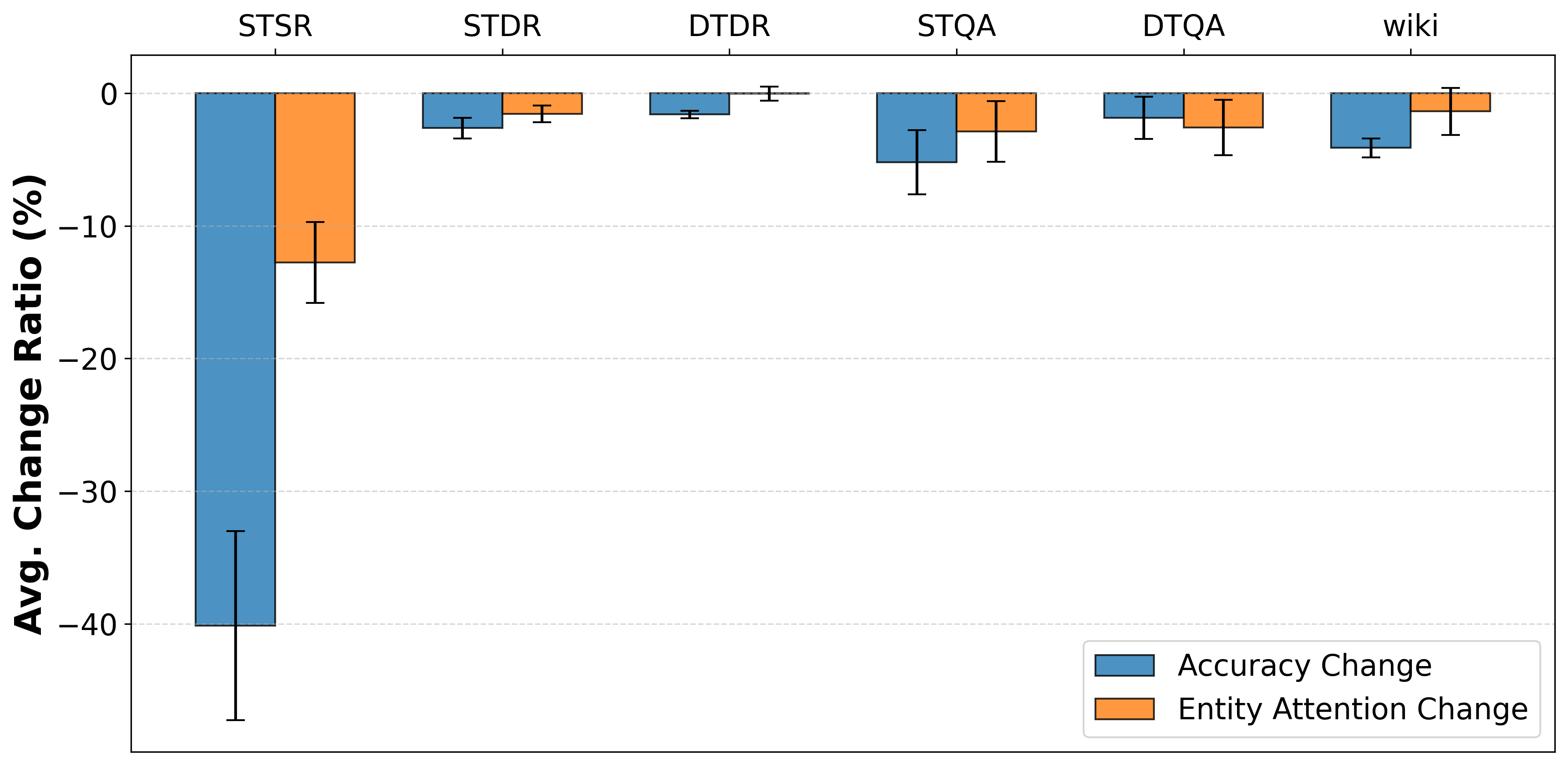}
    \caption{Accuracy and attention score changes when learning new knowledge in reasoning tasks. All experiments trained for 1 epoch.}
    \label{fig:reasoning_interpretability_bar_1ep}
\end{figure}

\begin{figure}[ht]
    \centering
    \includegraphics[width=0.48\textwidth]{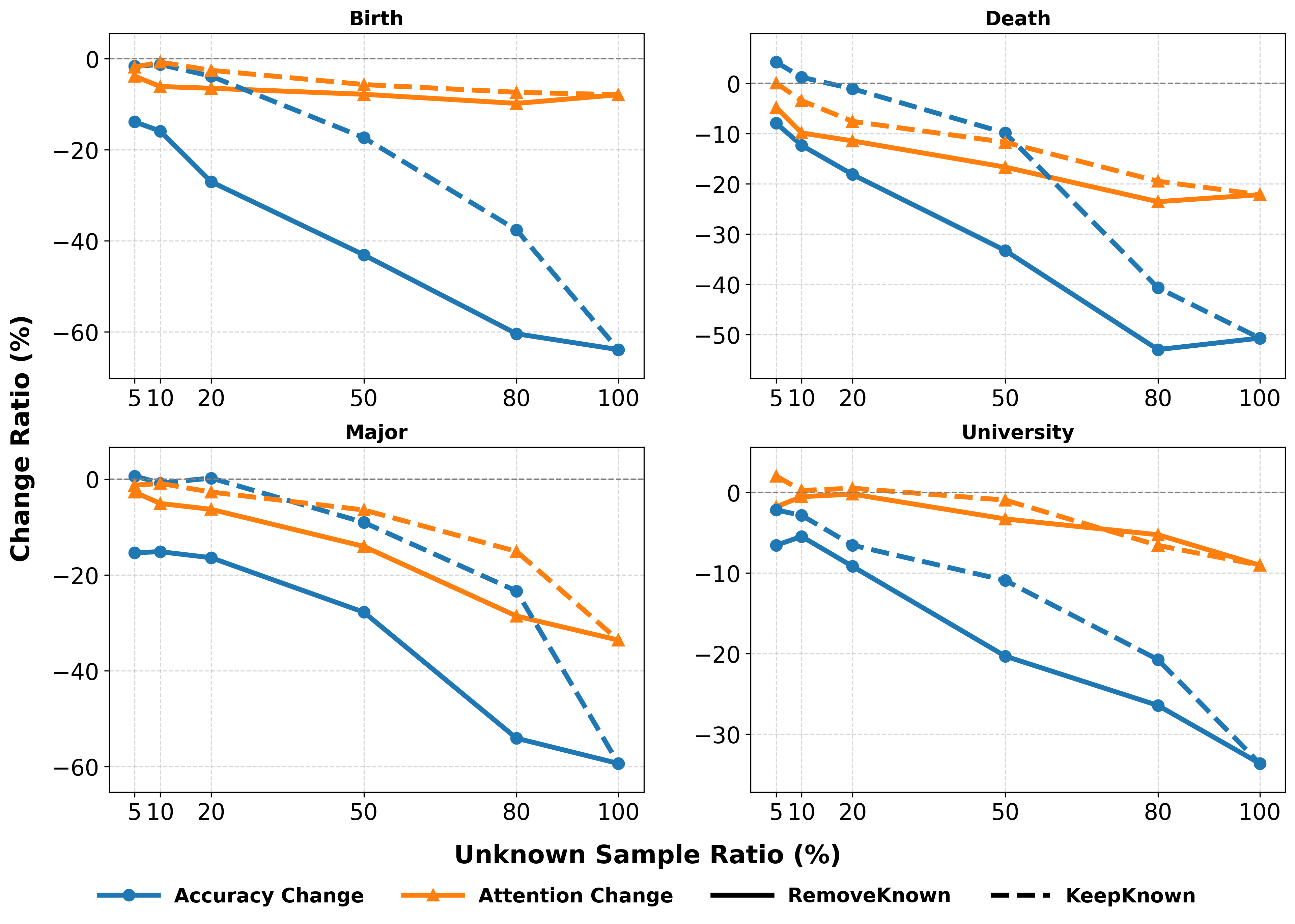}
    \caption{Accuracy and attention score changes with different unknown data ratio in certain type in QA tasks. All experiments trained for 1 epoch.}
    \label{fig:qa_unknown_percentage_interpretability_lines_1ep}
\end{figure}

\begin{figure}[ht]
    \centering
    \includegraphics[width=0.48\textwidth]{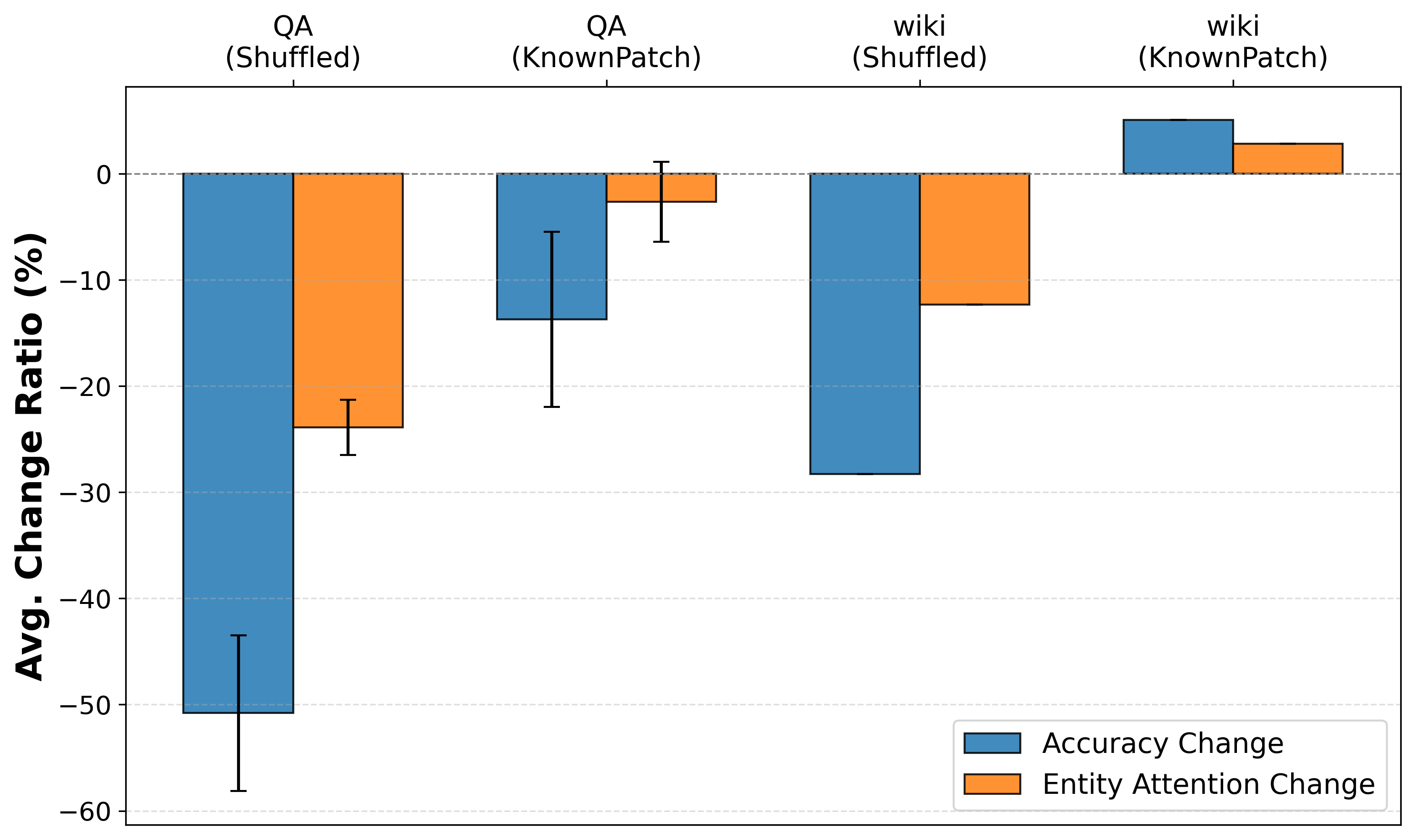}
    \caption{Performance and attention score changes when learning new knowledge in QA tasks, and after applying KnownPatch (with 20\% known data). QA represents the average across the four QA test sets, and error bars indicate standard deviations. All experiments trained for 1 epoch.}
    \label{fig:qa_mitigating_interpretability_bar_1ep}
\end{figure}

Table \ref{tab:qa_results_1ep} (similar to Table \ref{tab:qa_results}) provides the hallucination results in QA tasks when learning new knowledge; Figure \ref{fig:qa_unknown_percentage_1ep} (similar to Figure \ref{fig:qa_unknown_percentage}) shows the performance after learning different proportions of unknown knowledge; Figure \ref{fig:reasoning_big_table_1ep} (similar to Figure \ref{fig:reasoning_big_table}) shows the impact of new knowledge in reasoning tasks on different groups; 
Figure \ref{fig:KnownPatch_reasoning_20_1ep} (similar to Figure \ref{fig:KnownPatch_reasoning_20}) reports performance of KnownPatch on reasoning tasks when injecting 20\% known data; 
Figure \ref{fig:qa_mitigating_missing_cate_20_1ep} reports (similar to Figure \ref{fig:qa_mitigating_missing_cate_u800k200-3ep}) performance of KnownPatch when one knowledge type is missing with 20\% injection ratio; 
Figure \ref{fig:reasoning_interpretability_bar_1ep} (similar to Figure \ref{fig:reasoning_interpretability_bar}) reports the accuracy and attention score changes when learning new knowledge in reasoning tasks; Figure \ref{fig:qa_unknown_percentage_interpretability_lines_1ep} (similar to Figure \ref{fig:qa_unknown_percentage_interpretability_lines}) reports the accuracy and attention score changes after learning different proportions of unknown knowledge; Figure \ref{fig:qa_mitigating_interpretability_bar_1ep} (similar to Figure \ref{fig:qa_mitigating_interpretability_bar}) reports the performance and attention score changes before and after applying KnownPatch.

\begin{table}[ht]
    \centering
    \begin{tabular}{c c c}
    \toprule
    STQA & DTQA & Wiki \\
    \midrule
    -51.89 {\footnotesize($\pm$ 13.35)} 
    & -2.53 {\footnotesize($\pm$ 3.72)} 
    & -7.07 {\footnotesize($\pm$ 6.40)} \\
    \bottomrule
    \end{tabular}
    \caption{Hallucination induced by training on different unknown knowledge types in QA tasks. All experiments trained for 1 epoch.}
    \label{tab:qa_results_1ep}
\end{table}


\subsection{5 Epochs}
\begin{figure}[ht]
    \centering
    \includegraphics[width=0.48\textwidth]{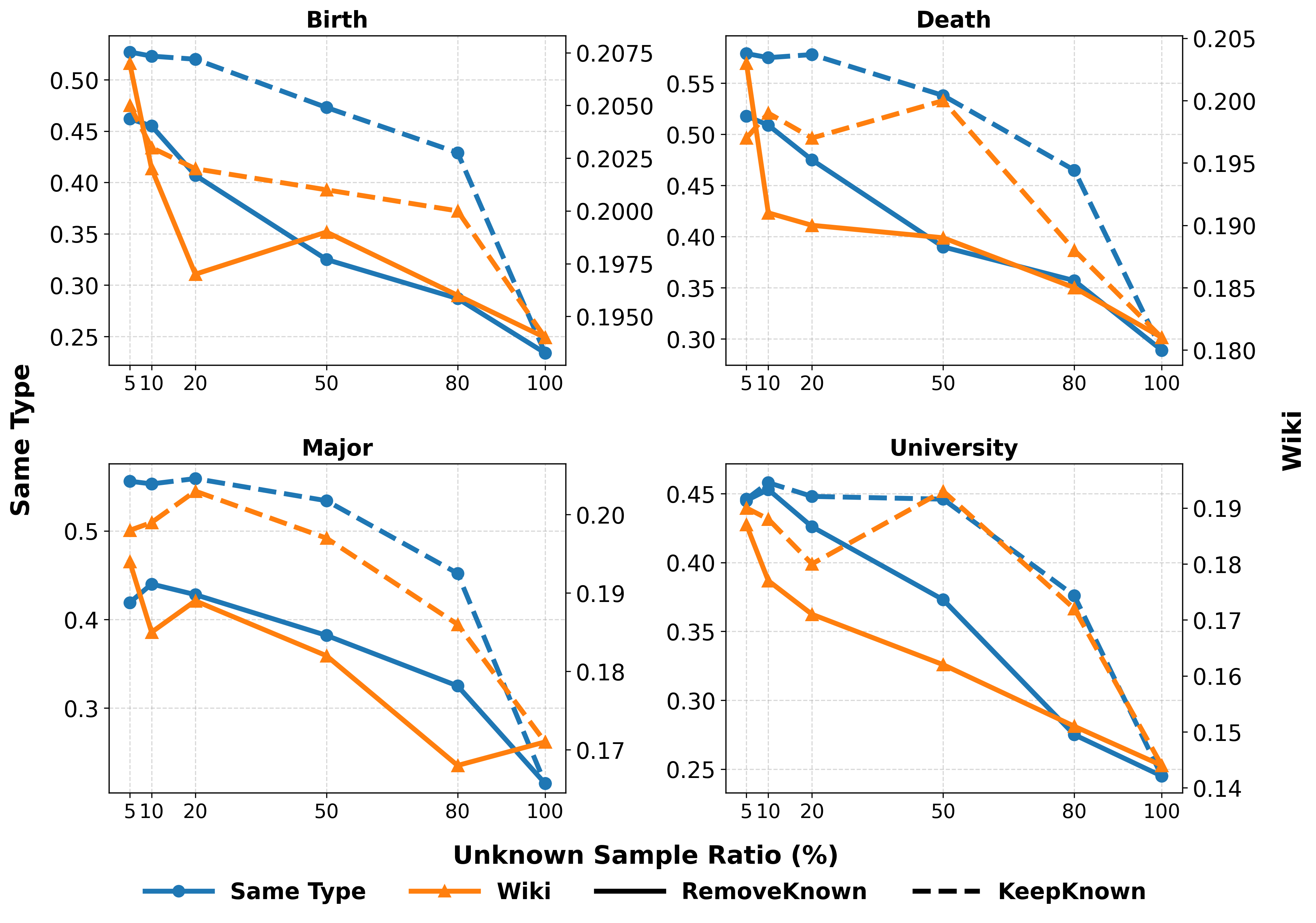}
    \caption{Performance in QA tasks under two settings with different proportions of unknown knowledge in the same type and wiki test set. All experiments trained for 5 epoch.}
    \label{fig:qa_unknown_percentage_5ep}
\end{figure}

\begin{figure}[ht]
    \centering
    \includegraphics[width=0.48\textwidth]{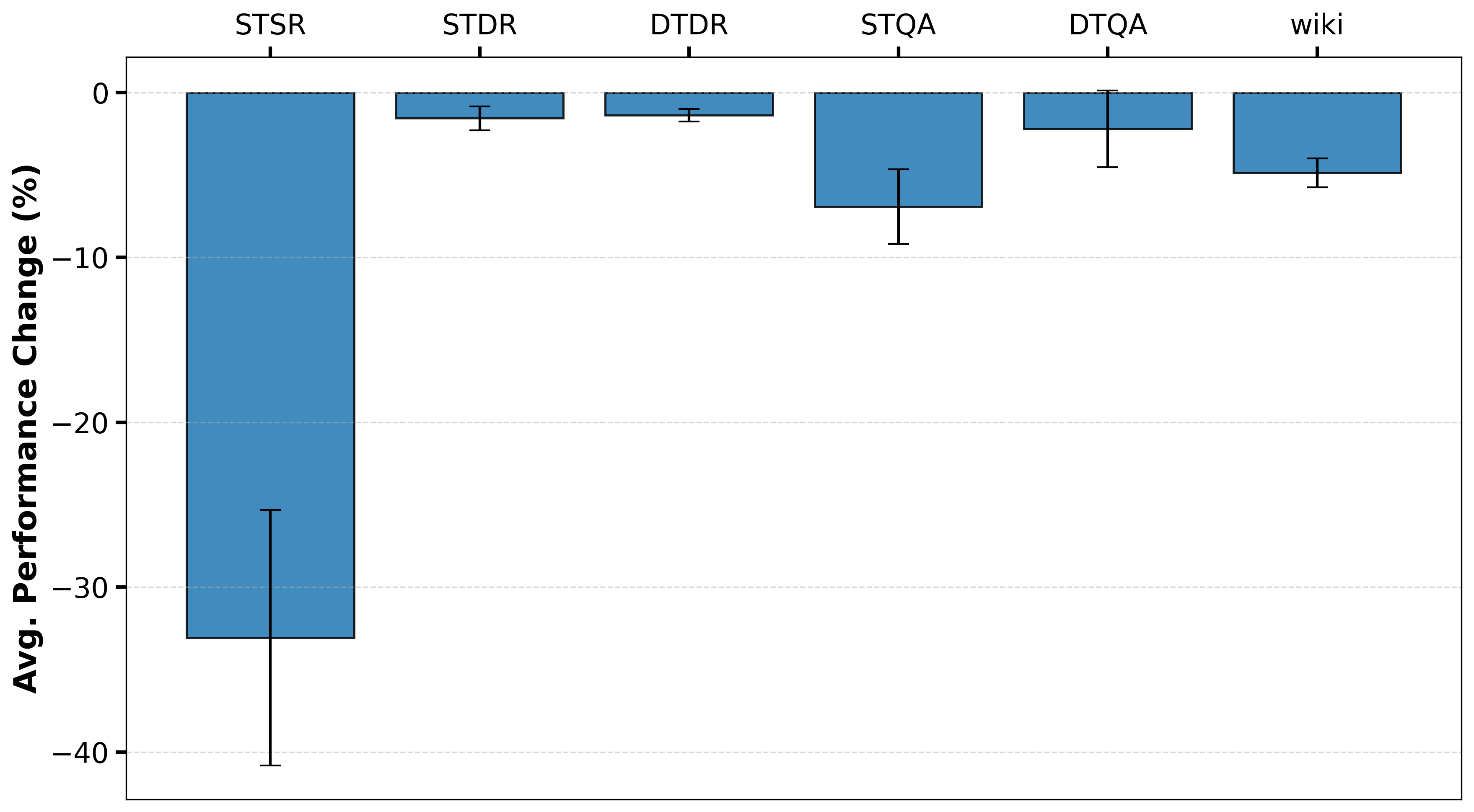}
    \caption{The impact of learning new knowledge in reasoning tasks on the average performance of different groups. All experiments trained for 5 epoch.}
    \label{fig:reasoning_big_table_5ep}
\end{figure}

\begin{figure}[ht]
    \centering
    \includegraphics[width=0.48\textwidth]{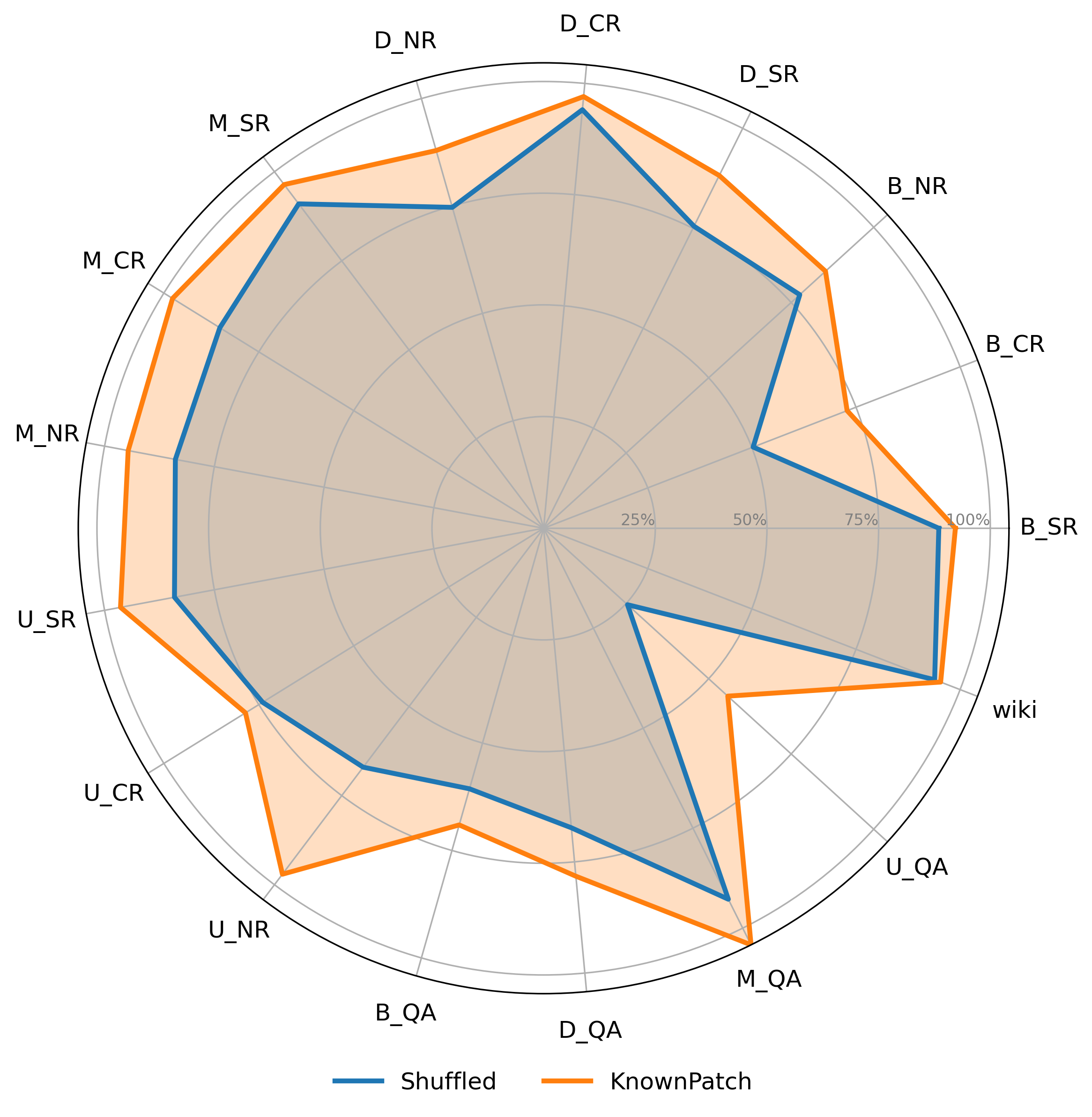}
    \caption{Performance of KnownPatch on reasoning task when injecting 20\% known data. The value here represents the accuracy percentage of this model compared to the fully known baseline model. All experiments trained for 5 epoch.}
    \label{fig:KnownPatch_reasoning_20_5ep}
\end{figure}

\begin{figure}[ht]
    \centering
    \includegraphics[width=0.48\textwidth]{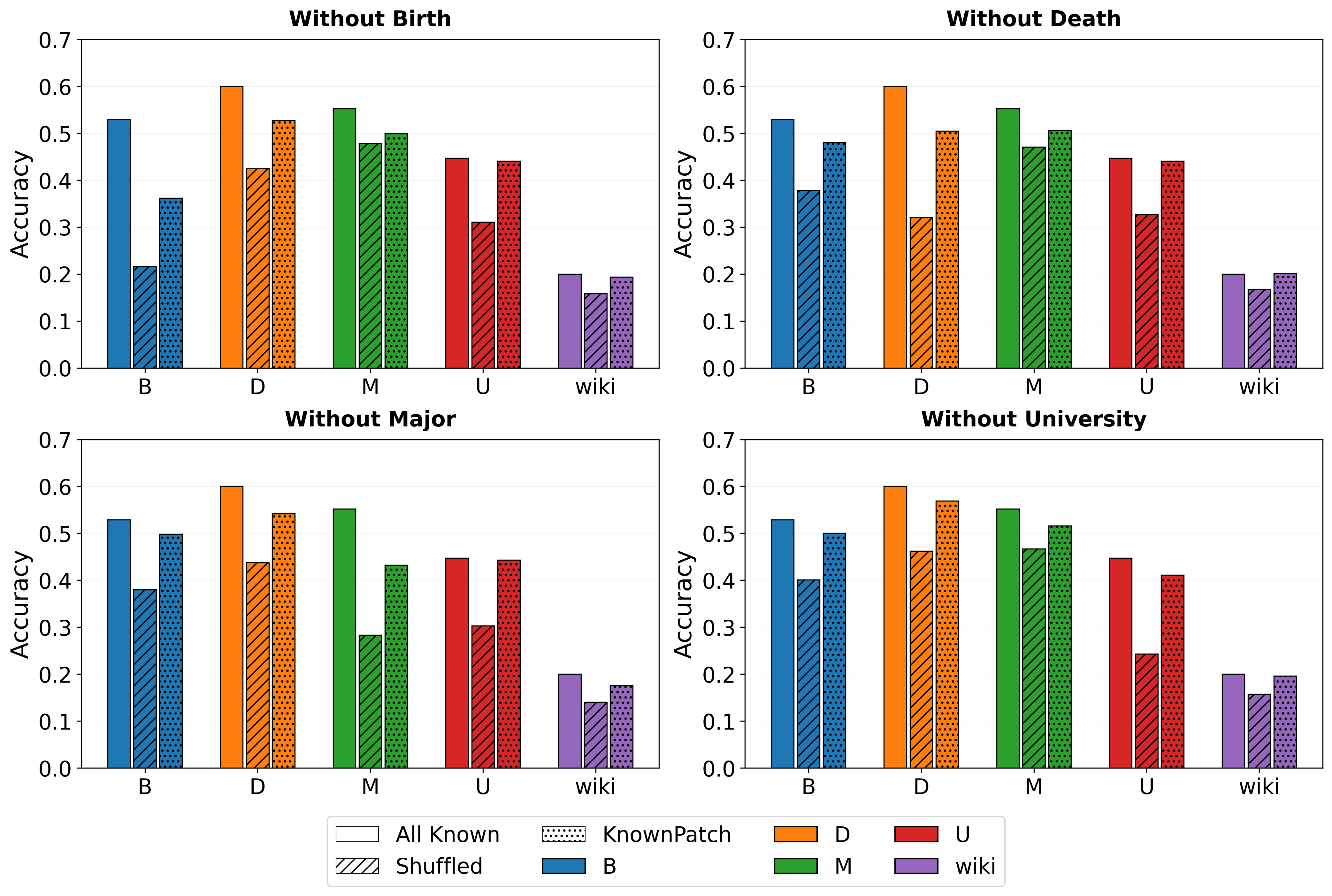}
    \caption{KnownPatch (missing one knowledge type) on QA tasks with an injection ratio of 20\%. All experiments trained for 5 epoch.}
    \label{fig:qa_mitigating_missing_cate_20_5ep}
\end{figure}

\begin{figure}[ht]
    \centering
    \includegraphics[width=0.48\textwidth]{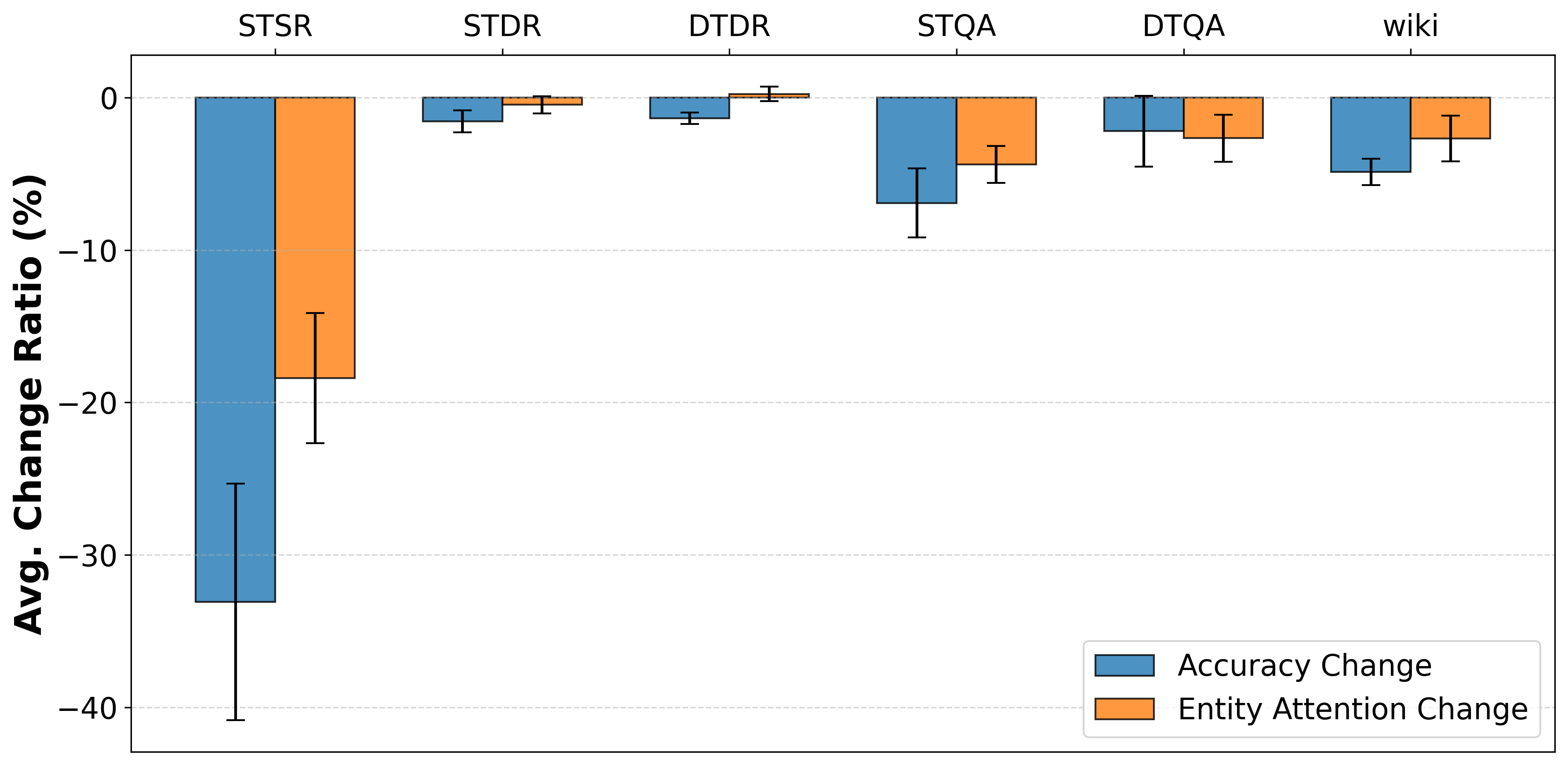}
    \caption{Accuracy and attention score changes when learning new knowledge in reasoning tasks. All experiments trained for 5 epoch.}
    \label{fig:reasoning_interpretability_bar_5ep}
\end{figure}

\begin{figure}[ht]
    \centering
    \includegraphics[width=0.48\textwidth]{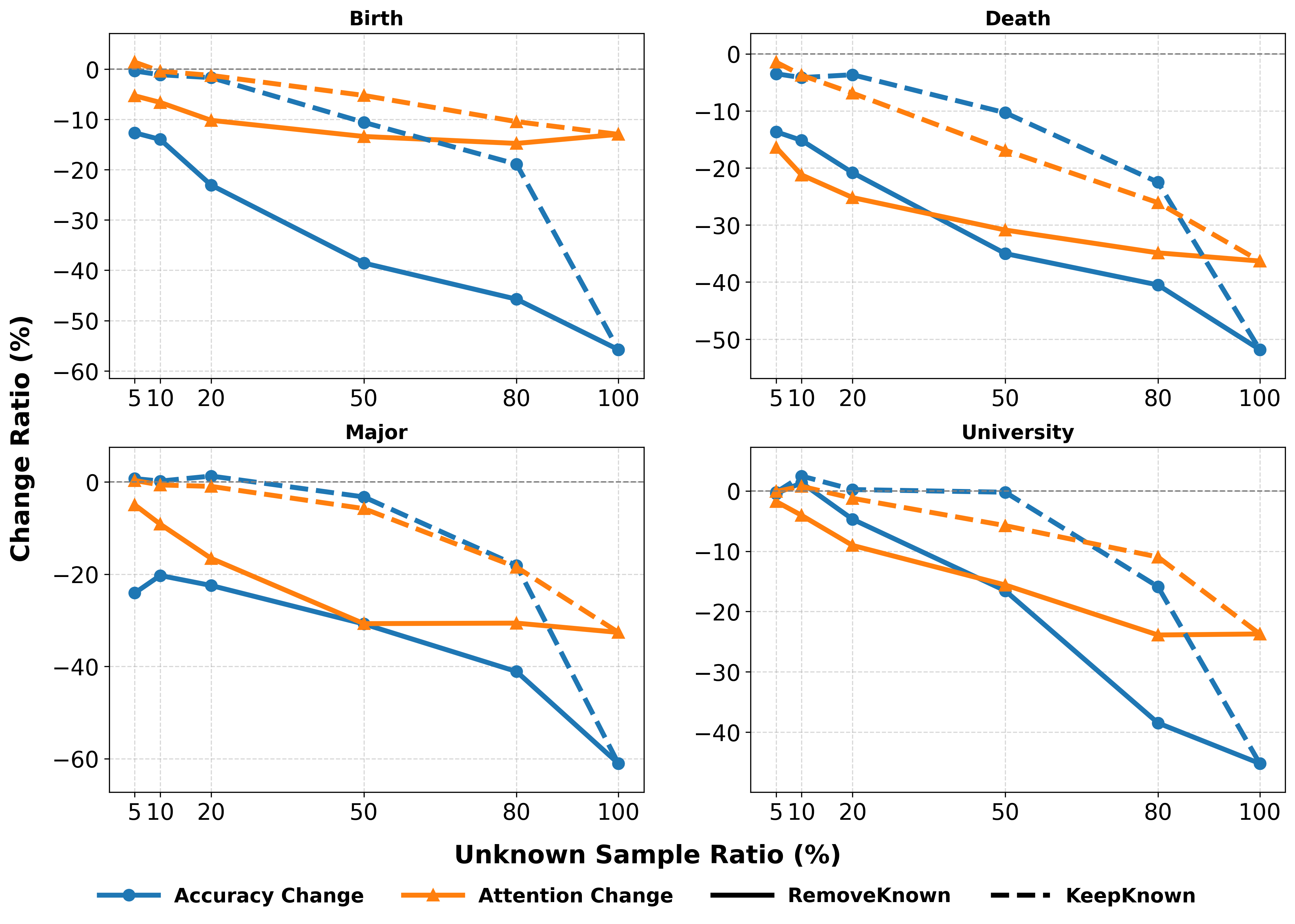}
    \caption{Accuracy and attention score changes with different unknown data ratio in certain type in QA tasks. All experiments trained for 5 epoch.}
    \label{fig:qa_unknown_percentage_interpretability_lines_5ep}
\end{figure}

\begin{figure}[ht]
    \centering
    \includegraphics[width=0.48\textwidth]{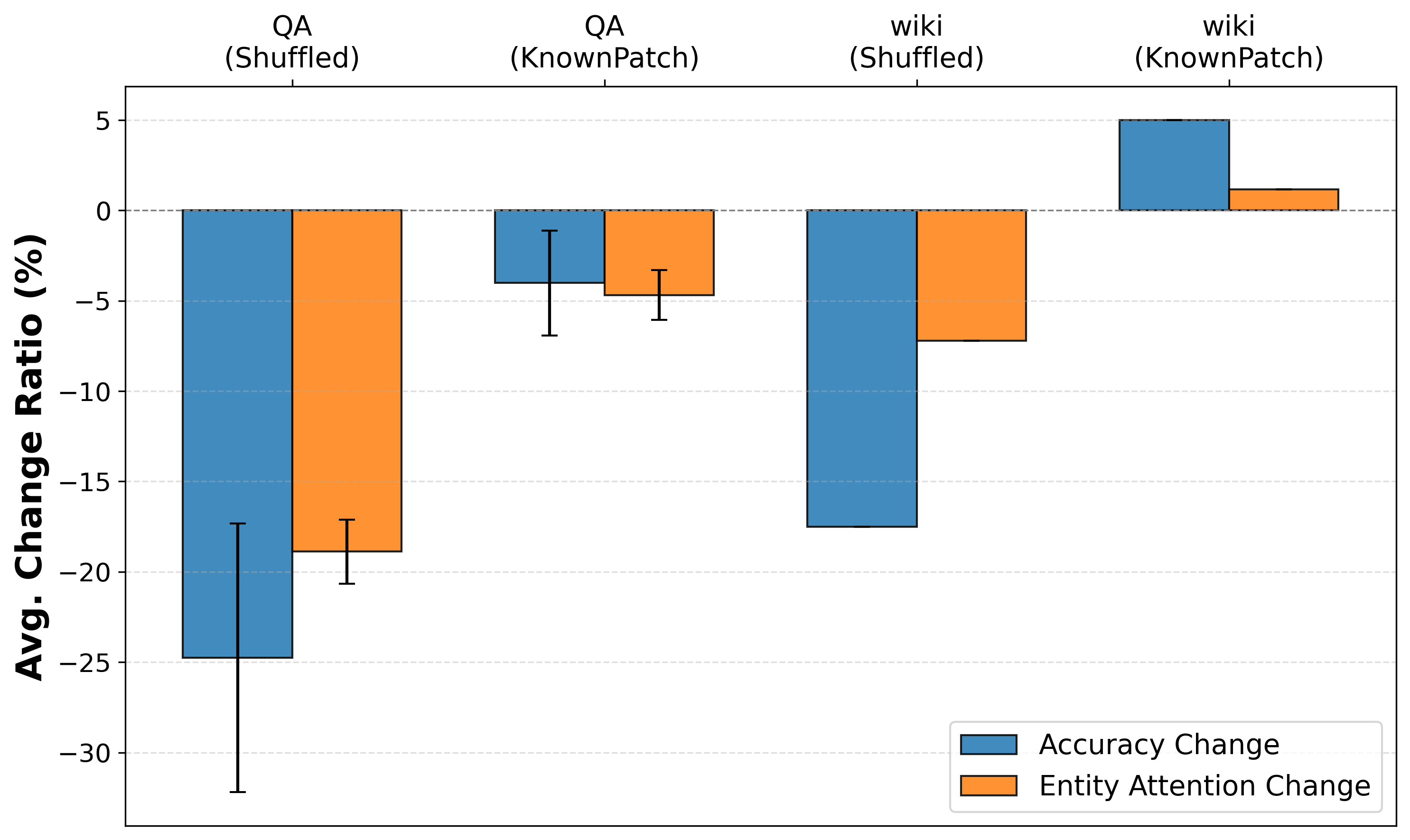}
    \caption{Performance and attention score changes when learning new knowledge in QA tasks, and after applying KnownPatch (with 20\% known data). QA represents the average across the four QA test sets, and error bars indicate standard deviations. All experiments trained for 5 epoch.}
    \label{fig:qa_mitigating_interpretability_bar_5ep}
\end{figure}

Table \ref{tab:qa_results_5ep} (similar to Table \ref{tab:qa_results}) provides the hallucination results in QA tasks when learning new knowledge; Figure \ref{fig:qa_unknown_percentage_5ep} (similar to Figure \ref{fig:qa_unknown_percentage}) shows the performance after learning different proportions of unknown knowledge; Figure \ref{fig:reasoning_big_table_5ep} (similar to Figure \ref{fig:reasoning_big_table}) shows the impact of new knowledge in reasoning tasks on different groups; 
Figure \ref{fig:KnownPatch_reasoning_20_5ep} (similar to Figure \ref{fig:KnownPatch_reasoning_20}) reports performance of KnownPatch when injecting 20\% known data; 
Figure \ref{fig:qa_mitigating_missing_cate_20_5ep} (similar to Figure \ref{fig:qa_mitigating_missing_cate_u800k200-3ep}) reports performance of KnownPatch when one knowledge type is missing when injecting 20\% known data; 
Figure \ref{fig:reasoning_interpretability_bar_5ep} (similar to Figure \ref{fig:reasoning_interpretability_bar}) reports the accuracy and attention score changes when learning new knowledge in reasoning tasks; Figure \ref{fig:qa_unknown_percentage_interpretability_lines_5ep} (similar to Figure \ref{fig:qa_unknown_percentage_interpretability_lines}) reports the accuracy and attention score changes after learning different proportions of unknown knowledge; Figure \ref{fig:qa_mitigating_interpretability_bar_5ep} (similar to Figure \ref{fig:qa_mitigating_interpretability_bar}) reports the performance and attention score changes before and after applying KnownPatch.

\begin{table}[ht]
    \centering
    \begin{tabular}{c c c}
    \toprule
    STQA & DTQA & Wiki \\
    \midrule
    -53.46 {\footnotesize($\pm$ 6.68)} 
    & -1.79 {\footnotesize($\pm$ 2.05)} 
    & -13.75 {\footnotesize($\pm$ 10.60)} \\
    \bottomrule
    \end{tabular}
    \caption{Hallucination induced by training on different unknown knowledge types in QA tasks. All experiments trained for 5 epoch.}
    \label{tab:qa_results_5ep}
\end{table}


\subsection{20 Epochs}
\begin{figure}[ht]
    \centering
    \includegraphics[width=0.48\textwidth]{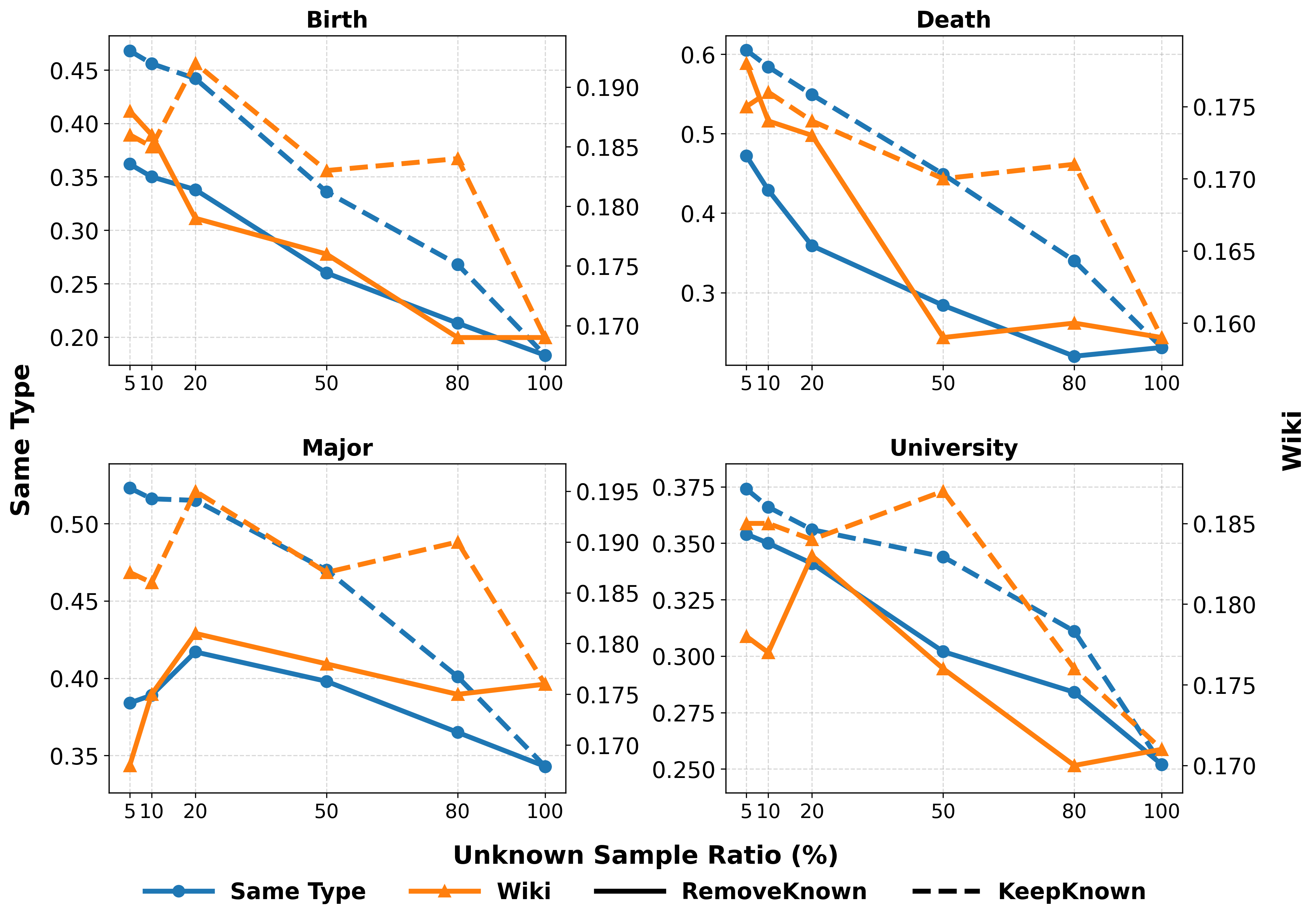}
    \caption{Performance in QA tasks under two settings with different proportions of unknown knowledge in the same type and wiki test set. All experiments trained for 20 epoch.}
    \label{fig:qa_unknown_percentage_20ep}
\end{figure}

\begin{figure}[ht]
    \centering
    \includegraphics[width=0.48\textwidth]{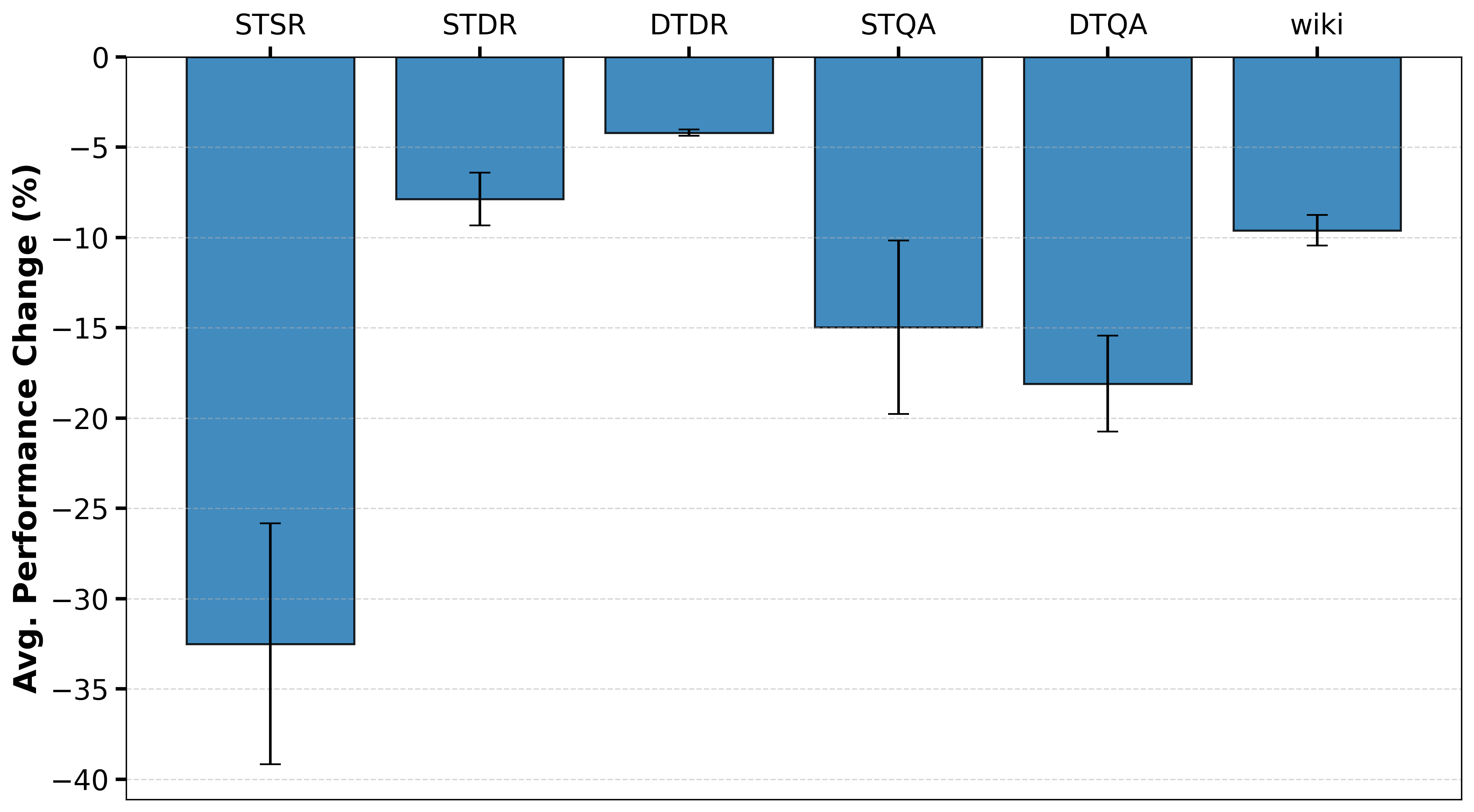}
    \caption{The impact of learning new knowledge in reasoning tasks on the average performance of different groups. All experiments trained for 20 epoch.}
    \label{fig:reasoning_big_table_20ep}
\end{figure}

\begin{figure}[ht]
    \centering
    \includegraphics[width=0.48\textwidth]{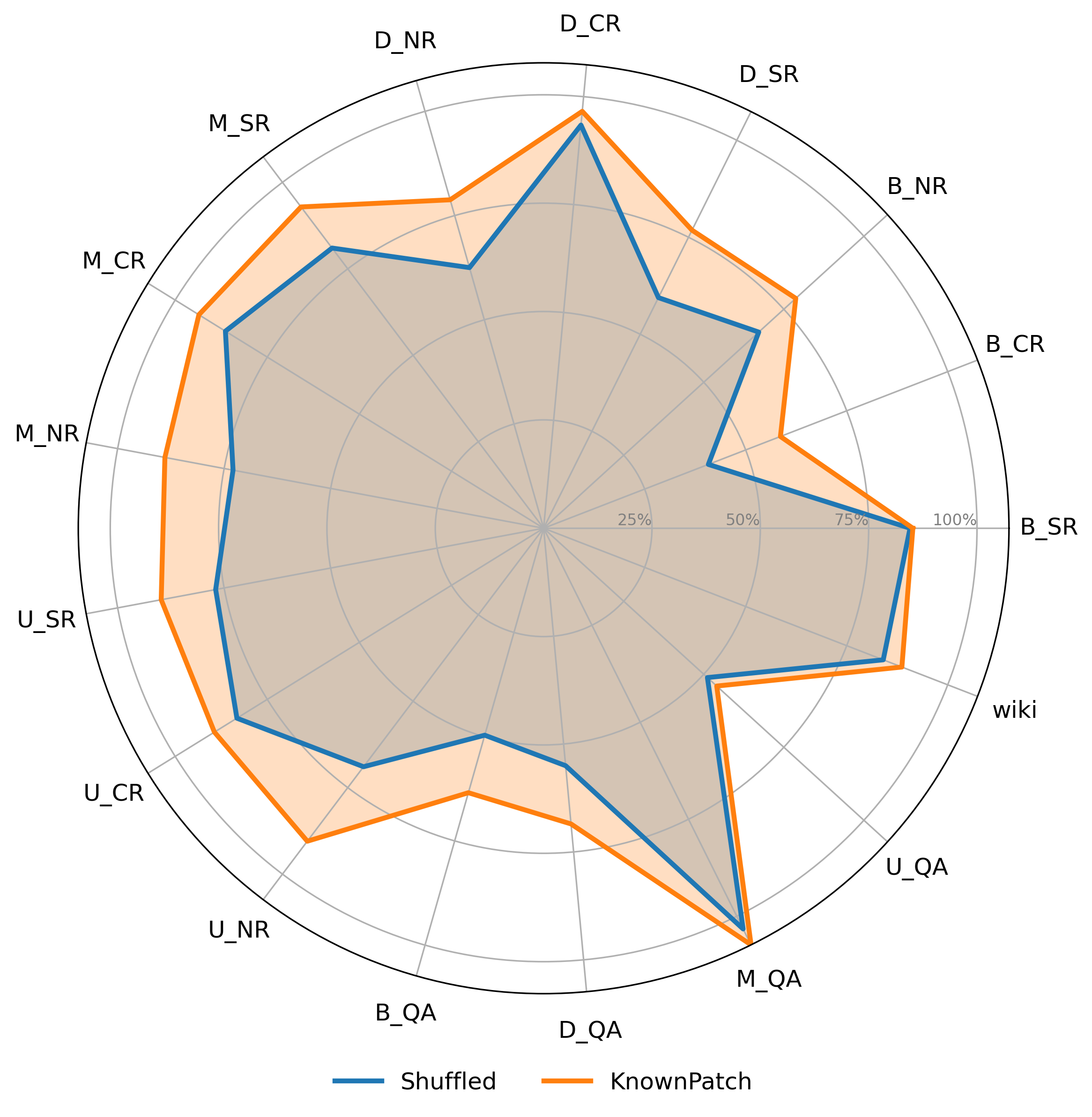}
    \caption{Performance of KnownPatch on reasoning task when injecting 20\% known data. The value here represents the accuracy percentage of this model compared to the fully known baseline model. All experiments trained for 20 epoch.}
    \label{fig:KnownPatch_reasoning_20_20ep}
\end{figure}

\begin{figure}[ht]
    \centering
    \includegraphics[width=0.48\textwidth]{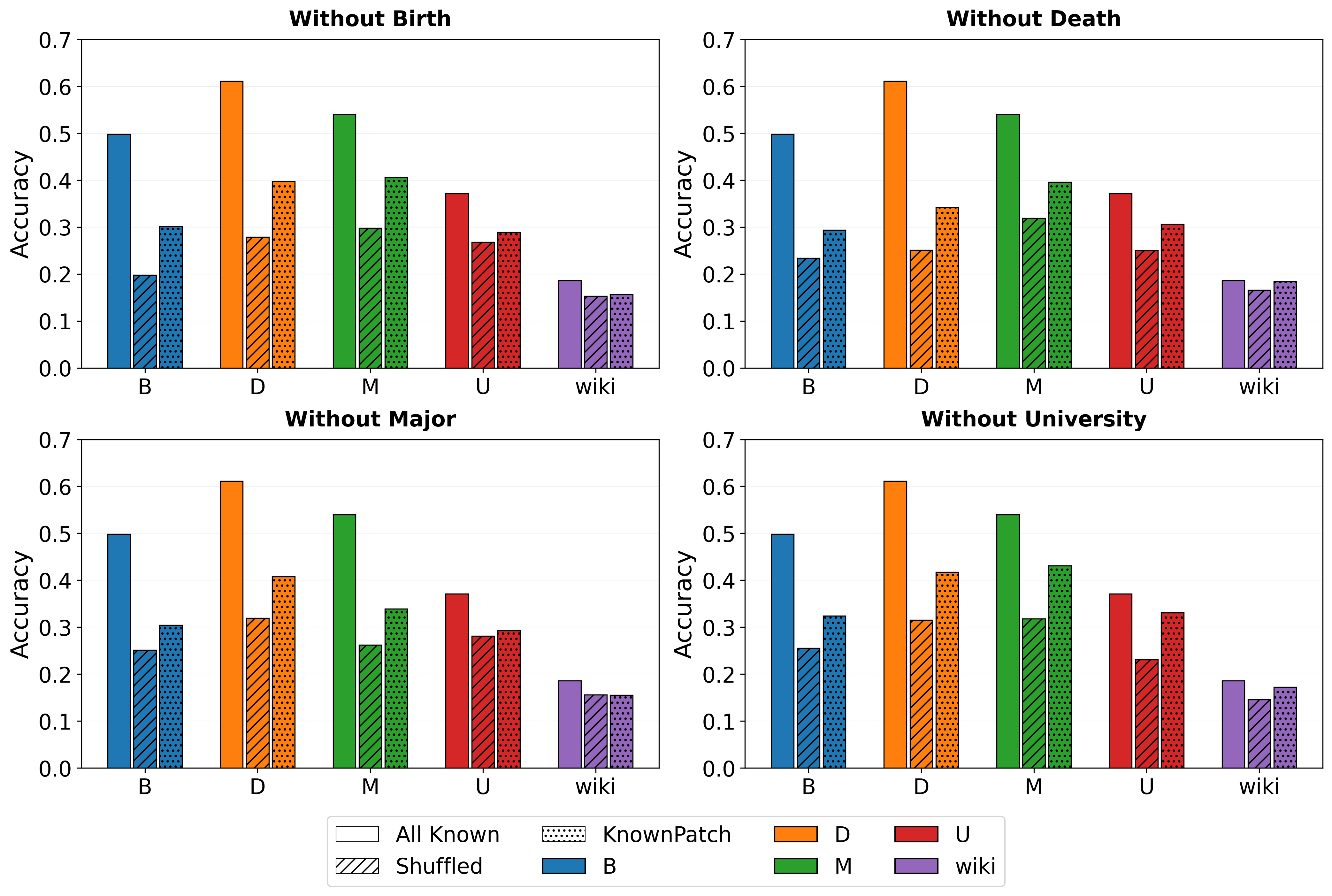}
    \caption{KnownPatch (missing one knowledge type) on QA tasks with an injection ratio of 20\%. All experiments trained for 20 epoch.}
    \label{fig:qa_mitigating_missing_cate_20_20ep}
\end{figure}

\begin{figure}[ht]
    \centering
    \includegraphics[width=0.48\textwidth]{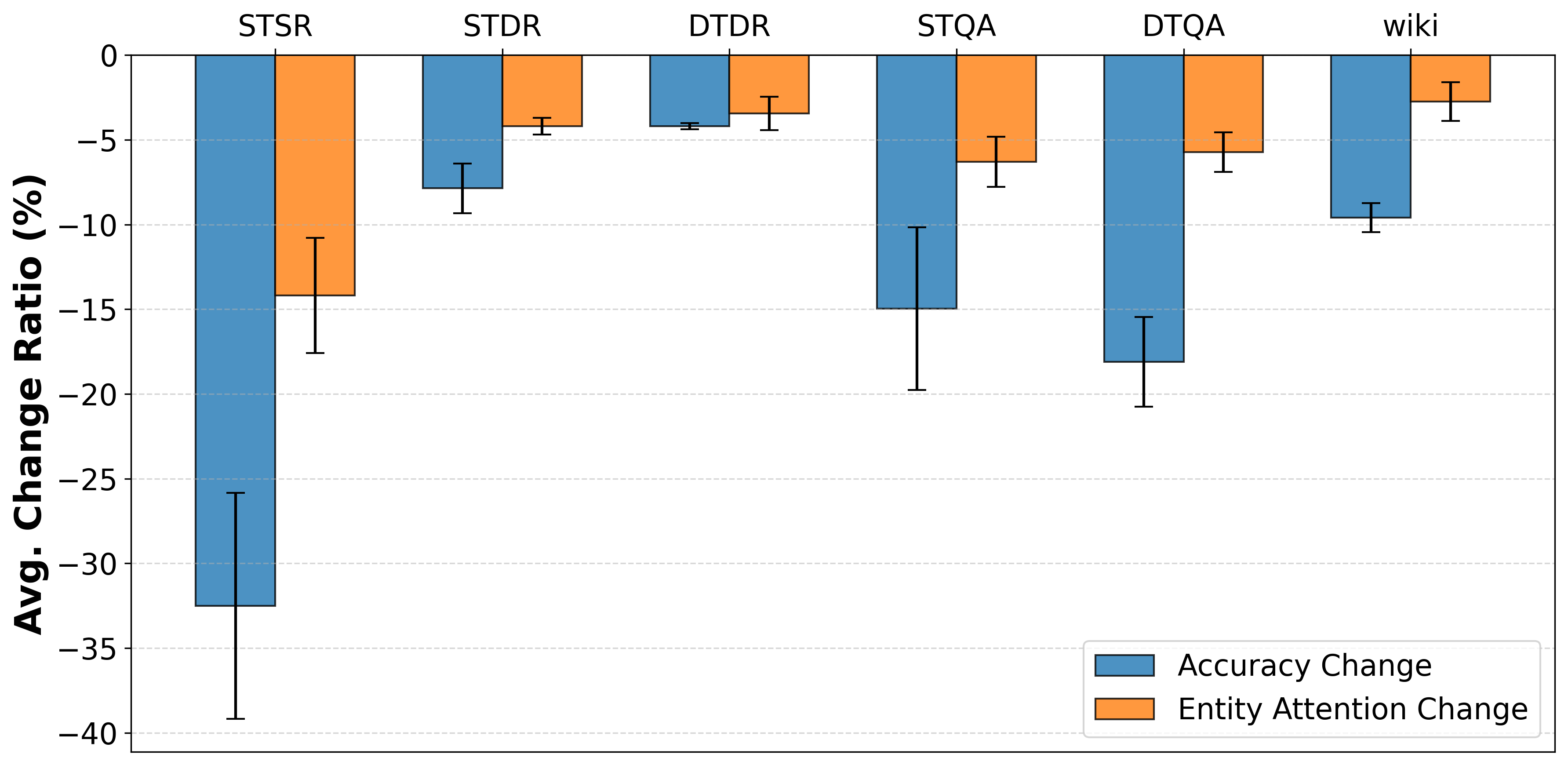}
    \caption{Accuracy and attention score changes when learning new knowledge in reasoning tasks. All experiments trained for 20 epoch.}
    \label{fig:reasoning_interpretability_bar_20ep}
\end{figure}

\begin{figure}[ht]
    \centering
    \includegraphics[width=0.48\textwidth]{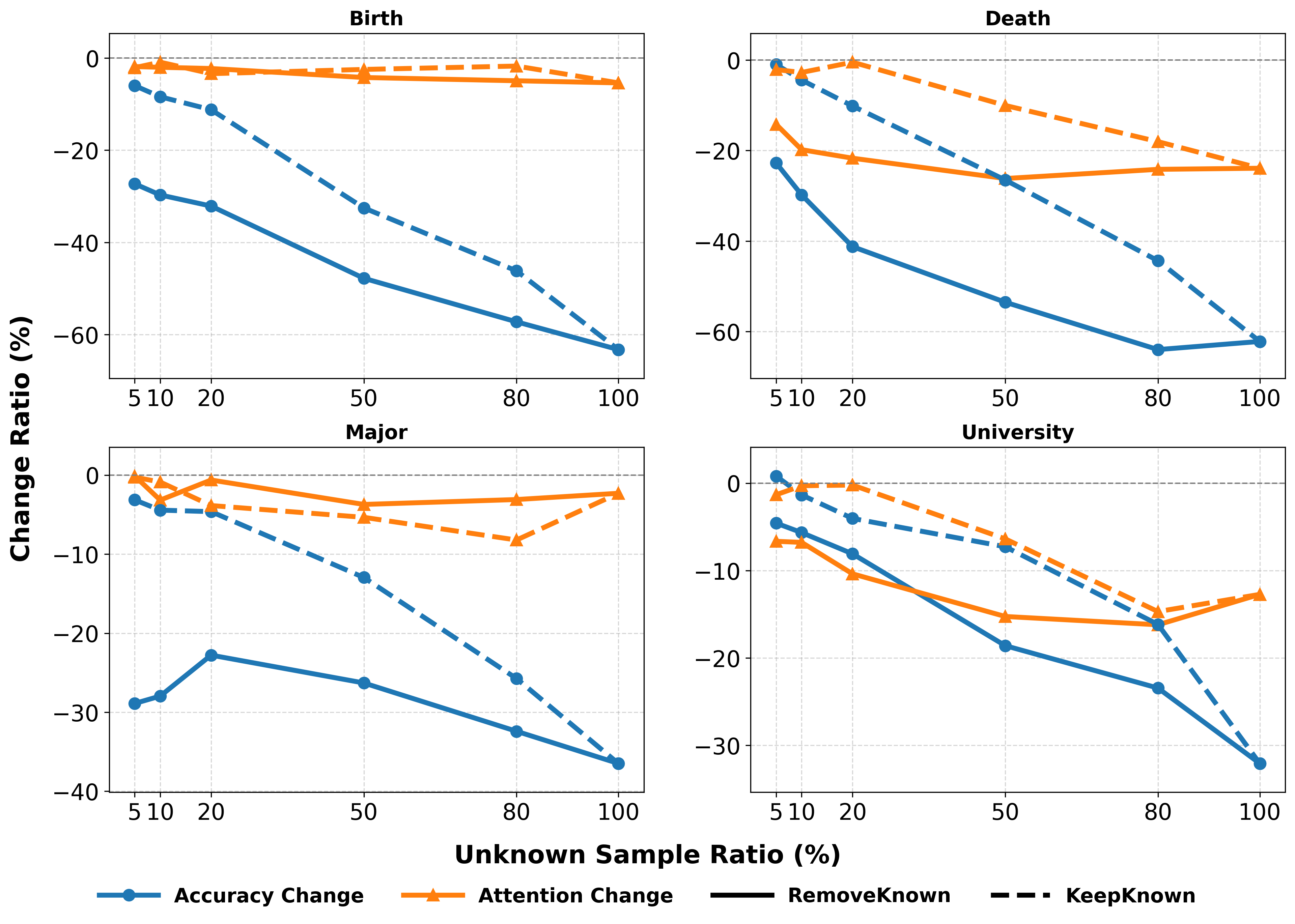}
    \caption{Accuracy and attention score changes with different unknown data ratio in certain type in QA tasks. All experiments trained for 20 epoch.}
    \label{fig:qa_unknown_percentage_interpretability_lines_20ep}
\end{figure}

\begin{figure}[ht]
    \centering
    \includegraphics[width=0.48\textwidth]{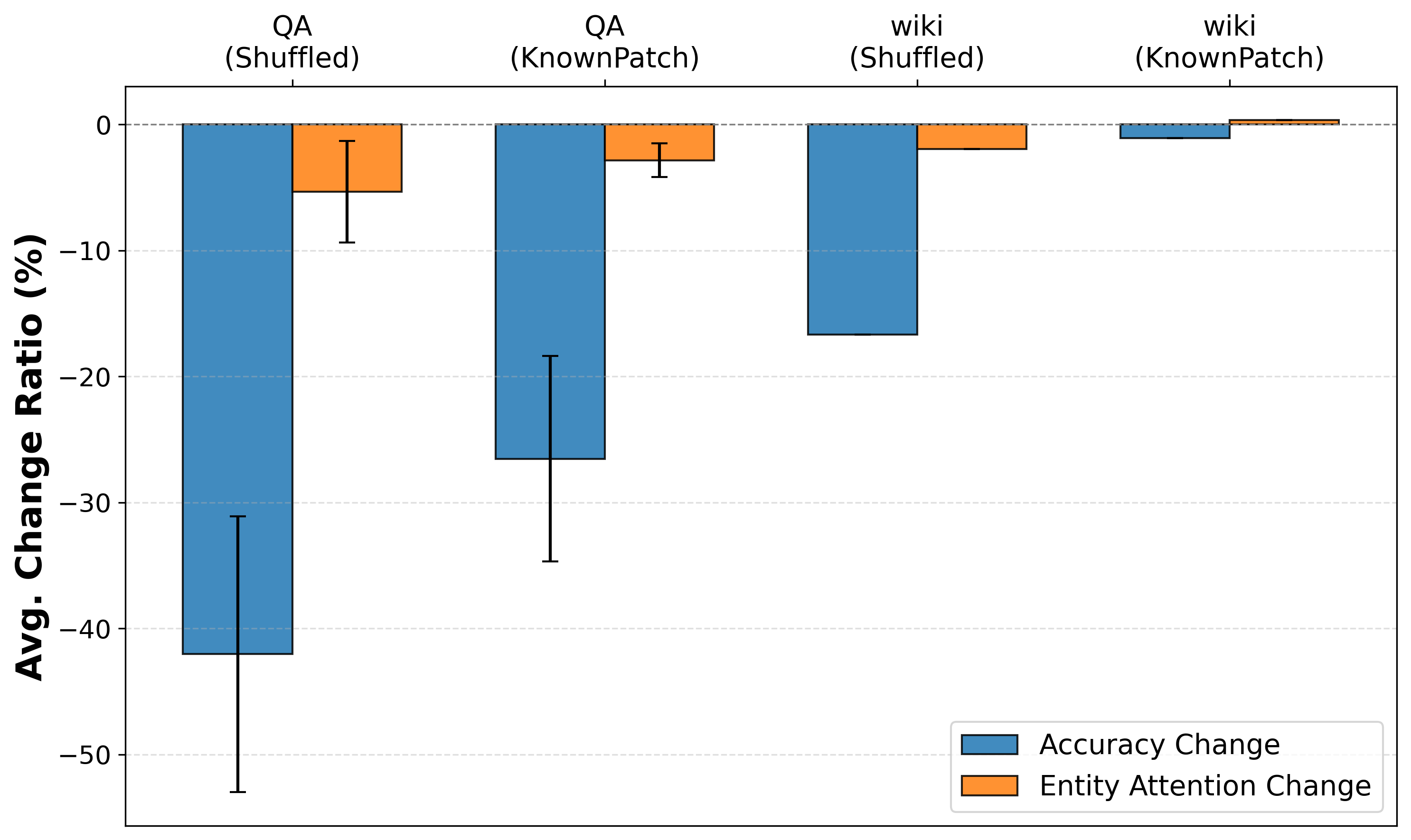}
    \caption{Performance and attention score changes when learning new knowledge in QA tasks, and after applying KnownPatch (with 20\% known data). QA represents the average across the four QA test sets, and error bars indicate standard deviations. All experiments trained for 20 epoch.}
    \label{fig:qa_mitigating_interpretability_bar_20ep}
\end{figure}

Table \ref{tab:qa_results_20ep} (similar to Table \ref{tab:qa_results}) provides the hallucination results in QA tasks when learning new knowledge; Figure \ref{fig:qa_unknown_percentage_20ep} (similar to Figure \ref{fig:qa_unknown_percentage}) shows the performance after learning different proportions of unknown knowledge; Figure \ref{fig:reasoning_big_table_20ep} (similar to Figure \ref{fig:reasoning_big_table}) shows the impact of new knowledge in reasoning tasks on different groups; 
Figure \ref{fig:KnownPatch_reasoning_20_20ep} (similar to Figure \ref{fig:KnownPatch_reasoning_20}) reports performance of KnownPatch when injecting 20\% known data; 
Figure \ref{fig:qa_mitigating_missing_cate_20_20ep} (similar to Figure \ref{fig:qa_mitigating_missing_cate_u800k200-3ep}) reports performance of KnownPatch when one knowledge type is missing when injecting 20\% known data; 
Figure \ref{fig:reasoning_interpretability_bar_20ep} (similar to Figure \ref{fig:reasoning_interpretability_bar}) reports the accuracy and attention score changes when learning new knowledge in reasoning tasks; Figure \ref{fig:qa_unknown_percentage_interpretability_lines_20ep} (similar to Figure \ref{fig:qa_unknown_percentage_interpretability_lines}) reports the accuracy and attention score changes after learning different proportions of unknown knowledge; Figure \ref{fig:qa_mitigating_interpretability_bar_20ep} (similar to Figure \ref{fig:qa_mitigating_interpretability_bar}) reports the performance and attention score changes before and after applying KnownPatch.

\begin{table}[ht]
    \centering
    \begin{tabular}{c c c}
    \toprule
    STQA & DTQA & Wiki \\
    \midrule
    -48.50 {\footnotesize($\pm$ 16.53)} 
    & -7.73 {\footnotesize($\pm$ 5.06)} 
    & -8.06 {\footnotesize($\pm$ 4.58)} \\
    \bottomrule
    \end{tabular}
    \caption{Hallucination induced by training on different unknown knowledge types in QA tasks. All experiments trained for 20 epoch.}
    \label{tab:qa_results_20ep}
\end{table}


\clearpage

\end{document}